\documentclass[preprint,12pt]{article}

\usepackage{amssymb}
\usepackage{amsmath}

\usepackage{longtable} 
\usepackage{subcaption}
\usepackage{adjustbox}
\usepackage{booktabs}
\usepackage{multirow}
\usepackage[table]{xcolor}
\usepackage{lscape}
\usepackage{array}
\usepackage{geometry}
\usepackage{bm}
\usepackage{caption}
\usepackage{graphicx}
\usepackage{pifont}
\usepackage{tikz}
\usepackage[normalem]{ulem} 
\usepackage[english]{babel}
\usepackage{natbib}
\usepackage{hyphenat}
\usepackage{textcomp}
\usepackage{mathtools}
\usepackage{comment}
\usepackage{pdflscape} 

\usepackage{tabularx} 

\usepackage{longtable}
\usepackage{algpseudocode}

\newcommand{\warningicon}{%
  \tikz[baseline=-0.5ex]\node[text=red, font=\bfseries]{!};%
}

\usepackage{pifont}
\newcommand{\cmark}{\textcolor{green}{\ding{51}}}%
\newcommand{\xmark}{\textcolor{red}{\ding{55}}}%

\newcommand{\albert}[1]{{\color{black}{#1}}}
\newcommand{\carlos}[1]{{\color{black}{#1}}}
\newcommand{\vanessa}[1]{{\color{black}{#1}}}

\usepackage[utf8]{inputenc} 
\usepackage[T1]{fontenc}    
\usepackage{hyperref}       
\usepackage{url}            
\usepackage{booktabs}       
\usepackage{amsfonts}       
\usepackage{nicefrac}       
\usepackage{microtype}      
\usepackage{xcolor}  

\usepackage{amsmath,amssymb,amsfonts}
\usepackage{graphicx}
\usepackage{algorithm}
\usepackage{hyperref}
\usepackage{textcomp}
\usepackage{mathtools}
\usepackage{comment}
\usepackage[symbol]{footmisc}
\providecommand{\keywords}[1]{\textbf{\textit{Index terms---}} #1}

\newcommand{\R}{\mathbb{R}} 
\newcommand{\E}{\mathbb{E}} 
\newcommand{\N}{\mathcal{N}} 
\newcommand{\M}{\mathcal{M}} 
\newcommand{\const}{\text{const}} 
\DeclareMathOperator{\Tr}{Tr} 

\DeclarePairedDelimiter\p{(}{)}
\DeclarePairedDelimiter\cor{[}{]}
\DeclarePairedDelimiter\llav{\{}{\}}
\DeclarePairedDelimiter\ang{\langle}{\rangle}

\DeclareMathOperator{\XndmT}{x_{n,d}^{(m)^\top}} 

\DeclareMathOperator{\acerotau}{\alpha_{0}^{\tau}}
\DeclareMathOperator{\bcerotau}{\beta_{0}^{\tau}}
\DeclareMathOperator{\taus}{\tau}
\DeclareMathOperator{\tauE}{\langle\tau\rangle}

\DeclareMathOperator{\unmed}{\frac{1}{2}}
\DeclareMathOperator{\lnpi}{\ln(2\pi)}
\DeclareMathOperator{\lntau}{\ln(\tau)}

\DeclareMathOperator{\taum}{\tau^{(m)}} 

\DeclareMathOperator{\tm}{\mathbf{T}^{(m)}} 
\DeclareMathOperator{\tnm}{\mathbf{t}_{n,:}^{(m)}} 
\DeclareMathOperator{\tndm}{t_{n,d}^{(m)}} 

\DeclareMathOperator{\sumn}{\sum\limits_{n}^{N}} 
\DeclareMathOperator{\summ}{\sum\limits_{m}^{M}} 
\DeclareMathOperator{\sumk}{\sum\limits_{k}^{K}} 
\DeclareMathOperator{\sums}{\sum\limits_{s}^{S}} 
\DeclareMathOperator{\sumd}{\sum\limits_{d}^{D^{(m)}}} 
\DeclareMathOperator{\sumc}{\sum\limits_{c}^{C}} 

\newcommand{\lnp}[1]{\ln\p*{#1}} 
\newcommand{\eqline}{\enskip&\enskip}
\newcommand{\eqeq}{\enskip=&\enskip}

\DeclareMathOperator{\lm}{\text{\boldmath$\lambda$}^{(m)}} 
\DeclareMathOperator{\ldm}{\lambda_d^{(m)}} 

\DeclareMathOperator{\y}{\mathbf{y}} 

\DeclareMathOperator{\Z}{\mathbf{Z}} 
\DeclareMathOperator{\ZT}{\mathbf{Z}^\top} 
\DeclareMathOperator{\Zn}{\mathbf{z}_{n,:}} 
\DeclareMathOperator{\ZnT}{\mathbf{z}_{n,:}^\top} 

\DeclareMathOperator{\Wk}{\mathbf{w}_{k,:}} 
\DeclareMathOperator{\Wkd}{w_{k,d}^{(m)}} 

\DeclareMathOperator{\X}{\mathbf{X}} 

\DeclareMathOperator{\Y}{\mathbf{Y}} 
\DeclareMathOperator{\Yn}{\mathbf{y}_{n,:}} 

\DeclareMathOperator{\V}{\mathbf{V}} 

\DeclareMathOperator{\Xm}{\mathbf{X}^{(m)}} 
\DeclareMathOperator{\XmT}{\mathbf{X}^{(m)^\top}} 
\DeclareMathOperator{\Xnm}{\mathbf{x}_{n,:}^{(m)}} 
\DeclareMathOperator{\XnmT}{\mathbf{x}_{n,:}^{(m)^\top}} 
\DeclareMathOperator{\Xndm}{x_{n,d}^{(m)}} 
\DeclareMathOperator{\Xdm}{\mathbf{x}_{:,d}^{(m)}} 
\DeclareMathOperator{\XM}{\mathbf{X}^{\{\M\}}} 

\DeclareMathOperator{\Wm}{\mathbf{W}^{(m)}} 
\DeclareMathOperator{\WmoT}{\mathbf{W}^{(M+1)^\top}} 
\DeclareMathOperator{\WmT}{\mathbf{W}^{(m)^\top}} 
\DeclareMathOperator{\Wkm}{\mathbf{w}_{k,:}^{(m)}} 
\DeclareMathOperator{\Wkdm}{w_{k,d}^{(m)}} 

\DeclareMathOperator{\gamm}{\text{\boldmath$\gamma$}^{(m)}} 
\DeclareMathOperator{\gamdm}{\gamma_d^{(m)}} 
\DeclareMathOperator{\dm}{\text{\boldmath$\delta$}^{(m)}} 
\DeclareMathOperator{\dsm}{\delta_s^{(m)}} 

\DeclareMathOperator{\tpredc}{t_{\ast,c}}

\DeclareMathOperator{\ypredc}{y_{\ast,c}}

\DeclareMathOperator{\xpredin}{\mathbf{x}_{\ast,:}^{\M}}

\DeclareMathOperator{\Vm}{\mathbf{V}^{(m)}} 
\DeclareMathOperator{\VmT}{\mathbf{V}^{(m)^\top}} 
\DeclareMathOperator{\Vsm}{\mathbf{v}_{:,s}^{(m)}} 
\DeclareMathOperator{\Vdm}{\mathbf{v}_{d,:}^{(m)}} 
\DeclareMathOperator{\VdmT}{\mathbf{v}_{d,:}^{(m)^\top}} 
\DeclareMathOperator{\Vdsm}{v_{d,s}^{(m)}} 

\DeclareMathOperator{\U}{\mathbf{U}} 
\DeclareMathOperator{\UT}{\mathbf{U}^{\top}} 

\DeclareMathOperator{\G}{\mathbf{G}} 
\DeclareMathOperator{\GT}{\mathbf{G}^\top} 
\DeclareMathOperator{\Gn}{\mathbf{g}_{n,:}} 
\DeclareMathOperator{\GnT}{\mathbf{g}_{n,:}^\top} 

\DeclareMathOperator{\Ik}{\mathbf{I}_{K}} 
\DeclareMathOperator{\Is}{\mathbf{I}_{S}} 
\DeclareMathOperator{\Ic}{\mathbf{I}_{C}} 

\DeclareMathOperator{\psim}{\psi^{(m)}} 
\DeclareMathOperator{\deltam}{\text{\boldmath$\delta$}^{(m)}} 
\DeclareMathOperator{\deltasm}{\delta_s^{(m)}} 
\DeclareMathOperator{\phim}{\text{\boldmath$\phi$}^{(m)}} 
\DeclareMathOperator{\phikm}{\phi_k^{(m)}} 

\title{Interpretable Generative and Discriminative Learning for Multimodal and Incomplete Clinical Data}

\author{
  Albert Belenguer-Llorens$^{1,2,}$\thanks{Corresponding author (e-mail: abelenguer@ethz.ch)} \and
  Carlos Sevilla-Salcedo$^{2}$ \and
  Janaina Mourao-Miranda$^{3}$ \and
  Vanessa G\'{o}mez-Verdejo$^{2,4}$
}

\date{
\footnotesize
  $^{1}$ Department of Health Sciences and Technology (DHEST), ETH Zurich, 8092, Zürich, Switzerland \\
  $^{2}$ Department of Signal Theory and Communications, Universidad Carlos III de Madrid, Leganés, 28911, Spain \\
  $^{3}$ UCL Hawkes Institute, Department of Computer Science, University College London, London, UK \\
  $^{4}$ Instituto de Investigación Sanitaria Gregorio Marañón (IiSGM), Madrid, 28009 Spain
}

\begin{document}

\maketitle

\begin{abstract}

Real-world clinical problems are often characterized by multimodal data, usually associated with incomplete views and limited sample sizes in their cohorts, posing significant limitations for machine learning algorithms. In this work, we propose a Bayesian approach designed to efficiently handle these challenges while providing interpretable solutions. Our approach integrates (1) a generative formulation to capture cross-view relationships with a semi-supervised strategy, and (2) a discriminative task-oriented formulation to identify relevant information for specific downstream objectives. This dual generative-discriminative formulation offers both general understanding and task-specific insights; thus, it provides an automatic imputation of the missing views while enabling robust inference across different data sources. The potential of this approach becomes evident when applied to the multimodal clinical data, where our algorithm is able to capture and disentangle the complex interactions among biological, psychological, and sociodemographic modalities.

\end{abstract}


\section{Introduction}
\label{sec:intro}

Advances in modern medicine increasingly rely on the integration of heterogeneous data sources, including neuroimaging, clinical variables, molecular profiles, and psychosocial assessments, to obtain a comprehensive view of disease mechanisms. This multimodal perspective is particularly valuable for complex and heterogeneous disorders, such as mental and neurological conditions, where no single modality is sufficient to capture the biological, psychological, and social factors involved \cite{who2001icf}. By combining complementary modalities, researchers can achieve a richer understanding of disease mechanisms and improve the accuracy of prognostic and diagnostic tools.
However, analysing multimodal datasets brings persistent and interrelated challenges. 

One of the most persistent obstacles in multimodal clinical data is the presence of incomplete datasets. In practice, neuroimaging scans may be missing due to contraindications or limited scanner availability, molecular profiling may be skipped to reduce costs, and psychosocial questionnaires may be incomplete due to patient fatigue. Such missingness is rarely random; rather, it reflects structured patterns in data acquisition shaped by site-specific protocols, patient subpopulation characteristics, and cost-driven prioritization of certain modalities 
\cite{austin2021missing}. 
Conventional imputation techniques, when applied without accounting for these patterns, risk amplifying acquisition biases, inadvertently introducing spurious associations, and degrading the reliability of downstream predictions 
\cite{heymans2022handling}.

A second major limitation in clinical studies is the scarcity of data. Even large-scale collaborative efforts often produce datasets that are modest in size when compared to the high dimensionality of multimodal measurements. Recruitment challenges, the rarity of certain conditions, and the high cost of acquisition protocols all contribute to small sample sizes 
\cite{cao2024small}. 
This scarcity places strict limits on the complexity of models that can be reliably trained, forcing researchers to either reduce the dimensionality of the data—potentially discarding informative features—or risk overfitting. 

In addition to these statistical and methodological issues, clinical adoption of Machine Learning (ML) tools depends critically on interpretability. Clinicians must be able to understand and trust a model’s reasoning before integrating it into diagnostic or prognostic decision-making. While post-hoc explanation techniques 
\cite{lundberg2017unified} 
can provide useful local insights, they often lack global consistency, making it difficult to ensure that the model’s decisions remain coherent across patients and settings. In safety-critical domains such as healthcare, this limitation poses a barrier to real-world deployment, as opaque models can fail silently in ways that are difficult to detect.

These three challenges, structured missingness, limited data availability, and the need for trustworthy interpretability, are deeply intertwined. A solution that handles missing data but cannot generalize in low-sample-size regimes, or one that achieves strong predictive accuracy but remains opaque to human experts, will fall short of the requirements for clinical impact. While existing ML techniques offer partial remedies to these problems, they are rarely addressed in a unified manner. Generative models, for example, can effectively capture multimodal relations and naturally support missing-data imputation, but their lack of task-oriented optimization often limits predictive performance. Conversely, discriminative models excel at task-specific objectives but cannot directly take advantage of partially observed data without a separate imputation step, which may discard valuable cross-modal information. Addressing these limitations in isolation has yielded partial progress, but a more promising path lies in designing integrated approaches that explicitly combine solutions for all three challenges within a single framework.

Building on this motivation, we propose the \textit{Optimized Sparse Inference and Reconstruction for Interdependent Spaces} (OSIRIS), a unified ML model designed to address these three challenges concurrently. OSIRIS combines the strengths of generative and discriminative modeling within a single architecture, enabling it to learn joint latent representations that capture cross-modal structure while remaining directly optimized for the downstream clinical task. The generative component allows the model to integrate information from incomplete or partially observed datasets without the need for a separate imputation pipeline, preserving subtle cross-modal dependencies that might otherwise be lost. The discriminative component ensures that the learned representations are explicitly tuned for predictive accuracy, even in small-sample regimes. \albert{Thus, OSIRIS has three main advantages: (1) As a Bayesian model, OSIRIS is able to quantify the uncertainty of the predicted outputs, a critical and required capability when working in medical domains. (2) It integrates sparse priors within its Bayesian formulation, enabling a genuine Feature Selection (FS) and the identification of a minimal subset of relevant biomarkers. (3) It is adaptable across data modalities and clinical applications, making it suitable for scenarios where it is desirable to combine multiple sources of information for prediction and disease understanding, e.g., health outcome prediction and biomarker discovery.}
This last point is possible thanks to its inherently interpretable nature, which is crucial when working in clinical scenarios.

The remainder of this work is organized as follows. Section~\ref{sec:related} reviews existing approaches for multimodal learning and highlights their limitations in addressing these challenges jointly. Section~\ref{sec:method} presents the proposed OSIRIS model. Section \ref{sec:exp} evaluates the discriminative and generative capabilities of the proposed model over several (clinical and no clinical) datasets, as well as its advantages in a challenging multimodal clinical datasets, and Section~\ref{sec:conclusions} discusses the implications of our findings, as well as some limitations.

\section{Related Work}
\label{sec:related}

Current state-of-the-art methods offer only partial solutions to the aforementioned challenges, with no unified and efficient approach addressing all challenges at once. To clarify existing strategies, we next review multimodal approaches and how they handle missing data, small samples, and interpretability.

The classical ML models, particularly linear ones, are widely used in multimodal analysis due to their interpretability. A common strategy is to concatenate all modalities into a single input vector \cite{fang2023comprehensive}. While simple, this often results in information loss when modalities are context-dependent, as in neuroimaging or genetic data, where features (e.g., voxels) meaning depends on the context \cite{yang2018multi}. To avoid this, alternative models aim to preserve modality-specific structure. Among them, Self-Representation (SR) methods are valued for their simplicity and interpretability 
\cite{liu2022efficient}. These approaches represent each sample in a shared space using weighted linear combinations of its modalities, enabling insight into the contribution of each view. Graph-based models offer another alternative, combining different modalities while leveraging cross-modal dependencies
through graph relations \cite{ektefaie2023multimodal}. 
In clinical settings, Multiple Kernel Learning SVMs (MKL-SVMs) are widely used due to their ability to model nonlinear cross-modal relationships via flexible kernel combinations 
\cite{zhang2024explaining}. 
However, when kernel construction is decoupled from the learning task, relevant information may be lost, and interpretability is limited to linear kernels. To improve alignment with prediction tasks, supervised multimodal methods like Partial Least Squares (PLS) have been proposed 
\cite{belenguer2024unified}. 
These models learn a shared latent space optimized for the target, improving task-specific performance while maintaining interpretability. Still, most of these approaches cannot natively handle missing data and rely on imputation techniques (which are often applied before the training). This may introduce bias, especially if missingness correlates with the target, and often ignores cross-modal dependencies unless views are concatenated. In contrast, multimodal Factor Analysis (FA) models 
\cite{anceschi2024bayesian} 
can handle and impute missing data while maintaining interpretability in their predictions. Nevertheless, these models are fundamentally generative in nature and are not specifically designed for predictive tasks, resulting in their frequent tendency to obtain less-than-ideal outcomes in terms of classification performance. \albert{In the medical domain, the use of such models appears to be particularly prominent in the field of multi-omics, where we can start to find algorithms capable of integrating multimodal omics datasets in a semi-supervised manner \cite{samorodnitsky2024bayesian}.}

Deep Learning (DL) models have emerged as powerful tools for multimodal data integration due to their capacity to model complex structures. Notable examples include Trusted Multi-view Classification (TMC) \cite{han2022trusted}, \albert{which dynamically integrates different data views by using the Dempster-Shafer theory while considering the evidence of each view}, and Cauchy–Schwarz
Multi-View Information Bottleneck (CSMVIB) \cite{zhang2025towards}, \albert{which, taking advantage of the mutual information measurement, defines a robust Neural Network (NN) able to compress and combine multimodal data into a low-dimensional, more representative space.} Some DL approaches combine multimodal data with the imputation of missing data. \albert{Within this group, we highlight} Cross Partial Multi-View Networks (CPM) \cite{zhang2019cpm}, a deep generative model that \albert{captures and combines the different views by searching the optimal trade-off between consistency and complementarity}; \albert{and} Deep Variational Information Bottleneck model (DeepIMV), \cite{lee2021variational} \albert{which} \albert{which uses a variational bottleneck approach to merge the information into low-dimensional representations capturing both intra-view and inter-view interactions.} Despite these advances, DL models face two major limitations in clinical settings: (1) poor generalization when trained on small datasets, and (2) limited interpretability, which can hinder clinical adoption. To address the first issue, Foundation Models (FMs) pre-trained on large datasets 
(e.g., LLaVA \cite{liu2023visual})
allow task adaptation via zero-shot learning or fine-tuning. Despite their potential, these models are largely focused on vision-language integration and lack support for broader types of data. \vanessa{In this direction, 
\carlos{
recent models like Sugar \cite{chow2024unified}}
have explored hybrid strategies that combine generative and discriminative training to improve multimodal modeling. By introducing discriminative objectives over generative latent spaces, this approach enhances both retrieval and generation performance. However, it remains tailored to large-scale vision–language models trained on massive datasets and lacks mechanisms to model uncertainty or handle missing information. In addition, the absence of multimodal foundation models capable of operating beyond neuroimaging and genetics further limits their applicability in healthcare contexts.}
\vanessa{Moreover, these models suffer from a major limitation: their lack of interpretability. While,}
post-hoc methods such as LIME \cite{ribeiro2016should} and SHAP \cite{lundberg2017unified} highlight relevant input features and improve on saliency maps; their explanations remain local, fail to generalize across datasets, and often lack the consistency and reliability needed in clinical contexts.


\begin{table}[ht!]
    \centering
    \caption{Comparison of related work models with OSIRIS across three key challenges: task-oriented, handling of missing data, and interpretability.}
    \begin{tabular}{c c c c}
    \toprule
         Algorithm  & Task-oriented & Handling of missing data & Interpretability\\
         \midrule
         Concatenation   & \cmark & \xmark & \cmark \\
         MKL-SVM   & \cmark & \xmark & \warningicon\ (linear) \\
         SR   & \cmark & \xmark & \cmark \\
         Graph   & \xmark & \xmark & \warningicon\ (limited) \\
         M-PLS   & \cmark & \xmark & \cmark \\
         M-FA  & \xmark & \cmark & \cmark \\
         CPM  & \cmark & \cmark & \xmark \\
         TMC  & \cmark & \xmark & \xmark \\
         CSMVIB  & \cmark & \xmark &  \xmark\\
         DeepIMV  & \cmark & \cmark & \xmark \\
         Sugar & \cmark & \cmark & \xmark \\
         FM  & \warningicon\ (fine-tuning) & \xmark & \xmark \\
         OSIRIS & \cmark & \cmark & \cmark \\
         \bottomrule
    \end{tabular}
    \label{tab:advantages_OSIRIS}
\end{table}

\albert{In the missing data imputation techniques realm, several methodological approaches have been proposed. The simplest strategies rely on basic statistics such as mean, median, or mode \cite{zhang2016missing}. While computationally efficient, they fail to capture the underlying complex distribution of the data, leading to biased imputations and loss of variability. More advanced statistical approaches, such as Multivariate Imputation by Chained Equations (MICE) \cite{van2011mice}, iteratively update missing values through conditional models, thereby reducing bias. However, MICE is computationally expensive in high-dimensional settings and sensitive to model misspecification. Thus, ML–based methods have gained prominence by using feature interactions for imputation. Widely used examples include MissForest \cite{stekhoven2012missforest} and k-Nearest Neighbors (kNN) imputer, valued for their accuracy and simplicity. More recently, HyperImputer was proposed \cite{jarrett2022hyperimpute}, which dynamically selects and combines models for improved imputations. Nonetheless, ML-based approaches often struggle with scalability and high proportions of missingness \cite{lin2020missing}. Recently, DL methods have emerged to capture higher-order dependencies. Representative models include Generative Adversarial Imputation Nets (GAIN) \cite{yoon2018gain}, which adapts Generative Adversarial Networks to generate realistic imputations; ReMasker \cite{du2024remasker}, which reconstructs masked entries using contextual neural representations; and DiffPuter \cite{zhang2025diffputer}, which applies diffusion models for iterative refinement. While these methodologies are robust, they require large sample sizes, demand significant computational resources, and pose a risk of overfitting when applied to smaller datasets, a frequent issue in healthcare contexts \cite{sun2023deep}.}

To summarize and highlight the strengths of our proposal, Table \ref{tab:advantages_OSIRIS} provides a comparative overview of the key characteristics of the reviewed approaches with respect to the three main challenges addressed in this work.

\section{Proposed model}
\label{sec:method}

To address the limitations of the state-of-the-art, we introduce \textbf{O}ptimized \textbf{S}parse \textbf{I}nference and \textbf{R}econstruction for \textbf{I}nterdependent \textbf{S}paces (OSIRIS). Our proposal uses a combination of generative and discriminative formulations, enabling learning in multimodal scenarios, with support for semi-supervised training\footnote{The semi-supervised setting refers to the training process where the model have access to all the data (both train and test) except the test labels.} and robust handling of missing data through imputation. These two last aspects are closely related, as the test labels are masked and treated as missing values, allowing the model to work with the entire dataset without accessing the test labels. In addition, OSIRIS identifies two latent representations (or cross-modality factors): one captures discriminative components that facilitate the identification of predictive biomarkers, and the other captures generative components that enables efficient data imputation and enhances our understanding of relationships across different modalities.

Let us consider a multimodal classification problem with $N$ data samples observed across $M$ different modalities or views, denoted by $\{(\mathbf{x}_{n,:}^{\{\mathcal{M}\}}, \mathbf{t}_{n,:})\}_{n=1}^N$, where $\mathcal{M} = \{1, 2, \dots, M\}$. Each $\mathbf{x}_{n,:}^{(m)} \in \mathbb{R}^{D^{(m)}}$ represents the $n$-th observation from the $m$-th modality, and $\mathbf{t}_{n,:} \in \{0,1\}^{C}$ is its one-hot encoded label over $C$ possible classes.  Throughout the paper, we use $\mathbf{I}_{D}$ to denote the $D \times D$ identity matrix. Moreover, we will denote as $\Lambda_\eta$ the diagonal matrix filled by vector $\eta = [\eta_1, \dots, \eta_N]$, i.e., $\Lambda_\eta = \text{diag}(\eta)$; and as $\text{Tr}(\cdot)$ the trace operator. Finally, for variational developments, the mean and covariance of a random variable (r.v.) $\eta$ will be denoted as $\langle\eta\rangle$ and $\Sigma_{\eta}$, respectively.


\subsection{Generative Model}

\begin{figure*}[ht!]
    \begin{subfigure}[b]{0.49\textwidth}
        \centering
        \includegraphics[width=\linewidth]{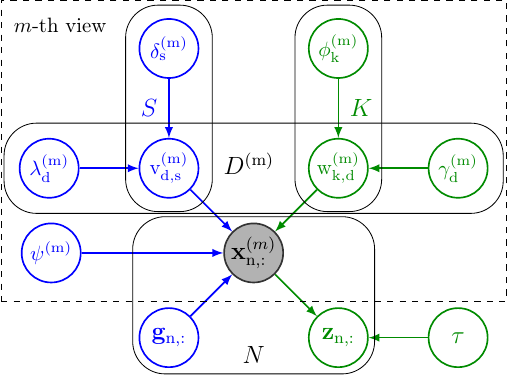}
        \caption{Input view/modality}
        \label{fig:OSIRIS_in}
    \end{subfigure}
    \begin{subfigure}[b]{0.49\textwidth}
        \centering
        \includegraphics[width=\textwidth]{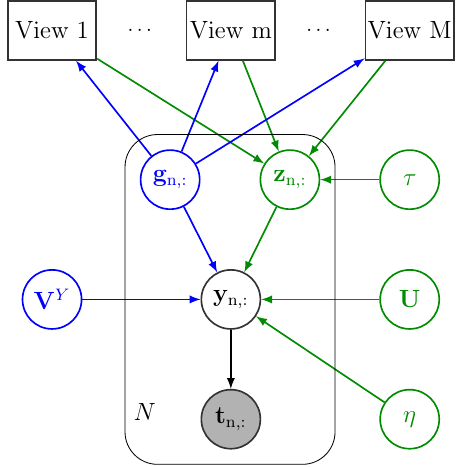}
        \caption{Output view/modality}
        \label{fig:OSIRIS_out}
    \end{subfigure}
    \caption{Diagram of the graphical model of OSIRIS and a zoom into each input view. Grey circles denote observed variables, white circles unobserved variables. Blue font denotes generative variables and green font task-oriented variables. Rectangles represent node groups dependent on the view's feature space.}
    \label{fig:OSIRIS}
\end{figure*}

Our goal is to learn a latent representation that models both inter-view and intra-view structure while capturing the discriminative information for the classification task.
To this end, we introduce OSIRIS, a novel ML model combining generative and discriminative utilities. OSIRIS uses two interdependent latent spaces: a generative latent space, $\G \in \R^{N \times S}$, which captures shared and unique structures across modalities, enabling missing data imputation and semi-supervised learning; and a task-oriented latent space, $\Z \in \R^{N \times K}$, which captures discriminative features optimized for classification. The graphical model in Figure~\ref{fig:OSIRIS_in} illustrates how the different input views contribute to the construction of both latent spaces, which in turn jointly model the output view. \albert{Besides, for further explanations, in order to be consitent with the multimodal FA literature, the term loadings were used for variables that map the latent space to the observed data while weights were used for variables that map from observed data to the latent space.} We now describe in detail the role and structure of each latent space. 


\textbf{The generative latent space} is modeled with a standard Gaussian prior, $\mathbf{g}_{\rm n,:} \sim \mathcal{N}(\mathbf{0}, \mathbf{I}_S)$, and is used to reconstruct each input view as follows:
\begin{equation}
    \mathbf{x}_{\rm n,:}^{(m)} \mid \mathbf{g}_{\rm n,:} \sim \mathcal{N}\left(\mathbf{g}_{\rm n,:} \mathbf{V}^{(m)\top}, \psim^{-1} \mathbf{I}_{D^{(m)}}\right),
    \label{eq:xg}
\end{equation}
where $\mathbf{V}^{(m)}\in \R^{D^{(m)} \times S}$ represents the \carlos{loadings} for modality $m$, and $\psim$ is a view-specific noise precision parameter drawn from a Gamma prior: $\psim \sim \Gamma\left(\alpha^{\psim}, \beta^{\psim}\right)$. This formulation enables the generation of new data samples and imputation of missing observations, while naturally supporting semi-supervised learning using the shared latent structure across all input views. 
To jointly enforce sparsity over both the latent factors and the input features, we place a double Automatic Relevance Determination (ARD) prior on the \carlos{generative loadings}, defined as:
\begin{equation}
    \text{v}_{\rm d,s}^{(m)} \sim \mathcal{N}\left(\mathbf{0}, \left(\lambda_d^{(m)} \delta^{(m)}_s\right)^{-1}\right),
    \label{eq:v}
\end{equation}
where $\delta_s^{(m)} \sim \Gamma\left(\alpha_s^{\delta^{(m)}}, \beta_s^{\delta^{(m)}}\right)$ and $\lambda_d^{(m)} \sim \Gamma\left(\alpha_d^{\lambda^{(m)}}, \beta_d^{\lambda^{(m)}}\right)$ control sparsity across latent dimensions and input features, respectively, following the  ARD principle \citep{neal2012bayesian}.

In contrast, \textbf{the task-oriented latent space} is directly inferred from the observed data as a sparse, linear combination of all views:
\begin{equation}
    \mathbf{z}_{\rm n,:} \sim \N \left(\sum_{m=1}^{M} \mathbf{x}_{\rm n,:}^{(m)} \mathbf{W}^{(m)\top},\tau^{-1}\mathbf{I}_{K} \right),
\end{equation}
where $\mathbf{W}^{(m)} \in \mathbb{R}^{K \times D^{(m)}}$ represents the view-specific \carlos{weights}, and $\tau$ is a shared noise precision parameter drawn from a Gamma prior: $\tau \sim \Gamma\left(\alpha^\tau, \beta^\tau\right)$. 
To impose sparsity across features and latent dimensions, we define a double ARD prior:
\begin{equation} 
    w_{k,d}^{(m)} \sim \mathcal{N}\left(0, \left(\phi_k^{(m)} \gamma_d^{(m)}\right)^{-1}\right),
    \label{eq:wkd}
\end{equation}
where $\phi_k^{(m)} \sim \Gamma\left(\alpha_k^{\phi^{(m)}}, \beta_k^{\phi^{(m)}}\right)$ and $\gamma_d^{(m)} \sim \Gamma\left(\alpha_d^{\gamma^{(m)}}, \beta_d^{\gamma^{(m)}}\right)$ force sparsity over each latent component and feature, respectively. 
Note that, to enhance model interpretability and reduce overfitting, ARD priors are placed over the \carlos{loadings} $\mathbf{V}^{(m)}$ and \carlos{weights} $\mathbf{W}^{(m)}$, enforcing shrinkage on irrelevant latent dimensions and features. Thus, this hierarchical prior structure enables ARD for both reconstruction (via the generative space) and classification (via the task-oriented space).

\textbf{To model the output view}, OSIRIS computes the continuous representation $\mathbf{y}_{\rm n,:}$ of the output view label $\mathbf{t}_{\rm n,:}$ as a linear combination of the discriminative and generative latent variables (see Figure \ref{fig:OSIRIS_out}):
\begin{equation}
    \mathbf{y}_{\rm n,:} = \mathbf{z}_{\rm n,:} \mathbf{U}^\top + \mathbf{g}_{\rm n,:} \mathbf{V}^{\mathbf{Y}\top} + \boldsymbol{\epsilon}_{\mathbf{y}},
    \label{eq_output}
\end{equation}
where $\mathbf{U} \in \mathbb{R}^{C\times K}$ \carlos{represents the discriminative loadings}, whose prior is defined as $\mathbf{u}_{\rm :,k} \sim \mathcal{N}(\mathbf{0}, \mathbf{I}_C)$; $\mathbf{V}^{\mathbf{Y}} \in \mathbb{R}^{C\times S}$ contains the \carlos{generative loadings} of the output view and, therefore, follows Eq. \eqref{eq:v}; and $\boldsymbol{\epsilon}_{\mathbf{y}} \sim \N\left(\textbf{0},\eta^{-1}\Ic\right)$ models the output variability, with $\eta \sim \Gamma\left(\alpha^\eta, \beta^\eta\right)$.  

To complete the generative process, the observed categorical labels  $\mathbf{t}_{\rm n,:}$ are linked to their corresponding continuous representations $\mathbf{y}_{\rm n,:}$ through a Bayesian logistic regression as 
\begin{equation}
    p(\mathbf{t}_{\rm n,:}|\mathbf{y}_{\rm n,:}) = e^{\mathbf{y}_{\rm n,:} \mathbf{t}_{\rm n,:}}\sigma(-\mathbf{y}_{\rm n,:}),
    \label{eq:reg_mod}
\end{equation}
where $\sigma(x) = (1+e^{-x})^{-1}$ is the sigmoid function.


Including both latent spaces in the generation of $\mathbf{y}_{\rm n,:}$ serves two main purposes. First, treating the output as a generative view allows the latent variable $\mathbf{g}_{\rm n,:}$ to capture label-related structure, enabling semi-supervised learning and robust inference even when label data is partially missing, since $\mathbf{g}_{\rm n,:}$ is informed by the observed views. Second, this dual formulation encourages a clean separation of roles: the discriminative latent space $\mathbf{z}_{\rm n,:}$ is optimized to retain only task-relevant information, while $\mathbf{g}_{\rm n,:}$ absorbs complementary or noisy variability. This setup acts similarly to a PCA-like reconstruction term, denoising the label view and improving the discriminative focus of $\mathbf{z}_{\rm n,:}$, ultimately enhancing performance, as confirmed in our experimental results.

\begin{table*}[th!]
\caption{$q^{*}$ update rules for key variables obtained using the mean-field approximation. 
}
    \begin{adjustbox}{max width=\textwidth}
    \renewcommand{\arraystretch}{2}
        \centering
        \setlength{\tabcolsep}{1pt}
        \begin{tabular}{c c c}
        \toprule
        {\bf Variable} & $\mathbf{q^{*}}$ {\bf distribution} & {\bf Parameters}\\
        
        \midrule
        $\mathbf{z}_{\rm n,:}$ & $\N(\mathbf{z}_{\rm n,:}|\langle \mathbf{z}_{\rm n,:}\rangle,\Sigma_{\mathbf{Z}})$ & \begin{tabular}{@{}c@{}}$\Sigma_{\mathbf{Z}}^{-1} = \langle\tau\rangle \Ik + \langle\eta\rangle\langle\mathbf{U}^\top\mathbf{U}\rangle$
        
        \\ $\langle\mathbf{z}_{\rm n,:}\rangle = \left(\langle\tau\rangle\sum_m^M\mathbf{x}_{\rm n,:}^{(m)}\langle\mathbf{W}^{(m)}\rangle^\top  +\langle \eta \rangle \left(\langle\mathbf{y}_{\rm n,:}\rangle - \langle\mathbf{g}_{\rm n,:} \rangle \langle\mathbf{V}^{\mathbf{Y} }  \rangle^\top \right)\langle\mathbf{U}\rangle \right)\Sigma_{\mathbf{Z}}$\end{tabular}\\
        
        \midrule
        $\Wkm$ & $\N(\Wkm|\langle \Wkm\rangle,\Sigma_{\Wk}^{\rm (m)})$ & \begin{tabular}{@{}c@{}}$\Sigma_{\Wk}^{(m)-1} = \tauE \XmT\Xm + \langle\phi_k^{(m)} \rangle \Lambda_{\langle\boldsymbol{\gamma}^{(m)}\rangle}$
        
        \\ $\langle \Wkm\rangle = \tauE\sum_n^N \left[\langle z_{\rm n, k}\rangle - \sum_{m' \neq m}^{M}\langle\mathbf{w}_{\rm k,:}^{(m')}\rangle \mathbf{x}^{(m') \top}_{\rm n,:}\right] \mathbf{x}_{\rm n,:}^{(m)}\Sigma_{\Wk}^{\rm (m)}$\end{tabular}\\

        \midrule
        $\mathbf{g}_{\rm n,:}$ & $\N(\mathbf{g}_{\rm n,:}|\langle \mathbf{g}_{\rm n,:}\rangle,\Sigma_{\mathbf{G}})$ & \begin{tabular}{@{}c@{}}$\Sigma_{\mathbf{G}}^{-1} = \Is + \langle\eta\rangle\langle\mathbf{V}^{\mathbf{Y}\top} \mathbf{V}^{\mathbf{Y}}\rangle + \sum_m^{M}\langle\psi^{(m)}\rangle\langle\mathbf{V}^{(m)\top}\mathbf{V}^{(m)}\rangle$
        
        \\ $\langle\mathbf{g}_{\rm n,:}\rangle = \left( \sum_m^M\left(\langle\psi^{(m)}\rangle\mathbf{x}_{\rm n,:}^{(m)}\langle\mathbf{V}^{(m)}\rangle\right) + \langle\eta\rangle   \left( \langle\mathbf{y}_{n,:}\rangle -\langle\mathbf{z}_{\rm n,:}\rangle\langle\mathbf{U}\rangle^\top \right)\langle\mathbf{V}^{\mathbf{Y}}\rangle \right)\Sigma_{\mathbf{G}}$\end{tabular}\\
        
        \midrule
        $\mathbf{v}_{d ,:}^{(m)}$ & $\N(\mathbf{v}_{d ,:}^{(m)}|\langle \mathbf{v}_{d ,:}^{(m)}\rangle,\Sigma_{\mathbf{v}_{d ,:}}^{\rm (m)})$ & \begin{tabular}{@{}c@{}}$ \Sigma_{\mathbf{v}_{d ,:}}^{\rm (m)-1} = \langle\lambda_{d }^{(m)}\rangle\Lambda_{\langle\boldsymbol{\delta}^{(m)}\rangle} + \langle\psi^{(m)}\rangle\langle\mathbf{G}^\top \mathbf{G}\rangle$
        
        \\ $\langle \mathbf{v}_{d ,:}^{(m)}\rangle = \langle\psi^{(m)}\rangle\mathbf{x}_{\rm :, d }^{(m)\top}\langle\mathbf{G}\rangle\Sigma_{\mathbf{v}_{d ,:}}^{\rm (m)}$\end{tabular}\\

        \midrule
        $\mathbf{y}_{\rm n,:}$ & $\N(\mathbf{y}_{\rm n,:}|\langle \mathbf{y}_{\rm n,:}\rangle,\Sigma_{\mathbf{y}_{\rm n,:}})$ & \begin{tabular}{@{}c@{}c@{}c@{}}$\Sigma_{\mathbf{y}_{\rm n,:}}^{-1} = \langle\eta\rangle\Ic + 2\Lambda_{\boldsymbol{\xi}_{\rm n,:}}$ ~~ where ~~ $\boldsymbol{\xi}_{\rm n,:} = \sqrt{\langle\mathbf{y}_{\rm n,:}\rangle^2 + diag\left(\Sigma_{\mathbf{y}_{\rm n,:}}\right)}$  \\ 
        
        $\langle\mathbf{y}_{\rm n,:}\rangle = \left(\mathbf{t}_{\rm n,:} - \unmed + \langle\eta\rangle\left(\langle\mathbf{z}_{\rm n,:}\rangle\langle\mathbf{U}\rangle^\top + \langle\mathbf{g}_{\rm n,:}\rangle\langle\mathbf{V}^{\mathbf{Y}}\rangle^\top\right)\right)\Sigma_{\mathbf{y}_{\rm n,:}}$ \end{tabular}\\
        \bottomrule

        \end{tabular}
    \end{adjustbox}
    \label{tab:q_dist_main}
\end{table*}

\subsection{Variational Inference}

To perform inference over model variables, included in $\bm{\Theta}$, we approximate the intractable posterior $p(\bm{\Theta} \mid \XM,\Y)$ with a fully factorized mean-field variational distribution $q(\bm{\Theta})$ assumes independence between the model variables as
\begin{equation}
\begin{split}
q\left(\boldsymbol{\Theta}\right) = & \prod_{m=1}^{M} \Bigg(q(\mathbf{V}^{(m)})q(\mathbf{W}^{(m)}) \, q(\psim) \prod_{k}^K \p*{q(\delta_k^{(m)})q(\phi_k^{(m)})} \\
&\prod_{d}^{D^{(m)}} \p*{q(\lambda_d^{(m)})q(\gamma_d^{(m)})}\Bigg) q(\mathbf{V}^{\mathbf{Y}}) q(\mathbf{U}) \prod_{n}^{N} \p*{q(\mathbf{z}_{\rm n,:}) q(\mathbf{g}_{\rm n,:})} \\& q(\eta) q(\tau) q(\mathbf{Y}). \label{eq:qModel}
\end{split}
\end{equation}
where $q(\cdot)$ denotes the approximate posterior distribution of a variable. Furthermore, to evaluate and measure the quality of the approximation $q(\mathbf{\Theta})$, the mean-field method optimizes a variational lower bound $L(q)$, proportional to the Kullback–Leibler (KL) divergence between $q(\mathbf{\Theta})$ and the true posterior $p(\mathbf{\Theta} \mid \mathbf{X}^{\mathcal{M}}, \mathbf{t})$.
Thus, as presented in \cite{bishop2006pattern}, 
analyzing the expression $L(q)$ allows us to derive an approximate posterior distribution for any individual parameter $\theta_j$ in the model, as follows:
\begin{equation} 
    \ln q_j^* = \E_{-q_j}\left[\ln (p(\bm{\Theta}, \mathbf{t}, \mathbf{X}^{\M}))\right] + const, 
\label{eq:lnqopt}
\end{equation} 
where $\E_{-q_j}\left[\cdot\right]$ denotes the expectation with respect to all r.v. except the $j$-th. Furthermore, since Eq.\eqref{eq:lnqopt} requires the involved distributions to be conjugate, we apply a slight modification to Eq.~\eqref{eq:reg_mod}, following the approach introduced in \cite{jaakkola2000bayesian}. Specifically, we derive a lower bound for $p(t_{\rm n,c} \mid y_{\rm n,c})$ using a first-order Taylor expansion, given by:
\begin{equation}
\label{eq:lbb}
\begin{split}
    & p(t_{\rm n,c} = 1|y_{\rm n,c}) = 
    e^{y_{\rm n,c}}\sigma(-y_{\rm n,c}) \geqslant h(y_{\rm n,c}, \xi_{\rm n,c}) \\ &= 
    e^{y_{\rm n,c} t_{\rm n,c}}\sigma(\xi_{\rm n,c})e^{-\frac{y_{\rm n,c} + \xi_{\rm n,c}}{2} - \lambda(\xi_{\rm n,c})(y_{\rm n,c}^2 - \xi_{\rm n,c}^2)},
\end{split}
\end{equation}
where $\lambda(a) = \frac{1}{2a}(\sigma(a) - \tfrac{1}{2})$, and $\xi_{\rm n,c}$ denotes the expansion point of the Taylor series around $y_{\rm n,c}$. It is important to note that $\xi_{\rm n,c}$ is treated as a learnable parameter, optimized during model training via variational inference. This approximation enables us to interpret $p(t_{\rm n,c} \mid y_{\rm n,c})$ as a Gaussian-like distribution, which can then be seamlessly incorporated into Eq.~\eqref{eq:lnqopt}.

\carlos{Furthermore, Table \ref{tab:q_dist_main} includes the variational distributions for the key model variables, the update for all model variables are available in the supplementary material.}
Thus, looking at these Tables, we can observe that, despite $\mathbf{G}$ and $\mathbf{Z}$ having independent priors, their variational posteriors become dependent due to their joint involvement in modeling the output view via Eq. \eqref{eq_output}, which couples both spaces during inference. Notably, $\langle\mathbf{g}_{n,:}\rangle$ reveals that the generative space does not attempt to reconstruct the full $\mathbf{y}_{n,:}$, but rather captures the residual between the output view and the component explained by the discriminative space, i.e., $\langle\mathbf{y}_{n,:}\rangle -\langle\mathbf{z}_{\rm n,:}\rangle\langle\mathbf{U}\rangle^\top$. This residual structure often corresponds to task-irrelevant variability. 
Conversely, the update for $\langle \mathbf{z}_{n,:} \rangle$ depends on the difference between $\mathbf{y}_{n,:}$ and the contribution from $\mathbf{g}_{n,:}$, effectively focusing the discriminative latent space on denoised output information. This encourages $\mathbf{Z}$ to encode features that are truly informative for classification, enhancing task performance.
These results corroborate the motivation discussed earlier for using both latent spaces to model the output view: by separating generative and task-specific information, the model promotes robust and interpretable representations that improve generalization. For the full derivations (including the posterior factorization) of all model variables, see Supplementary Material. The iterative training loop is detailed in Pseudocode \ref{alg:training}.

\setcounter{algorithm}{0}
\begin{algorithm}[!t]
\caption{Training Algorithm of OSIRIS}
\begin{algorithmic}[1]
    \Require Dataset $\{(\X^{\M}, \mathbf{t})\}$, convergence threshold $\epsilon$, maximum iterations $K$
    
    \State \textbf{Initialize} model variables $\bf \Theta = \{\mathbf{Y},\mathbf{Z},\G, \tau, \eta, \mathbf{V}^{Y}, \U, \mathbf{V}^{\M}, \mathbf{W}^{\M}, \psi^{\M}, \boldsymbol{\lambda}^{\M}, \boldsymbol{\delta}^{\M}, \boldsymbol{\phi}^{\M}, \boldsymbol{\gamma}^{\M} \}$
    
    \For{iteration $k = 1$ to $K$}
        \State \textbf{Update} distributions of $\boldsymbol{\Theta}$ (Table \ref{tab:q_dist_main})
        \State \textbf{Compute} $L(q)_k$ (Eq. \ref{eq:Lq})
        \If{$k \geq 100$}
            \State \textbf{Compute} $\Delta =  |L(q)_{k-100} - L(q)_k|$
            \If{$\Delta < \epsilon$}
                \State \textbf{break}  \Comment{Convergence reached}
            \EndIf
        \EndIf
    \EndFor
    
    \State \textbf{return} model parameters
\end{algorithmic}
\label{alg:training}
\end{algorithm}

We can obtain the final $L(q)$ if we introduce both $E_{q}\left[\ln (p(\boldsymbol{\Theta}, \mathbf{t}, \mathbf{X}^{\M}))\right]$ and $\E_{q}\left[\ln (q(\boldsymbol{\Theta}))\right]$ in Eq. \eqref{eq:lbb} and simplify as:
\begin{align}
\label{eq:Lq}
        L(q) = &\frac{N}{2}\ln |\Sigma_{\mathbf{Z}}|- (2 + \frac{N}{2} - \alpha_0^\tau)\ln(\beta^\tau) - (2 + \frac{N}{2} - \alpha_0^\eta)\ln(\beta^\eta)  + \frac{N}{2}\ln|\Sigma_{\mathbf{y}}| + \frac{C}{2}\ln|\Sigma_{\mathbf{U}}| \nonumber\\ &  + \sum_m^M\bigg[\left(\frac{D}{2} + \alpha_0^{\gamma^{\rm (m)}} -2\right)\sum_{\rm d}^{D^{\rm (m)}}\ln(\beta_{\rm d}^{\gamma^{\rm (m)}}) \nonumber\\ & + \left(\frac{K}{2} + \alpha_0^{\phi^{\rm (m)}} -2\right)\sum_{\rm k}^K\ln(\beta_{\rm k}^{\phi^{\rm (m)}}) + \sum_{\rm d}^{D^{\rm (m)}}\left(\beta_0^{\gamma^{\rm (m)}}\frac{\alpha_{\rm d}^{\gamma^{\rm (m)}}}{\beta_{\rm d}^{\gamma^{\rm (m)}}}\right) + \sum_{\rm k}^K\left(\beta_0^{\phi^{\rm (m)}}\frac{\alpha_{\rm k}^{\phi^{\rm (m)}}}{\beta_{\rm k}^{\phi^{\rm (m)}}}\right) \nonumber\\ & + \unmed\sum_{\rm k}^K\sum_{\rm d}^{D^{\rm (m)}}\left(\frac{\alpha_{\rm k}^{\delta^{\rm (m)}} \alpha_{\rm d}^{\gamma^{\rm (m)}}}{\beta_{\rm k}^{\delta^{\rm (m)}} \beta_{\rm d}^{\gamma^{\rm (m)}}}\langle\Wkd\rangle\right) \Bigg]\nonumber\\  & - \sum_n^N\left(\ln(\sigma(\xi_{\rm n,:})) + \langle\mathbf{y}_{\rm n,:}\rangle\mathbf{t}_{\rm n,:}^\top - \unmed(\langle\mathbf{y}_{\rm n,:}\rangle + \xi_{\rm n,:}) - \lambda(\xi_{\rm n,:})(\langle\mathbf{y}_{\rm n,:}^2\rangle - \xi_{\rm n,:}^2)\right)\nonumber\\
        & -\frac{1}{2} \Tr\{\ang{\GT \G}\} - \frac{N}{2} \ln|\Sigma_{\G}| \nonumber\\
        & + \sum_m^M\Bigg[\left(\frac{D^{(m)}}{2} + \alpha_0^{\lambda^{\rm (m)}} -2\right)\sum_{\rm d}^{D^{\rm (m)}}\ln(\beta_{\rm d}^{\lambda^{\rm (m)}}) +\left(\frac{S}{2} + \alpha_0^{\delta^{\rm (m)}} -2\right)\sum_{\rm s}^{S^{\rm (m)}}\ln(\beta_{\rm s}^{\delta^{\rm (m)}})\nonumber\\
        &  - (2 + \frac{N}{2} - \alpha_0^{\psim})\ln(\beta^{\psim})  -\left( \frac{D}{2}\ln\left|\Sigma_{\Vm}\right| \right)\Bigg]
\end{align}
where $\vert\cdot\vert$ denotes the determinant operator. For further mathematical developments of this expression, see the Supplementary Material.

The code of the algorithm and the means to reproduce the experiments of this work are publicly available \href{https://github.com/albello1/OSIRIS.git}{here}.

\subsection{Posterior predictive distribution}
\label{sec:pred}

Since the model incorporates two intermediate stages between the input data $\mathbf{X}^\M$ and the final classification determined by $\mathbf{t}$, predictions are performed in a hierarchical manner. Thus, to calculate the posterior predictive distribution, we begin marginalizing the posterior distribution of $\mathbf{t}$ over $\mathbf{y}$ for a new data point $\xpredin$ as
\begin{equation}
    p(\tpredc = 1|\xpredin) = \int_{\ypredc} p(\tpredc = 1|\ypredc)p(\ypredc|\xpredin)d\ypredc,
    \label{eq:dist_clas}
\end{equation}
where $\ypredc$ is the $c$-th column of the regression output of the model. 

The challenge we face is that Eq. \eqref{eq:dist_clas} does not have an analytical or closed-form solution. Thus, we will perform the approximation presented in \cite{bishop2006pattern}, which let us rewrite Eq. \eqref{eq:dist_clas} as
\begin{equation}
\label{eq:final_pred}
    p(\tpredc = 1|\xpredin) \simeq \sigma\left(\frac{\langle \ypredc\rangle }{(1 + \frac{\pi}{8}\Sigma_{\ypredc})^{\unmed}}\right).
\end{equation}
where $\Phi(\cdot)$ is the probit function.

Now, to infer the parameters $\langle \ypredc\rangle$ and $\Sigma_{\ypredc}$, it is necessary to parameterize $p(\ypredc|\xpredin)$. This involves marginalizing over the main model variables as
\begin{equation}
\label{eq:y_arr}
    p(\ypredc|\xpredin) = \int p(\ypredc|\boldsymbol{\Theta}) p\p*{\boldsymbol{\Theta}|\xpredin} d\boldsymbol{\Theta},
\end{equation}
where $\boldsymbol{\Theta} = \{\Z, \G, \U, \mathbf{V}^{\Y}, \eta\}$. Having developed mean-field inference over the model variables, we can rewrite Eq. \eqref{eq:y_arr} employing the obtained approximate posterior distributions as follows:
\begin{equation}
\begin{split}
\label{eq:yyy}
    p(\ypredc|\xpredin) = & \int_{\mathbf{z}_{\rm *, :}}\int_{\mathbf{g}_{\rm *,:}}\int_{\mathbf{u}_{\rm c,:}}\int_{\mathbf{v}_{\rm c,:}^{\Y}}\int_\eta N(\ypredc|\mathbf{z}_{\rm *,:} \mathbf{u}_{\rm c,:}^\top + \mathbf{g}_{\rm *,:} \mathbf{v}_{\rm c,:}^{\mathbf{Y}\top},\eta^{-1})q(\mathbf{z}_{\rm *, :})q(\mathbf{g}_{\rm *,:}) \\
    & q(\mathbf{u}_{\rm c,:})q(\mathbf{v}_{\rm c,:}^{\Y})q(\eta) d\mathbf{z}_{\rm *,:} d\mathbf{g}_{\rm *,:} d\mathbf{u}_{\rm c,:} d\mathbf{v}_{\rm c,:}^{\mathbf{Y}\top} d\eta.
\end{split}
\end{equation}

Thus, operating over Eq. \eqref{eq:yyy}, we can calculate analitically the mean as
\begin{equation}
\label{eq:mean_yy}
    \E[\ypredc] = \E[\mathbf{z}_{\rm *,:} \mathbf{u}_{c,:}^\top + \mathbf{g}_{\rm *,:} \mathbf{v}_{c,:}^{\mathbf{Y}\top} + \epsilon_{Y}] = \langle\mathbf{z}_{\rm *,:}\rangle\langle\mathbf{u}_{c,:}^\top\rangle + \langle\mathbf{g}_{\rm *,:}\rangle\langle\mathbf{v}_{c,:}^{\mathbf{Y}\top}\rangle,
\end{equation}
and the variance 
\begin{equation}
\begin{split}
\label{eq:var_yy}
    Var[\ypredc] & = \langle\eta\rangle + \sum_k^K\Big(\Sigma_{\text{z}_{\rm *,k}}\Sigma_{\text{u}_{\rm c,k}} + \langle\text{z}_{\rm *,k}\rangle\langle\text{z}_{\rm *,k}\rangle\Sigma_{\text{u}_{\rm c,k}} + \langle \text{u}_{\rm c,k}\rangle\langle\text{u}_{\rm c,k} \rangle\Sigma_{\mathbf{z}_{\rm *,k}}\Big) \\
    & + \sum_{s}^{S}\Big(\Sigma_{\mathbf{g}_{\rm *,s}}\Sigma_{\mathbf{v}_{\rm c,s}^{Y}} + \langle\mathbf{g}_{\rm *,s}\rangle\langle\mathbf{g}_{\rm *,s}\rangle\Sigma_{\mathbf{v}^{Y}_{\rm c,s}} + \langle \mathbf{
v}^{Y}_{\rm c,s}\rangle\langle\mathbf{v}^{Y}_{\rm c,s} \rangle\Sigma_{\mathbf{g}_{\rm *,s}}\Big).
\end{split}
\end{equation}

In that way, Eq. \eqref{eq:mean_yy} can be used for the hard prediction\footnote{As a Gaussian distribution, the expectation corresponds to the mean, mode, and median, which are the same. Thus, the expectation could be used a trustworthy predictor.} while Eq. \eqref{eq:var_yy} to estimate the uncertainity of the prediction. As explained at Section \ref{sec:intro}, this uncertainity estimation is crucial when working in biomedical settings.


\subsection{Missing data imputation}

To address missing values and enable semi-supervised training, we treat the missing entries as additional random variables to be inferred. Let $U$ denote the set of unobserved features and $O$ the set of observed ones. For a given data point in modality $m$, we write $\mathbf{x}_{*,U}^{(m)}$ for the unobserved features and $\mathbf{x}_{*,O}^{(m)}$ for the observed ones. The posterior distribution of the missing entries can then be expressed as
\begin{equation}
    p(\mathbf{x}_{*,U}^{(m)} \mid \mathbf{x}_{*,O}^{(m)}, \boldsymbol{\Theta})
    = \prod_{m=1}^{\M} \int p(\mathbf{x}^{(m)}_{*,U} \mid \mathbf{g}_{*,:}, \boldsymbol{\Theta})\, 
    p(\mathbf{g}_{*,:} \mid \mathbf{x}_{*,O}^{\M}, \boldsymbol{\Theta}) \, d\boldsymbol{\Theta}.
\end{equation}

The posterior mean and variance of the unobserved features are given by
\begin{equation}
    \langle\mathbf{x}_{*,U}^{(m)}\rangle = 
    \langle\mathbf{g}_{*,:}\rangle \langle\mathbf{V}_{U,:}^{(m)\top}\rangle,
\end{equation}
and
\begin{equation}
    \Sigma_{\mathbf{x}_{*,U}^{(m)}} = 
    \langle\psi^{(m)}\rangle^{-1} 
    + \Sigma_{\mathbf{g}_{*,:}}\Sigma_{\mathbf{V}_{U,:}^{(m)}} 
    + \langle\mathbf{g}_{*,:}\mathbf{g}_{*,:}^\top\rangle \Sigma_{\mathbf{V}_{U,:}^{(m)}} 
    + \langle\mathbf{v}_{u,:}^{(m)}\mathbf{v}_{u,:}^{(m)\top}\rangle \Sigma_{\mathbf{g}_{*,:}}.
\end{equation}

Finally, the marginal posterior distribution of $\mathbf{g}_{*,:}$ is computed as
\begin{equation}
    \Sigma_{\mathbf{G}}^{-1} = \Is
    + \langle\eta\rangle \langle\mathbf{V}^{\mathbf{Y}\top} \mathbf{V}^{\mathbf{Y}}\rangle 
    + \sum_{m=1}^{\M} \langle\psi^{(m)}\rangle 
      \langle\mathbf{V}_{U,:}^{(m)\top}\mathbf{V}_{U,:}^{(m)}\rangle,
\end{equation}
with mean
\begin{equation}
    \langle\mathbf{g}_{*,:}\rangle =
    \Bigg( 
        \sum_{m=1}^{\M} 
            \langle\psi^{(m)}\rangle 
            \langle\mathbf{x}_{*,U}^{(m)}\rangle 
            \langle\mathbf{V}_{U,:}^{(m)}\rangle 
        + \langle\eta\rangle 
            \big( \langle\mathbf{y}_{*,:}\rangle 
            - \langle\mathbf{z}_{*,:}\rangle \langle\mathbf{U}\rangle^\top \big) 
            \langle\mathbf{V}^{\mathbf{Y}}\rangle 
    \Bigg) \Sigma_{\mathbf{G}}.
\end{equation}

By treating the missing entries as r.v. inferred within the unsupervised iterative process, we avoid potential biases associated with specific classes. The algorithm iteratively parameterizes the distributions of the missing entries to optimize the generative space $\mathbf{G}$. 

A detailed explanation and further mathematical developments of the posterior predictive distribution and the missing-value imputation are available in the Supplementary Material.

\section{Experiments}
\label{sec:exp}

\albert{In this section, we analyze the capabilities and benefits of OSIRIS in an empirical way. First, we evaluate the model's performance over several clinical and non clinical datasets, both in abscense and precense of missing values. To do so, we compare the model against state-of-the-art baselines, both multi-modal classification models and missing data imputation algorithms. Second, we tested the model over a real clinical problem to assess its capacity to uncover interpretable discriminative and generative factors capturing associations across different modalities.}



\subsection{Classification Performance}

\begin{table}[th!]
    \centering
    \caption{Multimodal datasets used in this study. For each dataset (left to right), we report the number of classes, number of samples, modality dimensions (in parentheses), and a description of the input features.
    }
    \begin{adjustbox}{max width = \linewidth}
    \begin{tabular}{c c c c c}
    \toprule
         Database  & classes & Samples & Features & Description  \\
         \midrule
          Arrhythmia   & 2 & 452 &   (15;264) &  ECG features  \\
          Digit   & 10 & 2000 &  (216;76;64;6;240;47)  & Image features \\
          Sat  & 6 & 6435 & (4;4)   & Image features \\
          LFWA  & 7 & 1277 &  (2400;73)  & Flattened images + discrete features \\
          Fashion  & 4 & 2725 &  (512;512)  & Embeddings from CLIP \\
          Tadpole  & 3 & 1296 & (5;4;5;3;7;7) & Tabular medical data \\
          X-Ray  & 10 & 800 & (273;112)   & Image features \\
          Scene15  & 15 & 4485 & (8;90;10)  & Image features  \\
          ADNI  & 2 & 1082 & (114;9;13;152;4) & Brain regions\\
         \bottomrule
    \end{tabular}
    \end{adjustbox}
    \label{tab:databases}
\end{table}

To evaluate the effectiveness of our approach, we conduct a comprehensive comparison against state-of-the-art methods, including both classical ML techniques and recent DL models. Specifically, we benchmark OSIRIS against MKL-SVM with RBF kernel \cite{zhang2024explaining}, multimodal PLS (M-PLS) \cite{belenguer2024unified}, multimodal FA (M-FA) \cite{sevilla2022bayesian}, and four deep multimodal baselines: CPM \cite{zhang2019cpm}, TMC \cite{han2022trusted}, CSMVIB \cite{zhang2025towards}, and DeepIMV \cite{lee2021variational}. Regarding the hyperparameter selection criterion, for the Bayesian models (M-PLS, M-FA), no hyperparameter has been cross-validated since they are inferred iteratively; for DL models, we used author-recommended ranges; and for SVM the regularization parameter was cross-validated from $10^{-3}$ to $10^{3}$ in log-scale steps.

Table~\ref{tab:databases} summarizes the main characteristics of the datasets used in the comparison. Since the TADPOLE dataset contains substantial missing data, we applied median imputation to fill in missing values for the methods that cannot handle them (MKL-SVM, M-PLS, TMC and CSMVIB).
M-FA and OSIRIS’s generative latent space were both initialized with 100 latent dimensions, while M-PLS and OSIRIS’s discriminative latent space were set to a dimensionality equal to number of output classes minus one\footnote{As in Bayesian Partial Least Squares (BPLS), the number of latent components $K$ is bounded to $C - 1$ for classification with $C$ classes.
}. In the case of OSIRIS, the sparsity-inducing priors automatically pruned irrelevant dimensions, converging to a compact representation. In all Bayesian models, OSIRIS, M-PLS and M-FA, convergence was determined by the criterion $L(q)_{T-100} > L(q)_{T}(1 - 10^{-8})$, where $L(q)_T$ denotes the variational lower bound at iteration $T$.
We report results over 10 random train/test splits, evaluating performance using AUC and balanced accuracy (BACC), defined as $BACC = \frac{\text{sensitivity} + \text{specificity}}{2}$.

\begin{table}[!th]
\centering
\caption{Classification performance of OSIRIS (rightmost column) compared to all baseline methods. For each dataset and model, we report the AUC (white cells) and BACC (light gray cells).}
\begin{adjustbox}{max width=\textwidth}
    \begin{tabular}{ c c c c c c c c c}
    \toprule
    ~& MKL-SVM &  M-PLS & M-FA & CPM & TMC & CSMVIB & DeepIMV & OSIRIS\\ \midrule
    
    \multirow{2}{*}{Arrhythmia}     
                & 0.80 $\pm$ 0.06 & 0.80 $\pm$ 0.03 & 0.78 $\pm$ 0.07 & 0.71 $\pm$ 0.04 & 0.71 $\pm$ 0.09 & 0.56 $\pm$ 0.04 & 0.61 $\pm$ 0.09 & \textbf{0.82} $\pm$ \textbf{0.04} \\
                 &\cellcolor{gray!10} 0.72 $\pm$ 0.07  & \cellcolor{gray!10} 0.74 $\pm$ 0.06 & \cellcolor{gray!10} 0.72    $\pm$ 0.06 & \cellcolor{gray!10} 0.70 $\pm$ 0.08 &\cellcolor{gray!10} 0.68   $\pm$ 0.04 &\cellcolor{gray!10} 0.53  $\pm$ 0.04 &\cellcolor{gray!10} 0.58  $\pm$ 0.07 &\cellcolor{gray!10} \textbf{0.80}  $\pm$ \textbf{0.06}  \\     
    \midrule
    \multirow{2}{*}{Digit}     
                 &  0.98 $\pm$ 0.01 & 0.98 $\pm$ 0.01 & 0.88 $\pm$ 0.02  & \textbf{0.99} $\pm$ \textbf{0.01} & \textbf{0.99} $\pm$ \textbf{0.01} & 0.89 $\pm$ 0.01 & 0.97 $\pm$ 0.01 & \textbf{0.99} $\pm$ \textbf{0.01} \\
                 &\cellcolor{gray!10} 0.81 $\pm$ 0.02  & \cellcolor{gray!10} 0.89 $\pm$ 0.01  & \cellcolor{gray!10} 0.55   $\pm$ 0.07  & \cellcolor{gray!10} 0.92 $\pm$ 0.04 &\cellcolor{gray!10} \textbf{0.98}  $\pm$ \textbf{0.01} &\cellcolor{gray!10}  0.51 $\pm$ 0.01 &\cellcolor{gray!10} 0.70   $\pm$ 0.07 &\cellcolor{gray!10} \textbf{0.98}  $\pm$ \textbf{0.01} \\     
    \midrule
    \multirow{2}{*}{Sat}     
                  & \textbf{0.91} $\pm$ \textbf{0.01}  & 0.88 $\pm$ 0.01  & \textbf{0.91} $\pm$ \textbf{0.01} & \textbf{0.91} $\pm$ \textbf{0.02} & \textbf{0.91} $\pm$ \textbf{0.01} & 0.75 $\pm$ 0.04 &  \textbf{0.91} $\pm$ \textbf{0.01} & \textbf{0.91} $\pm$ \textbf{0.01}\\
                 &\cellcolor{gray!10} \textbf{0.80} $\pm$ \textbf{0.01}  & \cellcolor{gray!10} 0.53 $\pm$ 0.01  & \cellcolor{gray!10}  0.64  $\pm$ 0.01  & \cellcolor{gray!10} 0.77 $\pm$ 0.02  &\cellcolor{gray!10} 0.72  $\pm$ 0.01 &\cellcolor{gray!10} 0.35  $\pm$ 0.07 &\cellcolor{gray!10}  0.59 $\pm$ 0.01 &\cellcolor{gray!10}  0.73 $\pm$ 0.01 \\     
    \midrule
    \multirow{2}{*}{LFWA}     
                  & 0.90 $\pm$ 0.01 & 0.84 $\pm$ 0.01  & 0.97 $\pm$ 0.01  & \textbf{0.99} $\pm$ \textbf{0.01} & 0.92 $\pm$ 0.01  & 0.81 $\pm$ 0.02 & 0.91 $\pm$ 0.01 & \textbf{0.99} $\pm$ \textbf{0.01} \\
                  &\cellcolor{gray!10} 0.38 $\pm$ 0.01  & \cellcolor{gray!10} 0.52 $\pm$ 0.05 & \cellcolor{gray!10}    0.83 $\pm$ 0.03  & \cellcolor{gray!10} \textbf{0.92} $\pm$ \textbf{0.02}  &\cellcolor{gray!10}   0.73 $\pm$ 0.03 &\cellcolor{gray!10} 0.51  $\pm$ 0.02 &\cellcolor{gray!10}  0.31 $\pm$ 0.03 &\cellcolor{gray!10} \textbf{0.92}  $\pm$ \textbf{0.02} \\     
    \midrule
    \multirow{2}{*}{Fashion}     
                 & 0.72 $\pm$ 0.09 & \textbf{0.99} $\pm$ \textbf{0.01}  & \textbf{0.99} $\pm$ \textbf{0.01}  & \textbf{0.99} $\pm$ \textbf{0.01} & 0.98 $\pm$ 0.03 & 0.55 $\pm$ 0.02 & \textbf{0.99} $\pm$ 0.01 & \textbf{0.99} $\pm$ \textbf{0.01} \\
                  &\cellcolor{gray!10} 0.25  $\pm$ 0.01  & \cellcolor{gray!10} \textbf{0.99} $\pm$ \textbf{0.01} & \cellcolor{gray!10}  \textbf{0.99}  $\pm$ 0.01  & \cellcolor{gray!10} \textbf{0.99}$\pm$ \textbf{0.01} &\cellcolor{gray!10}   0.97 $\pm$ 0.07 &\cellcolor{gray!10}  0.27 $\pm$ 0.01 &\cellcolor{gray!10} \textbf{0.99}  $\pm$ \textbf{0.01} &\cellcolor{gray!10} \textbf{0.99} $\pm$ \textbf{0.01} \\     
    \midrule
    \multirow{2}{*}{Tadpole}     
                  & 0.83 $\pm$ 0.01 & 0.78 $\pm$ 0.01  & 0.82 $\pm$ 0.02  & \textbf{0.84} $\pm$ \textbf{0.02} & 0.82 $\pm$ 0.02 & 0.71 $\pm$ 0.02 &  0.73 $\pm$ 0.04 & 0.83 $\pm$ 0.01\\
                  &\cellcolor{gray!10} 0.73 $\pm$ 0.04  & \cellcolor{gray!10} 0.64 $\pm$ 0.02  & \cellcolor{gray!10} 0.66   $\pm$  0.05  & \cellcolor{gray!10} \textbf{0.76} $\pm$ \textbf{0.03} &\cellcolor{gray!10}  0.74 $\pm$ 0.03 &\cellcolor{gray!10}  0.56 $\pm$ 0.03 &\cellcolor{gray!10}  0.35 $\pm$ 0.04 &\cellcolor{gray!10}  \textbf{0.76} $\pm$ \textbf{0.02} \\ 
    \midrule
    \multirow{2}{*}{X-Ray}     
                  & \textbf{0.99} $\pm$ \textbf{0.01}  & 0.93 $\pm$ 0.01  & 0.96 $\pm$ 0.01  & \textbf{0.99} $\pm$ \textbf{0.01} & 0.98 $\pm$ 0.01 & 0.58 $\pm$ 0.03  & 0.97  $\pm$ 0.01 & \textbf{0.99} $\pm$ \textbf{0.01} \\
                  &\cellcolor{gray!10} 0.87 $\pm$ 0.04  & \cellcolor{gray!10} 0.46 $\pm$ 0.05  & \cellcolor{gray!10} 0.64  $\pm$ 0.05   & \cellcolor{gray!10} 0.88 $\pm$ 0.02&\cellcolor{gray!10}  0.87 $\pm$ 0.04  &\cellcolor{gray!10} 0.12  $\pm$ 0.02  &\cellcolor{gray!10}  0.46 $\pm$ 0.05 &\cellcolor{gray!10}  \textbf{0.92} $\pm$ \textbf{0.05} \\  
    \midrule
    \multirow{2}{*}{Scene15}     
                  & \textbf{0.91} $\pm$ \textbf{0.01}  & 0.81 $\pm$ 0.01 & 0.87 $\pm$ 0.01   & 0.88 $\pm$ 0.02 &  0.88 $\pm$ 0.01  & 0.74 $\pm$ 0.01   &  0.89 $\pm$ 0.01 & \textbf{0.91} $\pm$ \textbf{0.01} \\
                  &\cellcolor{gray!10} 0.57 $\pm$ 0.01  & \cellcolor{gray!10}0.17  $\pm$ 0.01   & \cellcolor{gray!10} 0.35  $\pm$ 0.04   & \cellcolor{gray!10} 0.52 $\pm$ 0.01&\cellcolor{gray!10}  0.40 $\pm$ 0.05  &\cellcolor{gray!10} 0.19  $\pm$ 0.01   &\cellcolor{gray!10}  0.37 $\pm$ 0.03  &\cellcolor{gray!10}  \textbf{0.60} $\pm$ \textbf{0.02} \\ 
    \midrule
    \multirow{2}{*}{ADNI}     
                  & 0.68 $\pm$ 0.04 & 0.72  $\pm$ 0.05 & \textbf{0.81} $\pm$ \textbf{0.04}   & \textbf{0.81} $\pm$ \textbf{0.05} & \textbf{0.81} $\pm$ \textbf{0.04}  & 0.52 $\pm$   0.01 &  0.75 $\pm$ 0.03 & \textbf{0.81} $\pm$ \textbf{0.04}\\
                  &\cellcolor{gray!10} 0.59 $\pm$ 0.04  & \cellcolor{gray!10} 0.70 $\pm$ 0.05   & \cellcolor{gray!10}  0.67 $\pm$  0.04  & \cellcolor{gray!10} 0.67 $\pm$ 0.02 &\cellcolor{gray!10} 0.34  $\pm$ 0.05  &\cellcolor{gray!10}  0.50 $\pm$    0.0&\cellcolor{gray!10} 0.60  $\pm$ 0.04  &\cellcolor{gray!10} \textbf{0.80} $\pm$  \textbf{0.04}\\ 
    \midrule
    \end{tabular}
    \end{adjustbox}
    \label{tab:results_standard}
\end{table}

\albert{

Table~\ref{tab:results_standard} shows that OSIRIS outperforms or matches all baselines in terms of AUC and BACC on every dataset except Sat, where OSIRIS obtained the same AUC as the best baselines. Its strength is particularly evident in multiclass settings, where it maintains competitive AUC scores while substantially outperforming other models in BACC. This advantage is most striking in the Arrhythmia dataset, where OSIRIS surpasses the rest of algorithms by a robust margin, highlighting its ability to effectively handle challenging classification tasks.

When comparing OSIRIS to individual baselines, the CPM model achieves the most similar results. This observation is consistent with prior reports in the literature, where CPM is often positioned among the strongest multimodal state-of-the-art methods. In contrast, other baselines such as DeepIMV, M-FA, and CSMVIB show less consistent behavior: while they reach comparable performance in some datasets, they often underperform significantly on others. For instance, DeepIMV achieves notably lower AUC scores than OSIRIS on Arrhythmia (0.61 vs. 0.82), Tadpole (0.73 vs. 0.83), and ADNI (0.75 vs. 0.81), underscoring the greater robustness of our approach. Furthermore, among traditional ML baselines, MKL-SVM stands out as the strongest competitor. This aligns with the broader literature, where MKL-SVM is frequently highlighted for its simplicity and robustness. Nevertheless, despite its competitiveness, MKL-SVM remains underperforming OSIRIS. We attribute this gap in performance to OSIRIS’s ability to compactly summarize multimodal information through a small number of latent factors, typically between 10 and 30 on average.

}



\albert{Moreover, }compared to M-FA and M-PLS, OSIRIS provides distinct advantages. While M-FA focuses on reconstructing all input variance, including task-irrelevant information, OSIRIS separates generative and discriminative factors, enabling the discriminative space to focus solely on task-relevant features. Additionally, OSIRIS outperforms M-PLS by using its generative capability for semi-supervised learning. This approach also results in a cleaner discriminative space, free from noise and other irrelevant factors modeled by the generative space, leading to a significant performance improvement.

\subsection{Missing data imputation}

To assess the advantages of OSIRIS's integrated imputation strategy, we analyzed how BACC evolves under varying rates of missing values (from 5$\%$ to 30$\%$). We compared standard OSIRIS (where data imputation is carried out by its generative model) with \albert{six} external imputation methods applied as a preprocessing step prior to OSIRIS: mean, K-Nearest Neighbors (KNN), 
missForest \cite{stekhoven2012missforest},
Multivariate Imputation by Chained Equations (MICE) \cite{van2011mice}, Generative Adversarial Imputation Nets (GAIN) \cite{yoon2018gain}, and HyperImpute \cite{jarrett2022hyperimpute}. To mitigate label bias, we generated ten random missing-data masks for each dataset and missing-rate combination, then averaged the resulting performance. In addition, the Variational AutoEncoder (VAE) imputer and ReMasker \cite{du2024remasker} were preliminarily evaluated; however, due to their significantly inferior performance compared to the other inputers, they were not incorporated in this analysis.

\begin{figure}[!th]
    \centering
    \begin{subfigure}{0.32\textwidth}
        \includegraphics[trim={0.5cm 0 2.5cm 0},clip,width=\linewidth]{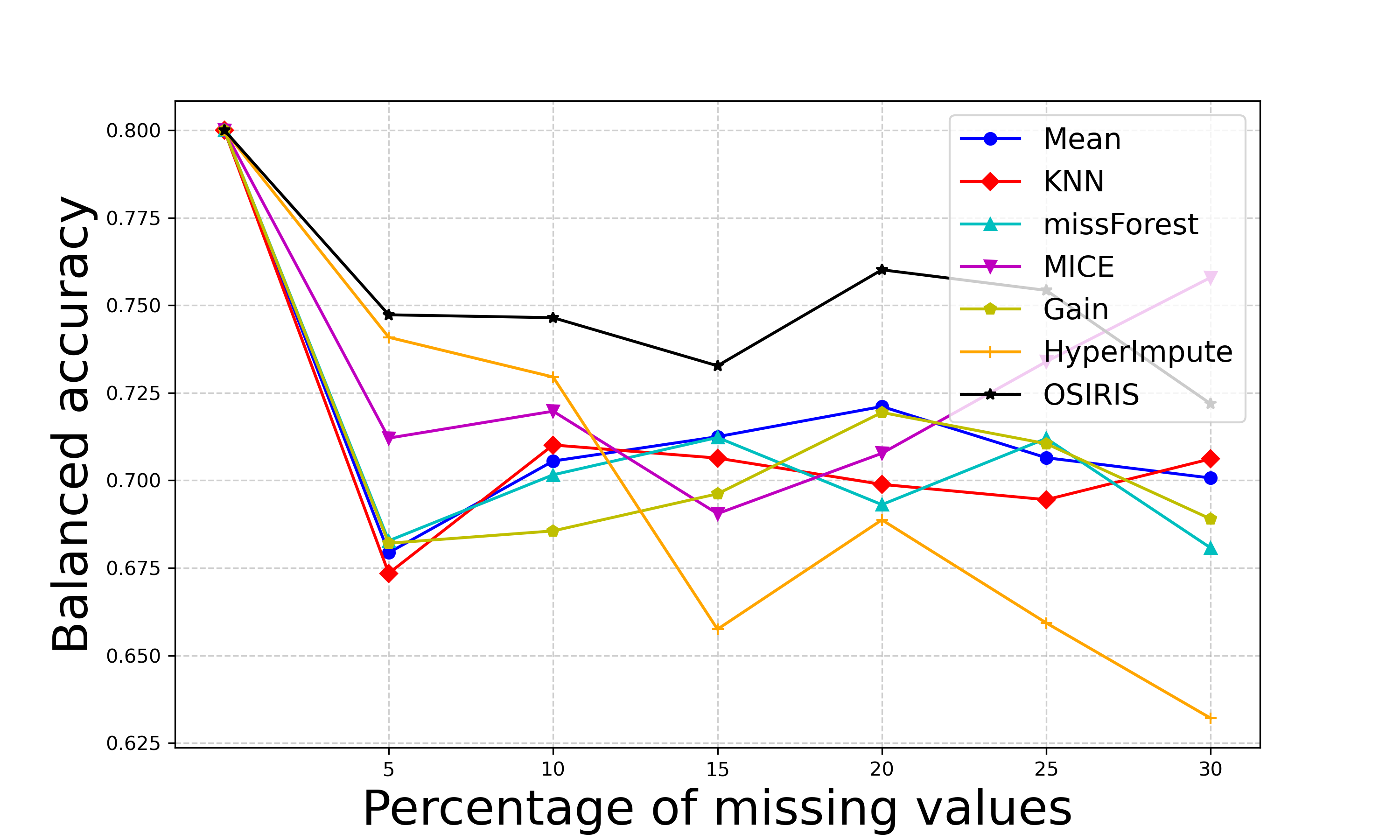}
        \caption{Arrhythmia}
    \end{subfigure}
    \hfill
    \begin{subfigure}{0.32\textwidth}
        \includegraphics[trim={0.5cm 0 2.5cm 0},clip,width=\linewidth]{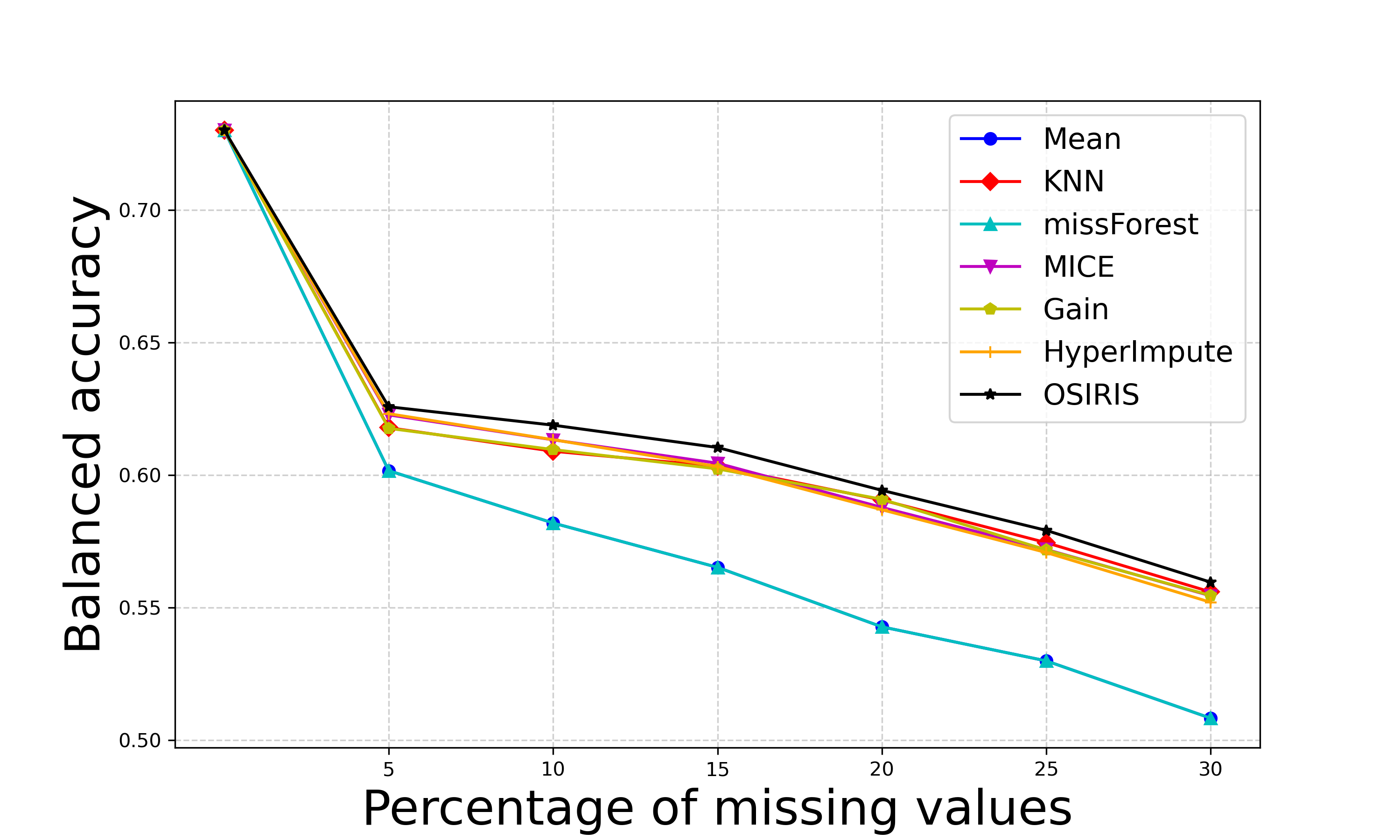}
        \caption{Sat}
    \end{subfigure}
    \hfill
    \begin{subfigure}{0.32\textwidth}
        \includegraphics[trim={0.5cm 0 2.5cm 0},clip,width=\linewidth]{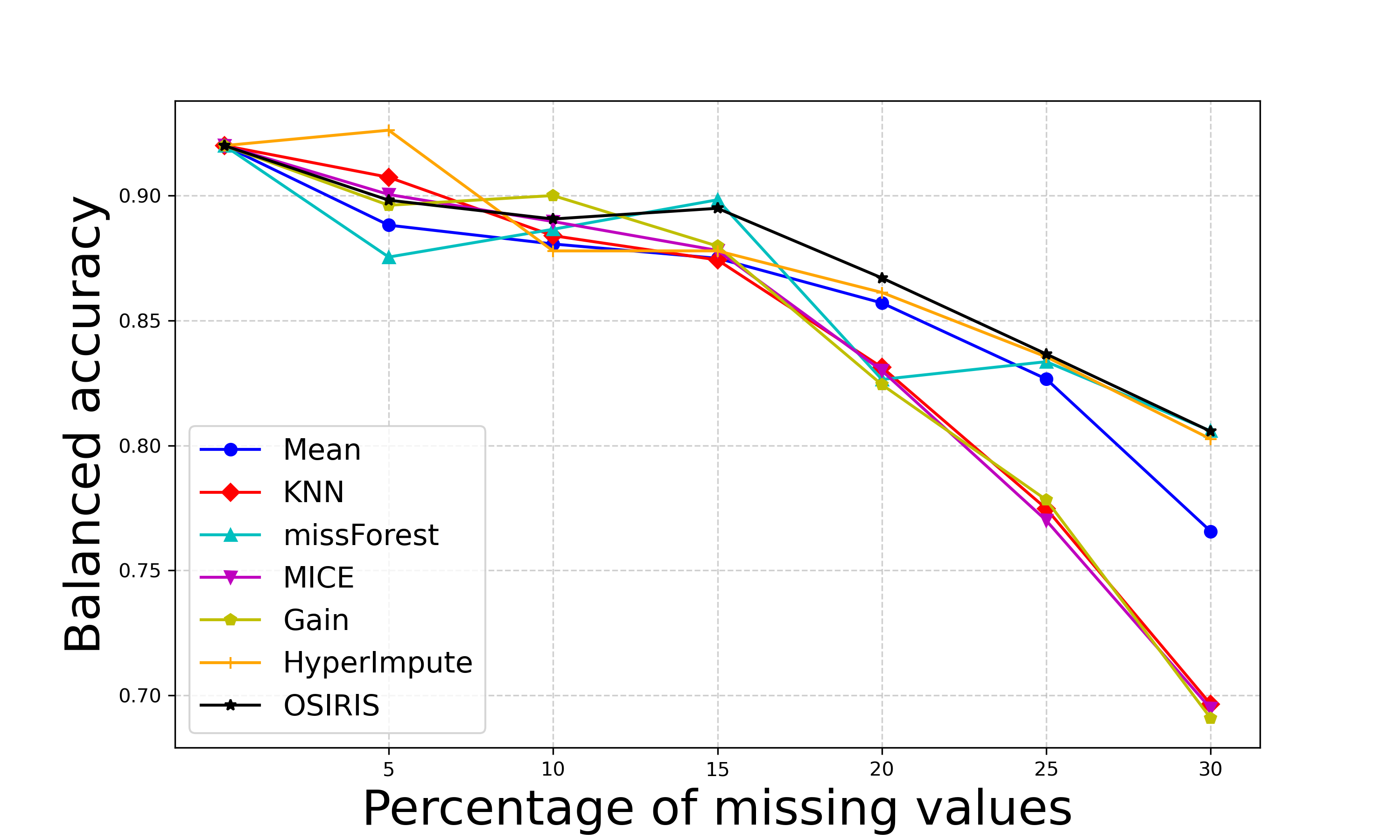}
        \caption{LFWA}
    \end{subfigure}

    \vspace{1em}

    \begin{subfigure}{0.32\textwidth}
        \includegraphics[trim={0.5cm 0 2.5cm 0},clip,width=\linewidth]{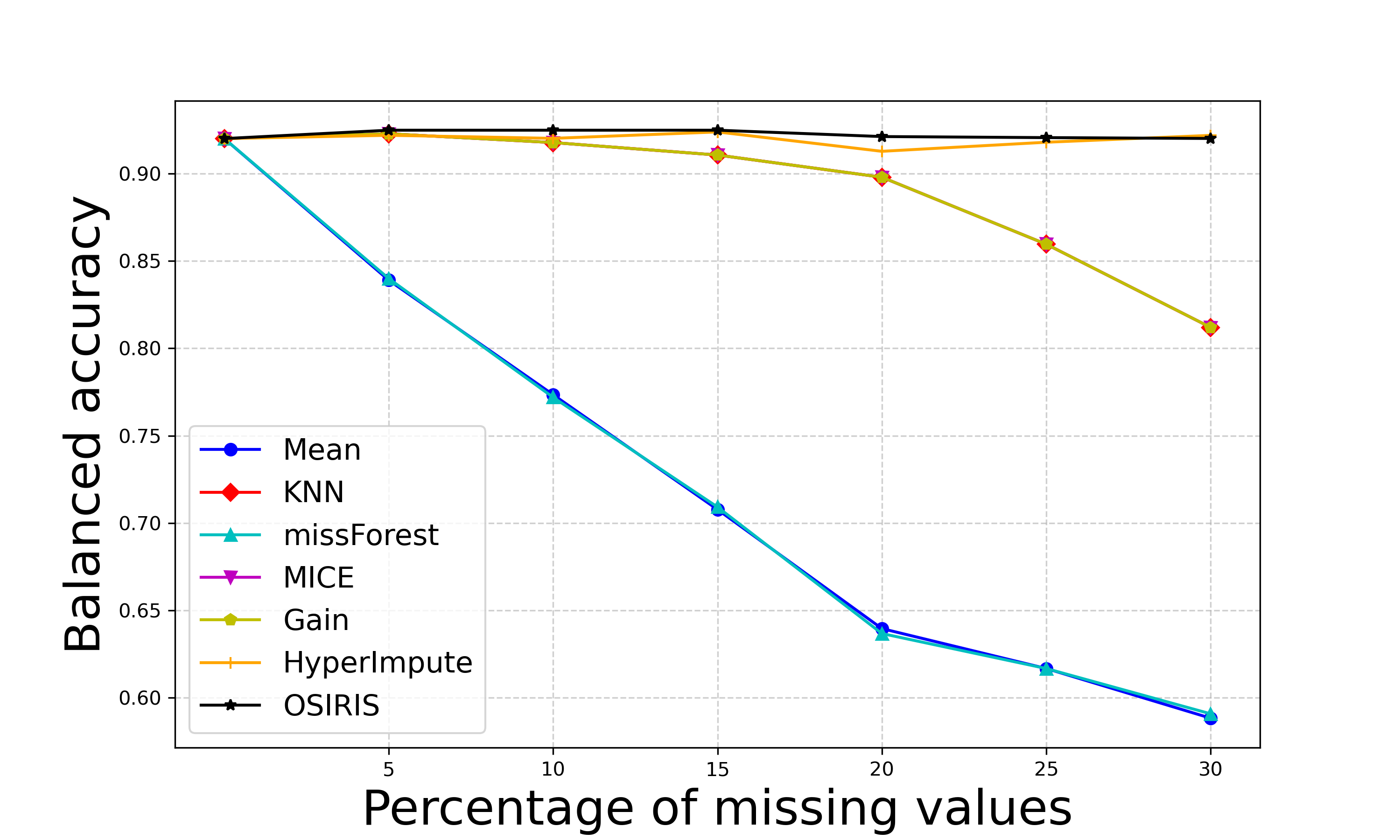}
        \caption{X-Ray}
    \end{subfigure}
    \hfill
    \begin{subfigure}{0.32\textwidth}
        \includegraphics[trim={0.5cm 0 2.5cm 0},clip,width=\linewidth]{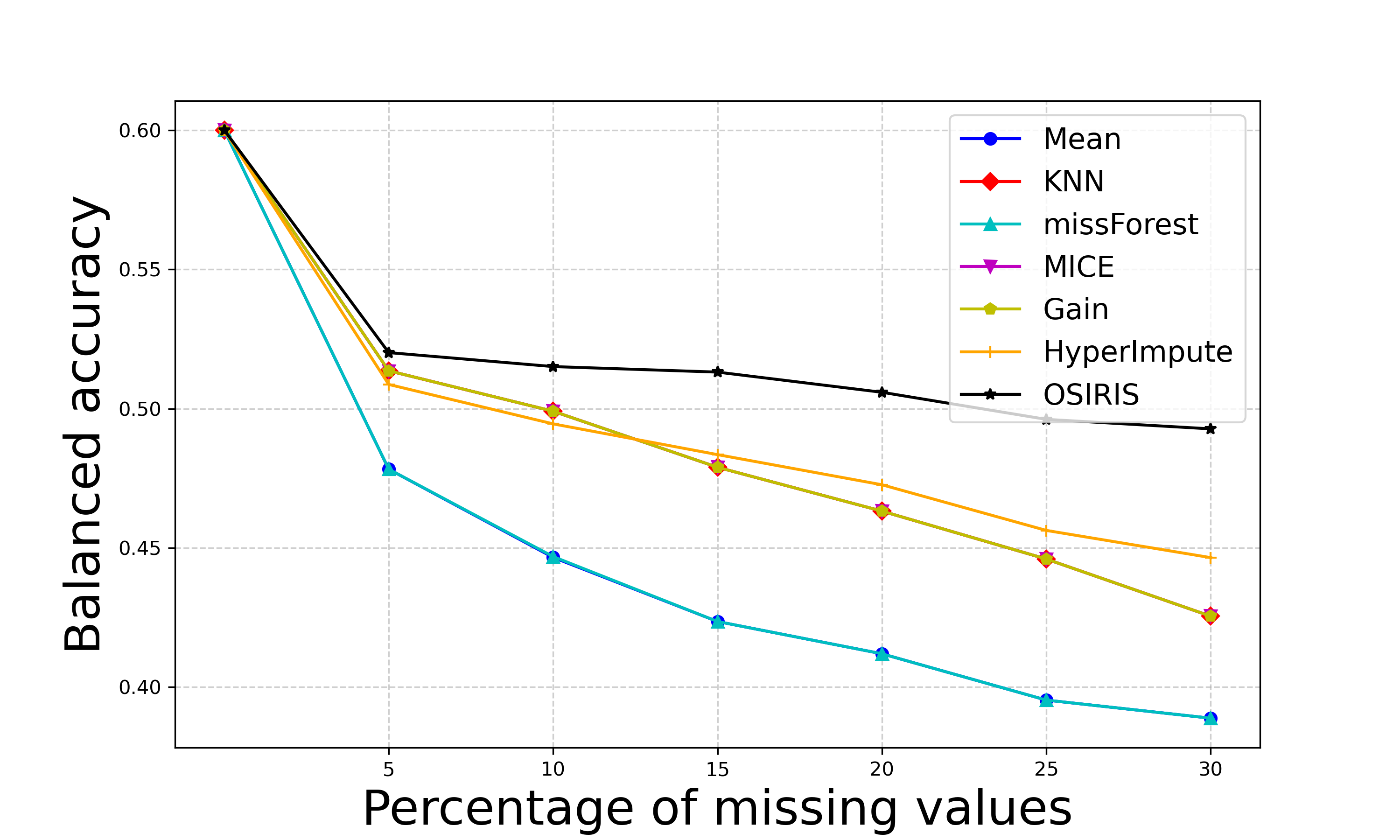}
        \caption{Scene15}
    \end{subfigure}
    \hfill
    \begin{subfigure}{0.32\textwidth}
        \includegraphics[trim={0.5cm 0 2.5cm 0},clip,width=\linewidth]{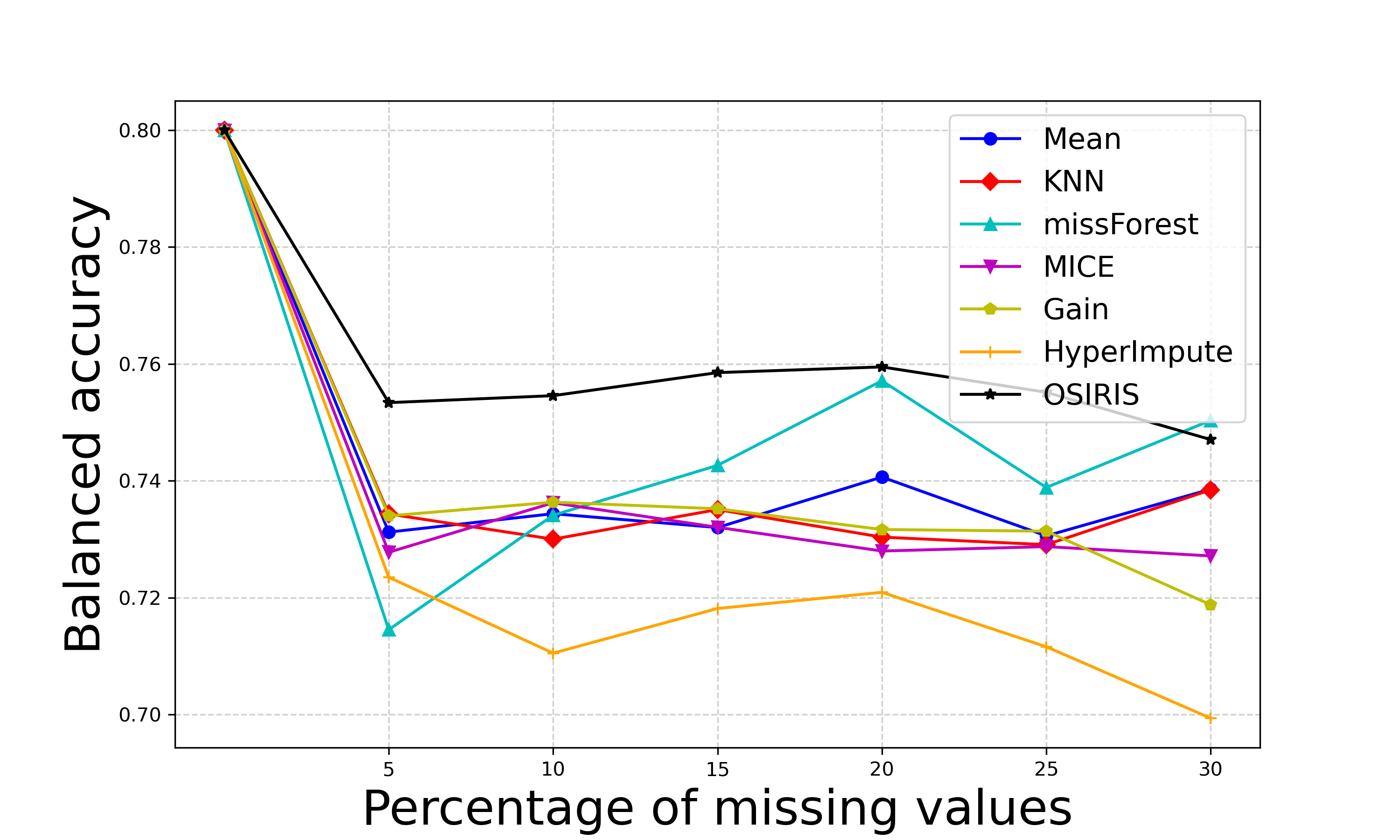}
        \caption{ADNI}
    \end{subfigure}
    \caption{Evolution of OSIRIS BACC as missing value rates increase, comparing its internal imputation mechanism to external imputation methods applied as preprocessing. \carlos{It displays the mean performance of each model across ten randomly generated missing masks.}
    }
    \label{fig:8figs_3x3}
\end{figure}

\albert{Figure \ref{fig:8figs_3x3} shows the performance results obtained across the most representative datasets\footnote{The Tadpole dataset was excluded due to its pronounced level of inherent missingness, while Fashion and Digit were omitted as their performance remained largely invariant across methods (potentially due to the simplicity of the task)}. Overall, the OSIRIS imputation strategy consistently achieves the highest BACC, most notably in the ADNI, Scene15, and XRay datasets, while remining competitive on the other datasets. Importantly, on the XRay dataset, OSIRIS exhibits a remarkable robustness, as it sustains high predictive accuracy even under conditions of progressively increasing missingness, a scenario in which competing baselines tend to deteriorate more rapidly. Furthermore, while in certain datasets the method attains performance levels comparable to the other algorithms, it systematically remains among the top-performing approaches.

These results highlight the advantages of imputing incomplete observations within the generative space $G$. This latent space, although constructed with partial supervision by incorporating label information, remains disentangled from the task-specific representation space $Z$, thereby mitigating the risk of introducing classification bias. Crucially, the imputation mechanism in OSIRIS is grounded in reconstructing missing modalities by maximizing the reconstruction fidelity of the observed ones, while explicitly encoding which features belong to each modality and modeling their interrelations. This design enables the method to capture complex cross-modality dependencies that conventional imputation techniques cannot represent, and explains its empirical superiority. Thus, OSIRIS can be used not only for prediction and discovery of multimodal relationships, but also as a principled framework for multimodal data imputation.}



\subsection{Application to multimodal clinical data}
\label{sec:neuro_bio}

In this section, we demonstrate how our approach applies to the clinical setting, in particular, to a multimodal problem from the ADNI consortium. The study incorporates 13 distinct data modalities (neurological, physiological, psychological, and demographic) collected from Alzheimer’s patients (see Table \ref{tab:ADNI_modalities}). It is important to note that, given the multimodal nature of the study, conducted across four phases, the dataset contains a substantial amount of missing data. In many cases, entire modalities were not collected during specific phases. Regarding the task, due to its clinical relevance in Alzheimer's research, we focused on distinguishing between the two stages of neurodegeneration preceding Alzheimer's disease: Early and Late Mild Cognitive Impairment (EMCI and LMCI), using only baseline data, i.e., information collected at the initial visit (timepoint 0) of the study.

\begin{table}[htbp]
\caption{Description of the modalities used for the ADNI database. Brain imaging data (ROIs, cortical parcellation, and amyloid PET scans) were preprocessed and extracted using the Computational Anatomy Toolbox 12 (CAT12) \cite{gaser2024cat}. Dataset description acronyms are: ROIs (Regions of Interest), MRI (Magnetic Resonance Imaging), PET (Positron Emission Tomography), GM (gray matter), WM (white matter), and CSF (cerebrospinal fluid).}
\label{tab:ADNI_modalities}
\centering
\begin{adjustbox}{max width=\textwidth}
\begin{tabular}{l r r @{\hskip 20pt} l r r}
\toprule
\textbf{Modality} & \textbf{\# Features} & 
\textbf{$\%$ Missing} & \textbf{Modality} & \textbf{\# Features} &
\textbf{$\%$ Missing} \\
\cmidrule(r){1-3} \cmidrule(l){4-6}
GM ROIs  & 118 & 0.73 & Montreal cog. assess.  &  35 & 23.33 \\
WM ROIs  &  9 & 0.73 & Hachinski ischemia scale  &  9& 0.01 \\
CSF ROIs  & 9& 0.73 & Amyloid PET &  328& 39.66 \\
Cortical parcellation  & 152& 1.31 & Metabolites & 271& 32.78   \\
General MRI  & 4 & 0.0 & Bile acids & 23& 21.99 \\
Geriatrical depression  & 17& 0.09 & Demographics & 15& 0.01 \\
Functional activities  & 11& 0.94 &              &  &    \\
\bottomrule
\end{tabular}
\end{adjustbox}
\end{table}

We begin the study by analyzing OSIRIS' performance in different configurations to evaluate the individual contributions of the discriminative and generative spaces to prediction and assess the benefits of semi-supervised learning. Thus, after training OSIRIS in both semi-supervised and fully supervised modes, we compared its performance using three prediction setups: (i) both $\mathbf{Z}$ and $\mathbf{G}$ (default setup), (ii) only $\mathbf{Z}$, and (iii) only $\mathbf{G}$. As reported in Table~\ref{tab:spaces_metrics}, two main conclusions emerge: first, semi-supervised learning yields significantly better results than fully supervised training, highlighting the benefit of including unlabeled data through $\mathbf{G}$; and second, most of the predictive power resides in $\mathbf{Z}$, as models using only $\mathbf{G}$ perform considerably worse. These findings empirically support our design: $\mathbf{Z}$ captures task-relevant information, while $\mathbf{G}$ enables learning from incomplete or unlabeled data and models structured variability not directly related to the predictive task. To verify that the $\mathbf{G}$ and $\mathbf{Z}$ spaces were truly independent, as implied by the structure of their variational posterior means, we calculated the cosine similarity between every dimension of $\mathbf{Z}$ and every component of $\mathbf{G}$. The highest similarity we observed was $-0.032$, confirming that the two spaces are effectively orthogonal.

\begin{table}[h!]
\centering
\caption{Results of OSIRIS under semi-supervised and supervised settings, using the default formulation (ZG), task-oriented latent space only (Z), and generative latent space only (G).}
\begin{adjustbox}{max width=\textwidth}
\begin{tabular}{c ccc ccc}
\toprule
    & \multicolumn{3}{c}{\textbf{Semi-Supervised}} & \multicolumn{3}{c}{\textbf{Supervised}} \\ \cmidrule(r){2-4} \cmidrule(l){5-7}
    Spaces & ZG & Z & G & ZG & Z & G \\ \midrule
    AUC     &  \textbf{0.83} $\pm$ \textbf{0.03}  &  \textbf{0.82} $\pm$ \textbf{0.03}  &   0.70 $\pm$ 0.03 &  0.73 $\pm$ 0.03  &  0.74 $\pm$ 0.02  &  0.66 $\pm$ 0.03  \\
    \cellcolor{gray!10} BACC  & \cellcolor{gray!10} \textbf{0.81} $\pm$ \textbf{0.02}  & \cellcolor{gray!10} \textbf{0.81} $\pm$ \textbf{0.02}   & \cellcolor{gray!10} 0.69 $\pm$ 0.02 & \cellcolor{gray!10} 0.72 $\pm$ 0.04 & \cellcolor{gray!10} 0.72 $\pm$ 0.04   & \cellcolor{gray!10} 0.64 $\pm$ 0.04     \\
    \bottomrule
\end{tabular}    
\end{adjustbox}
\label{tab:spaces_metrics}
\end{table}

Next, to illustrate the compactness and interpretability offered by OSIRIS, we analyze the latent factors and input features selected by the model in both the generative and discriminative spaces. This analysis relies on the \carlos{weights} $\mathbf{W}$ and \carlos{loadings} $\mathbf{V}$, which define how input variables contribute to the discriminative and generative latent representations, respectively. Since the model was trained using 10 folds, each yielding slightly different latent structures, we focused on identifying stable latent dimensions across folds. Specifically, we computed the pairwise cosine similarity between each task-oriented weights $\mathbf{w}_{:,k}$ (or generative loading $\mathbf{v}_{:,s}$) in a given fold and its counterparts in the others. A latent factor was considered stable if the absolute cosine similarity of their associated weights or loadings exceeded 0.95 in at least 8 out of 10 folds.

\begin{figure*}[ht!]
    \centering
    \begin{subfigure}[b]{0.48\linewidth}
        \centering
        \includegraphics[height=5.5cm]{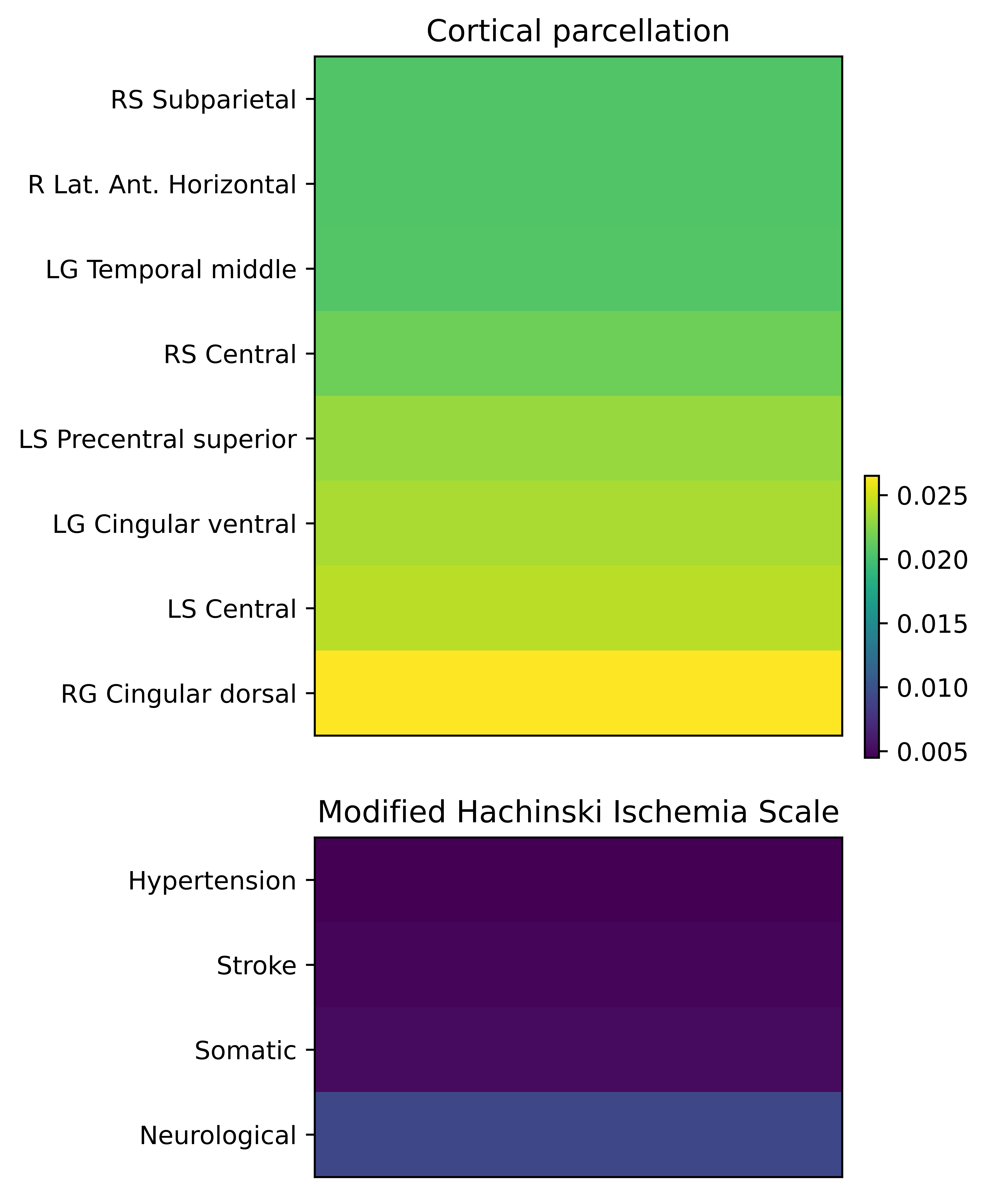}
        \caption{Stable discriminative \carlos{weight} matrix $\mathbf{W}$}
        \label{fig:stableW}
    \end{subfigure}
    \hspace{0.02\linewidth}
    \begin{subfigure}[b]{0.48\linewidth}
        \centering
        \includegraphics[height=5.5cm]{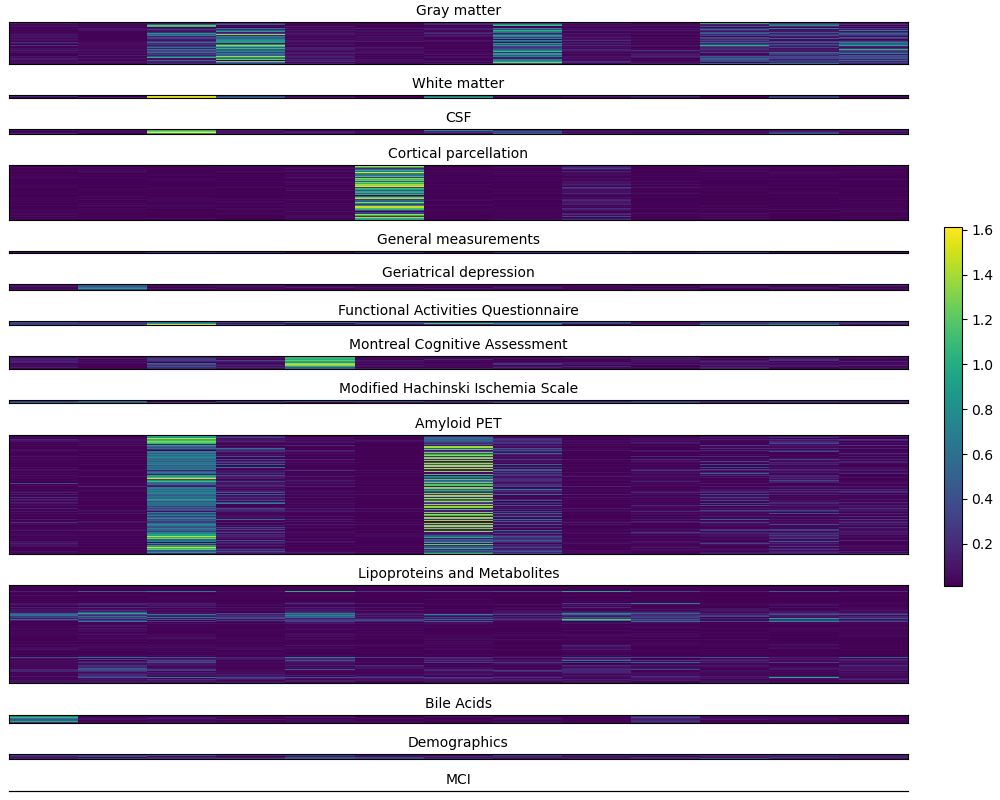}
        \caption{Stable generative \carlos{loading} matrix $\mathbf{V}$}
        \label{fig:stableV}
    \end{subfigure}
    \caption{Stable \carlos{weight} and \carlos{loading} matrices identified by both the task-oriented space (a) and the generative space (b). Columns represent the stable factors, while rows correspond to the features. In subfigure (a), the acronyms used denote the following: L (left), R (right), S (sulcus), and G (gyrus).}
    \label{fig:Both_spaces}
\end{figure*}

In particular, for the discriminative space $\mathbf{Z}$, which consists of a single latent factor dimension due to the binary nature of the classification task, we found that this latent was stable in 8 out of the 10 folds. Analysis of this stable factor (Figure \ref{fig:stableW}) showed that all views were pruned  out except for the cortical parcellation ROIs and the Hachinski Ischemia Scale,  and even within those two, only a subset of features was retained. Besides, the selected features have been previously reported in the literature as relevant for characterizing MCI, including regions such as the cingulate cortex 
\cite{fu2023decreased}, 
central sulcus 
\cite{liu2013longitudinal}, and temporal areas 
\cite{li2020accelerating}. 
Furthermore, there is substantial evidence supporting a strong link between cerebral ischemic lesions and the onset of MCI 
\cite{debette2010association}, commonly referred to as vascular dementia \cite{t2015vascular}.

In contrast, for the generative space $\mathbf{G}$, where the model tends to select around 25 latent dimensions, we identified 13 stable latent factors that consistently appeared in at least 8 out of 10 folds. By leveraging structured sparsity, our model separates modality-specific and shared latent factors that reflect meaningful neurobiological and psychological interactions. In Figure \ref{fig:stableV}, the absolute \carlos{loading} matrix $\mathbf{V}$ is shown with each of the 13 stable factors represented as a column. The color bar indicates each feature’s contribution (dark blue for no contribution and bright green to yellow for strong contribution). The pattern makes clear that each modality loads predominantly onto only a few factors, underscoring the model’s ability to uncover specific cross-modal associations.
\albert{The generative factors are presented in the Supplementary Material}.

\space

\section{Conclusions and Limitations}
\label{sec:conclusions}

We introduce a multimodal Bayesian framework designed to disentangle both discriminative and generative factors in complex, real-world multimodal clinical datasets. Its formulation enables robust generalization in low-sample-size scenarios, unbiased imputation of missing values, and extraction of interpretable discriminative and generative factors. Empirical results improve state-of-the-art performance, with the most robust imputation strategy and superior balanced accuracy compared to benchmark methods. Additionally, it enables the integration of heterogeneous clinical data and offers a practical way to uncover novel interactions among biological, psychological, and sociodemographic data. In future research, we intend to implement this model on clinical datasets containing more extensive multimodal information to throughly examine its potential. This will enable us to uncover latent factors that could have future clinical utility as multimodal biomarkers.

\albert{While the linear formulation may appear to be a limitation (preventing the model from capturing complex nonlinear relationships between modalities in more intricate settings), it significantly enhances model interpretability and provides robust solutions in wide-data scenarios. Moreover, its intrinsic feature selection (FS) mechanism further improves interpretability, enabling the identification of clinically relevant biomarkers}. Additionally, unlike some DL approaches, the model does not directly process structured data such as images or text. However, thanks to the wide availability of pretrained FMs, such data can still be integrated via precomputed embeddings, though this comes at the cost of reduced interpretability.

\newpage

\section*{Acknowledgements}

The ADNI and TADPOLE datasets used in this work were obtained from the ADNI study. The ADNI was launched in 2003 as a public-private partnership, led by Principal Investigator Michael W. Weiner, MD. For up-to-date information, see \url{http://www.adni-info.org}.

\section*{Author contributions}
ABL: conceptualization, data curation, formal analysis, investigation, methodology, software, writing original draft;
CSS: conceptualization, funding acquisition, investigation, methodology, resources, supervision, writing original draft;
JMM: conceptualization, project administration, supervision, writing - review and editing;
VGV: conceptualization, funding acquisition, project administration, resources, supervision, writing - original draft.

VGV and JMM jointly supervised this work as co-senior authors.

\section*{Data availability}

The data used in this project is publicly available and can be downloaded at the following sources: \href{https://archive.ics.uci.edu/dataset/5/arrhythmia}{Arrhythmia}, \href{https://scikit-learn.org/1.5/auto_examples/datasets/plot_digits_last_image.html}{Digit}, \href{https://landsat.gsfc.nasa.gov/data/}{Sat}, \href{https://www.kaggle.com/datasets/jessicali9530/lfw-dataset}{LFWA}, \href{https://www.kaggle.com/datasets/bothin/fashiongen-validation/data}{Fashion}, \href{https://tadpole.grand-challenge.org/Data/}{Tadpole}, \href{https://www.kaggle.com/datasets/nih-chest-xrays/data}{X-Ray}, \href{https://www.kaggle.com/datasets/zaiyankhan/15scene-dataset}{Scene15}, \href{https://adni.loni.usc.edu/}{ADNI}. To replicate the results, the specific data folds used in the experiments are provided in the project GitHub.

\appendix
\section{Model Inference}

\subsection{Approximate posterior distributions}

The model variables follow these prior distributions:
\begin{align}
    \Xnm  &\sim \N\p*{\Gn \VmT, \psim^{-1} \Is} \label{eq:x}\\
    \Gn   &\sim \N\p*{\bm{0}, \Ik} \\
    \mathbf{v}_{\rm d,s}^{\mathbf{Y}}  &\sim \N\p*{\bm{0}, (\delta_s^{(m+1)} \lambda_d^{(m+1)})^{-1}}\\
    \mathbf{v}_{\rm d,s}^{(m)}  &\sim \N\p*{\bm{0}, (\deltasm \lambda_d^{(m)})^{-1}} \\
    \deltasm &\sim \Gamma\p*{\alpha_s^{\deltam}, \beta_s^{\deltam}} \\
    \lambda_d^{(m)} &\sim \Gamma\p*{\alpha_d^{\boldsymbol{\lambda}^{(m)}}, \beta_d^{\boldsymbol{\lambda}^{(m)}}} \\
    \Zn  &\sim \N\p*{\sum_{m}^{M}\Xnm \WmT, \tau^{-1} \Ik} \\
    \Wkdm  &\sim \N\p*{0, (\phikm\gamdm)^{-1}} \\
    \phikm &\sim \Gamma\p*{\alpha_k^{\phim}, \beta_k^{\phim}} \\
    \gamdm &\sim \Gamma\p*{\alpha_d^{\gamm}, \beta_d^{\gamm}} \\
    \tau &\sim \Gamma\p*{\alpha^{\tau}, \beta^{\tau}} \\
    \Yn  &\sim \N\p*{\Zn \UT + \Gn \mathbf{V}^{\mathbf{Y}\top}, \eta^{-1} \Ic} \\
    \mathbf{u}_{\rm :,k}  &\sim \N\p*{\bm{0}, \Ic} \\
    \eta &\sim \Gamma\p*{\alpha^{\eta}, \beta^{\eta}}
\end{align}

To build the model, we use the mean-field approximation over the model variables $\mathbf{\Theta}$. As presented in \cite{bishop2006pattern}, the approximated posterior distribution of each $\theta_i$ will be defined as
\begin{equation}\label{eq:meanfield}
q^*(\theta_i) \propto \mathbb{E}_{\bm{\Theta}_{-i}} \left[ \log p(\bm{\Theta}, \XM, \Y) \right],
\end{equation}
where $\bm{\Theta}_{-i}$ includes all variables except $\theta_i$. Moreover, for further developments, the joint model distribution can be defined as:
\begin{equation}
\begin{split}
    p(\mathbf{Y},\mathbf{X}^\M,\mathbf{Z}, \mathbf{G},\tau, \eta, \mathbf{V}^\M,\mathbf{V}^{\mathbf{Y}}, \mathbf{W}^\M,\mathbf{U}, \boldsymbol{\delta}^\M, \boldsymbol{\lambda}^\M, \boldsymbol{\phi}^\M, \boldsymbol{\gamma}^\M) = p(\mathbf{T}|\mathbf{Y})p(\mathbf{Y}\vert \mathbf{Z},\mathbf{U}, \mathbf{G}, \mathbf{V}^{\mathbf{Y}}, \eta) \\
    p(\mathbf{X}^\M\vert \mathbf{G},\mathbf{V}^\M,\psi)p(\mathbf{G})p(\mathbf{V}^\M\vert \boldsymbol{\delta}^\M,\boldsymbol{\lambda}^\M)p(\mathbf{V}^{\mathbf{Y}}\vert \boldsymbol{\delta}^{(M+1)},\boldsymbol{\lambda}^{(M+1)})p(\boldsymbol{\delta}^\M)p(\boldsymbol{\lambda}^\M)p(\mathbf{Z}\vert \mathbf{W}^\M, \tau)\\
    p(\mathbf{W}^\M\vert \boldsymbol{\phi}^\M,\boldsymbol{\gamma}^\M)p(\boldsymbol{\phi}^\M)p(\boldsymbol{\gamma}^\M)p(\mathbf{U})p(\tau)p(\eta)p(\psi^\M)
\end{split}
\end{equation}

This way, we get that the approximated posterior distribution can be written as 
\begin{align}
q\left(\boldsymbol{\Theta}\right) \eqeq \prod_{m=1}^{M} \p*{q(\mathbf{V}^{(m)})q(\mathbf{W}^{(m)}) \, q(\psim) \prod_{k}^K \p*{q(\delta_k^{(m)})q(\phi_k^{(m)})} \prod_{d}^{D^{(m)}} \p*{q(\lambda_d^{(m)})q(\gamma_d^{(m)})}} \nonumber\\
\eqline q(\mathbf{V}^{\mathbf{Y}}) q(\mathbf{U}) \prod_{n}^{N} \p*{q(\mathbf{z}_{\rm n,:}) q(\mathbf{g}_{\rm n,:})} q(\eta) q(\tau). \label{eq:qModel}
\end{align}

\subsubsection{\texorpdfstring{Mean field approximation of $\mathbf{Z}$}{Mean field approximation of Z}}

Following the mean-field procedure defined in Eq. \eqref{eq:meanfield}, we define the approximated posterior distribution of $\mathbf{Z}$ as:
\begin{equation}
\label{eq:main_z}
    \ln q(\mathbf{Z}) = \E[\ln p(\mathbf{Y}|\mathbf{Z}, \mathbf{U},\mathbf{G}, \mathbf{V}^{\mathbf{Y}},\eta) + \ln p(\mathbf{Z}|\mathbf{W}^{\M},\tau)].
\end{equation}

Thus, we can develop the first term of the summation as:
\begin{equation}
\begin{split}
\label{eq:Y_des}
    &\ln p(\mathbf{Y}|\mathbf{Z}, \mathbf{U},\mathbf{G}, \mathbf{V}^{\mathbf{Y}},\eta)  = \sum_n^N \ln \N\left(\Zn\UT + \Gn\mathbf{V}^{\mathbf{Y}\top}, \eta^{-1}\Ic 
    \right) \\
    & =\sum_n^N\left(-\frac{1}{2}\ln(2\pi) + \frac{1}{2}\ln(\eta) - \frac{\eta}{2}\left(\Yn - \Zn\UT - \Gn\mathbf{V}^{\mathbf{Y}\top}\right)\left(\Yn - \Zn\UT - \Gn\mathbf{V}^{\mathbf{Y}\top}\right)^\top\right)\\
    & = \sum_n^N\Big(-\frac{1}{2}\ln(2\pi) + \frac{1}{2}\ln(\eta) - \frac{\eta}{2}\Big[\Yn\Yn^\top -2\Yn\U\Zn^\top - 2\Yn\mathbf{V}^{\mathbf{Y}}\Gn^\top + \Zn\UT\U\Zn^\top \\ 
    & + \Gn\mathbf{V}^{\mathbf{Y}\top}\mathbf{V}^{\mathbf{Y}}\Gn^\top +2\Zn\UT\mathbf{V}^{\mathbf{Y}}\Gn^\top\Big]\Big).
\end{split}
\end{equation}

Hence, if we adapt Eq. (\ref{eq:Y_des}) focusing solely on $\mathbf{Z}$, treating as constants all terms that do not depend on this variable:
\begin{equation}
\begin{split}
\label{eq:z_first}
    & \ln p(\mathbf{Y}|\mathbf{Z}, \mathbf{U},\mathbf{G}, \mathbf{V}^{\mathbf{Y}},\eta) = \sum_n^N -\frac{\eta}{2}\left(-2 -2\Yn\U\Zn^\top + \Zn\UT\U\Zn^\top +2\Zn\UT\mathbf{V}^{\mathbf{Y}}\Gn^\top\right) \\
    & = \eta\sum_n^N\left(\Yn\U\Zn^\top - \frac{1}{2}\Zn\UT\U\Zn^\top -\Zn\UT\mathbf{V}^{\mathbf{Y}}\Gn^\top \right) + \const.
\end{split}
\end{equation}

The second term of Eq. (\ref{eq:main_z}) can be expressed as:
\begin{equation}
\label{eq:z_second}
    \ln p(\mathbf{Z}|\mathbf{W}^{\M},\tau) = \sum_n^N -\frac{1}{2}\tau\left(\Zn\Zn^\top - \Zn\sum_m^M\left(\Xnm\WmT\right)\right) + \const.
\end{equation}

We can now combine Eq. (\ref{eq:z_first}) and Eq. (\ref{eq:z_second}) in order to apply the expectation operator and identify the relevant terms:
\begin{equation}
\begin{split}
    & \ln q(\mathbf{Z}) = \E\Big[\sum_n^N\Big(\Zn\eta\UT\Yn^\top - \frac{1}{2}\Zn\eta\UT\U\Zn^\top -\Zn\eta\UT\mathbf{V}^{\mathbf{Y}\top}\Gn^\top - \frac{1}{2}\Zn\tau\Zn^\top \\
    & + \Zn\tau\sum_m^M\left(\Xnm\WmT\right)\Big)\Big] + \const \\
    & = \sum_n^N\E\Bigg[\Zn\left(\eta\UT\Yn^\top - \eta\UT\mathbf{V}^{\mathbf{Y}}\Gn^\top + \tau\sum_m^M\left(\Xnm\WmT\right)\right) \\
    & - \frac{1}{2}\Zn\left(\eta\UT\U + \tau\Ik\right)\Zn^\top\Bigg] + \const \\
    & = \sum_n^N\Bigg(\Zn\left(\langle\eta\rangle\langle\UT\rangle\langle\Yn^\top\rangle - \langle\eta\rangle\langle\UT\rangle\langle\mathbf{V}^{\mathbf{Y}}\rangle\langle\Gn^\top\rangle + \langle\tau\rangle\sum_m^M\left(\Xnm\langle\WmT\rangle\right)\right) \\
    & - \frac{1}{2}\Zn\left(\langle\eta\rangle\langle\UT\U\rangle + \langle\tau\rangle\Ik\right)\Zn^\top\Bigg) + \const
\end{split}
\end{equation}

Thus, we can identify the different elements and subsequently parameterize the approximated posterior
\begin{equation}
    q(\Zn) = \N(\langle\Zn\rangle, \Sigma_{\Z}^{-1})
\end{equation}
where
\begin{align}
    \Sigma_{\mathbf{Z}}^{-1} & = \langle\tau\rangle \Ik + \langle\eta\rangle\langle\mathbf{U}^\top\mathbf{U}\rangle \\
    \langle\mathbf{z}_{\rm n,:}\rangle & = \left(\langle\tau\rangle\sum_m^M\mathbf{x}_{\rm n,:}^{(m)}\langle\mathbf{W}^{(m)}\rangle^\top  +\langle \eta \rangle \left(\langle\mathbf{y}_{\rm n,:}\rangle - \langle\mathbf{g}_{\rm n,:} \rangle \langle\mathbf{V}^{\mathbf{Y} }  \rangle^\top \right)\langle\mathbf{U}\rangle \right)\Sigma_{\mathbf{Z}}
\end{align}

\subsubsection{\texorpdfstring{Mean field approximation of $\mathbf{G}$}{Mean field approximation of G}}

Following Eq. \eqref{eq:meanfield}, we define
\begin{equation}
    \ln q(\G) = \E[\ln p(\G) + \ln p(\X\vert \G, \V^\M, \psi^\M) + \ln p(\Y\vert \Z, \U, \G, \mathbf{V}^{(M+1)}, \eta)]
\end{equation}

First term is developed as follows:
\begin{equation}
\label{eq:g_first}
    \ln p(\G) = \sum_n^N\left(-\frac{1}{2}\Gn\Gn^\top\right) + \const.
\end{equation}

Second term as:
\begin{equation}
\label{eq:g_second}
    \ln p(\X\vert \G, \V^\M, \psi^\M) = \sum_m^M\sum_n^N\left(\psi^{(m)}\Xnm\V\Gn^\top - \frac{\psim}{2}\Gn\VmT\Vm\Gn^\top\right) + \const.
\end{equation}

 Third term is analogous to Eq. (\ref{eq:Y_des}) as:
 \begin{equation}
 \begin{split}
 \label{eq:g_third}
     \ln p(\Y\vert \Z, \U, \G, \mathbf{V}^{(M+1)}, \eta)  &= \sum_n^N\Big(-\frac{\eta}{2}\Big[-2\Yn\mathbf{V}^{\mathbf{Y}}\Gn^\top + \Gn\mathbf{V}^{\mathbf{Y}}\mathbf{V}^{\mathbf{Y}\top}\Gn^\top \\& + 2\Zn\UT\mathbf{V}^{\mathbf{Y}}\Gn^\top\Big]\Big) + \const.
 \end{split}
 \end{equation}

 We combine Eq. (\ref{eq:g_first}), (\ref{eq:g_second}) and (\ref{eq:g_third}) as:
 \begin{equation}
    \begin{split}
        &\ln q(\G) = \E\Bigg[\sum_n^N\Bigg(\sum_m^M\left(\psim\Xnm\Vm\right)\Gn^\top - \frac{1}{2}\Gn\sum_m^M\left(\psim\VmT\Vm\right)\Gn^\top \\
        & -\frac{1}{2}\Gn\Gn^\top + \eta\Yn\mathbf{V}^{\mathbf{Y}}\Gn^\top - \frac{1}{2}\Gn\left(\eta\mathbf{V}^{\mathbf{Y}}\mathbf{V}^{\mathbf{Y}\top}\right)\Gn^\top - \eta\Zn\UT\mathbf{V}^{\mathbf{Y}}\Gn^\top\Bigg)\Bigg] + \const \\
        & = \sum_n^N\Bigg(\Bigg[\sum_m^M\langle\psim\rangle\Xnm\langle\Vm\rangle + \langle\eta\rangle\langle\Yn\rangle\langle\mathbf{V}^{\mathbf{Y}}\rangle - \langle\eta\rangle\langle\Zn\rangle\langle\UT\rangle\langle\mathbf{V}^{\mathbf{Y}}\rangle\Bigg]\Gn^\top\\
        & - \frac{1}{2}\Gn\left(\sum_m^M\left(\langle\psim\rangle\langle\VmT\Vm\rangle\right) + \Is + \langle\eta\rangle\langle\mathbf{V}^{\mathbf{Y}\top}\mathbf{V}^{\mathbf{Y}}\rangle\right)\Gn^\top\Bigg) + \const.
    \end{split}
 \end{equation}

Hence, identifying the elements we can parameterize:
\begin{equation}
    q(\Gn) = \N\left(\langle\Gn\rangle, \Sigma_{\G}^{-1}\right),
\end{equation}
where
\begin{align}
    \Sigma_{\mathbf{G}}^{-1} &= \Is + \langle\eta\rangle\langle\mathbf{V}^{\mathbf{Y}\top} \mathbf{V}^{\mathbf{Y}}\rangle + \sum_m^{M}\langle\psi^{(m)}\rangle\langle\mathbf{V}^{(m)^\top}\mathbf{V}^{(m)}\rangle \\
    \langle\mathbf{g}_{\rm n,:}\rangle &= \left( \sum_m^M\left(\langle\psi^{(m)}\rangle\mathbf{x}_{\rm n,:}^{(m)}\langle\mathbf{V}^{(m)}\rangle\right) + \langle\eta\rangle   \left( \langle\mathbf{y}_{n,:}\rangle -\langle\mathbf{z}_{\rm n,:}\rangle\langle\mathbf{U}\rangle^\top \right)\langle\mathbf{V}^{\mathbf{Y}}\rangle \right)\Sigma_{\mathbf{G}}
\end{align}

\subsubsection{\texorpdfstring{Mean field approximation of $\U$}{Mean field approximation of U}}

By following Eq. \eqref{eq:meanfield}, we define:
\begin{equation}
    \ln p(\U) = \E\left[\ln p(\U) + \ln p(\Y\vert \Z, \U, \G, \mathbf{V}^{\mathbf{\Y}},\eta)\right].
\end{equation}

First term can be expressed as
\begin{equation}
\label{eq:u_first}
    \ln p(\U) = -\frac{1}{2}\sum_k^K\mathbf{u}_{\rm :,k}^\top\mathbf{u}_{\rm :,k} + \const.
\end{equation}

As in Eq. (\ref{eq:Y_des}), second term can be expressed as
\begin{equation}
\begin{split}
\label{eq:u_second}
    &\ln p(\Y\vert \Z, \U, \G, \mathbf{V}^{\mathbf{\Y}},\eta) \\
    & = \sum_n^N -\frac{\eta}{2}\left[-2\Yn\U\Zn^\top + \Zn\UT\U\Zn^\top + 2\Zn\UT\mathbf{V}^{\mathbf{Y}}\Gn^\top\right] + \const \\
    & = -\frac{\eta}{2}\sum_c^C\left(-2\mathbf{y}_{\rm :,c}^\top\Z\mathbf{u}_{\rm c,:}^\top + \mathbf{u}_{\rm c,:}\ZT\Z\mathbf{u}_{\rm c,:}^\top + 2 \mathbf{v}_{\rm c,:}^{\mathbf{Y}}\GT\Z\mathbf{u}_{\rm c,:}^\top\right) + \const.
\end{split}
\end{equation}

If we join Eq. (\ref{eq:u_first}) and (\ref{eq:u_second}), apply expectation and identify terms, we can parameterize
\begin{equation}
    q(\mathbf{u}_{\rm c,:}) = \N\left(\langle\mathbf{u}_{\rm c,:}\rangle, \Sigma_{\mathbf{u}_{\rm c,:}}^{-1}\right),
\end{equation}
where
\begin{align}
    \Sigma_{\mathbf{u}_{\rm c,:}}^{-1} & = \Ik + \langle\eta\rangle\langle\ZT\Z\rangle \\
    \langle\mathbf{u}_{\rm c,:}\rangle & = \left[\langle\eta\rangle\langle\mathbf{y}_{\rm :,c}^\top\rangle\langle\Z\rangle - \langle\eta\rangle\langle\mathbf{v}_{\rm c,:}^{\mathbf{Y}}\rangle\langle\GT\rangle\langle\Z\rangle\right]\Sigma_{\mathbf{u}_{\rm c,:}}.
\end{align}

\subsubsection{\texorpdfstring{Mean field approximation of $\mathbf{V}^{\mathbf{Y}}$}{Mean field approximation of V Y}}

As in previous variables, we express Eq. (\ref{eq:meanfield}) as
\begin{equation}
    \ln q(\mathbf{V}^{\mathbf{Y}}) = \E\left[\ln p\left(\Y\vert \Z, \U, \G, \mathbf{V}^{\mathbf{Y}}, \eta\right) + \ln p\left(\mathbf{V}^{\mathbf{Y}}\vert\boldsymbol{\lambda}^{(M+1)}, \boldsymbol{\delta}^{(M+1)}\right)\right].
\end{equation}

Similarly to Eq. (\ref{eq:Y_des}), we develop the first term as
\begin{equation}
\begin{split}
\label{eq:vy_first}
    & \ln p\left(\Y\vert \Z, \U, \G, \mathbf{V}^{\mathbf{Y}}, \eta\right) \\
    & = \sum_n^N -\frac{\eta}{2}\left(-2 \Yn\mathbf{V}^{\mathbf{Y}}\Gn^\top + \Gn\mathbf{V}^{\mathbf{Y}\top}\mathbf{V}^{\mathbf{Y}}\Gn^\top + 2\Zn\UT\mathbf{V}^{\mathbf{Y}}\Gn^\top\right) + \const \\
    & = -\frac{\eta}{2}\sum_c^C\sum_n^N\left(-2\mathbf{v}_{\rm c,:}^{\mathbf{Y}} \Gn^\top y_{n,c}\right) - \frac{\eta}{2}\sum_c^C\left(\mathbf{v}_{\rm c,:}^{\mathbf{Y}}\GT\G\mathbf{v}_{\rm c,:}^{\mathbf{Y}\top}\right) \\
    & - \frac{\eta}{2}\sum_c^C\sum_n^N\left(2\mathbf{v}_{\rm c,:}^{\mathbf{Y}}\Gn^\top\mathbf{u}_{\rm c,:}\Zn^\top\right) + \const.
\end{split}
\end{equation}

Second term can be expressed as
\begin{equation}
\begin{split}
\label{eq:vy_second}
    & \ln p\left(\mathbf{V}^{\mathbf{Y}}\vert\boldsymbol{\delta}^{(M+1)}, \boldsymbol{\lambda}^{(M+1)}\right) = 
    \sum_s^S \ln \N\left(\mathbf{v}_{\rm :,s}^{\mathbf{Y}}\vert 0, \delta_s^{(M+1)-1}\Lambda_{\boldsymbol{\lambda}^{(M+1)}}^{-1}\right) \\
    & = \sum_s^S \left(-\unmed\lnpi + \unmed\ln(\delta_s^{(M+1)}\Lambda_{\boldsymbol{\lambda}^{(M+1)}}) - \frac{\delta_s^{(M+1)}}{2}\left(\mathbf{v}_{\rm :,s}^{\mathbf{Y}\top}\Lambda_{\boldsymbol{\lambda^{(M+1)}}}\mathbf{v}_{\rm :,s}^{\mathbf{Y}}\right)\right) + \const \\
    & = -\frac{1}{2}\sum_c^C \left(\lambda_c^{(M+1)}\mathbf{v}_{\rm c,:}^{\mathbf{Y}}\Lambda_{\boldsymbol{\delta}}\mathbf{v}_{\rm c,:}^{\mathbf{Y}\top}\right) + \const.
\end{split}
\end{equation}

We can join Eq. (\ref{eq:vy_first}) and (\ref{eq:vy_second}), apply expectations and identify terms to parameterize
\begin{equation}
    q\left(\mathbf{v}_{\rm c,:}^{\mathbf{Y}}\right) = \N\left(\langle\mathbf{v}_{\rm c,:}^{\mathbf{Y}}\rangle, \Sigma_{\mathbf{v}_{\rm c,:}^{\mathbf{Y}}}^{-1}\right),
\end{equation}
where
\begin{align}
    \Sigma_{\mathbf{v}_{\rm c,:}^{\mathbf{Y}}}^{-1} & = \langle\eta\rangle \langle\GT\G\rangle + \langle\gamma_c^{(M+1)}\rangle\Lambda_{\langle\boldsymbol{\phi}\rangle} \\
    \langle\mathbf{v}_{\rm c,:}^{\mathbf{Y}}\rangle & = \langle\eta\rangle\sum_n^N\langle\Gn\rangle\langle y_{\rm n,c}\rangle - \langle\eta\rangle\sum_n^N\langle\Zn\rangle\langle\mathbf{u}_{\rm c,:}^\top\rangle\langle\Gn\rangle
\end{align}

\subsubsection{\texorpdfstring{Mean field approximation of $\eta$}{Mean field approximation of eta}}

We use Eq. (\ref{eq:meanfield}) to get
\begin{equation}
    \ln q(\eta) = \E\left[\ln p(\eta) + \ln p(\Y\vert\Z,\U, \G, \mathbf{V}^{\mathbf{Y}},\eta)\right].
\end{equation}

The first term, as follows a gamma distribution, can be expressed as
\begin{equation}
\label{eq:eta_first}
    \ln p(\eta) = \beta_0^\eta\alpha^\eta + (\alpha_0^\eta -1)\ln (\alpha^\eta) + \const.
\end{equation}

Second term is analogous to Eq. (\ref{eq:Y_des}), however, we operated over the model variables with the Trace operator to be able to apply further expectations. Thus, Eq. (\ref{eq:Y_des}) can be expressed as
\begin{equation}
\begin{split}
\label{eq:eta_second}
    &\ln p(\mathbf{Y}|\mathbf{Z}, \mathbf{U},\mathbf{G}, \mathbf{V}^{\mathbf{Y}},\eta)  =  \sum_n^N\Big(-\frac{1}{2}\ln(2\pi) + \frac{1}{2}\ln(\eta) - \frac{\eta}{2}\Big[\Yn\Yn^\top -2\Yn\U\Zn^\top  \\ &- 2\Yn\mathbf{V}^{\mathbf{Y}}\Gn^\top + \Zn\UT\U\Zn^\top + \Gn\mathbf{V}^{\mathbf{Y}\top}\mathbf{V}^{\mathbf{Y}}\Gn^\top +2\Zn\UT\mathbf{V}^{\mathbf{Y}}\Gn^\top\Big]\Big) \\
    & = \sum_n^N\Big(\frac{1}{2}\ln\eta - \frac{\eta}{2}\Big[\Yn\Yn^\top -2 \Yn\U\Zn^\top - 2\Yn\mathbf{V}^{\mathbf{Y}}\Gn^\top + Trace\left(\UT\U\Zn^\top\Zn\right)\\
    &+ Trace\left(\mathbf{V}^{\mathbf{Y}\top}\mathbf{V}^{\mathbf{Y}}\Gn^\top\Gn\right) +2\Zn\UT\mathbf{V}^{\mathbf{Y}}\Gn^\top\Big]\Big) + \const
\end{split}
\end{equation}

Now, we join Eq. (\ref{eq:eta_first}) and (\ref{eq:eta_second}) and apply expectattions to parameterize
\begin{equation}
    q\left(\eta\right) = \Gamma(\alpha^\eta, \beta^\eta),
\end{equation}
where
\begin{equation}
    \alpha^\eta  = \frac{NC}{2} + \alpha^\eta_0
\end{equation}
and
\begin{equation}
\begin{split}
    \beta^\eta & = \beta_0^\eta + \frac{1}{2}\sum_n^N\Big(\langle\Yn\Yn^\top\rangle -2\langle\Yn\rangle\langle\U\rangle\langle\Zn^\top\rangle -2 \langle\Yn\rangle\langle\mathbf{V}^{\mathbf{Y}}\rangle\langle\Gn^\top\rangle \\
    & + Trace\left(\langle\UT\U\rangle\langle\Zn^\top\Zn\rangle\right) + Trace\left(\langle\mathbf{V}^{\mathbf{Y}\top}\mathbf{V}^{\mathbf{Y}}\rangle\langle\Gn^\top\Gn\rangle\right) \\& + 2\langle\Zn\rangle\langle\UT\rangle\langle\mathbf{V}^{\mathbf{Y}}\rangle\langle\Gn^\top\rangle\Big)
\end{split}
\end{equation}


\subsubsection{\texorpdfstring{Mean Field Approximation of $\mathbf{W}^\M$}{Mean Field Approximation of W M}}

\label{sec:inf_A}

As stated in the mean-field inference procedure, we use Eq. (\ref{eq:meanfield}) to obtain:
\begin{equation}
\label{eq:E_a}
    \ln q(\mathbf{W}^\M) = \E\left[\ln p(\mathbf{Z}|\mathbf{W}^{\M},\tau) + \ln p(\mathbf{W}^\M|\boldsymbol{\phi}^\M, \boldsymbol{\gamma}^\M)\right],
\end{equation}
where $\M = \{1, 2, \dots, M\}$ corresponds to the set of the $M$ views.

Thus, we can develop the first term as:
\begin{equation}
\begin{split}
\label{eq:z_para_a}
    & \ln p(\mathbf{Z}|\mathbf{W}^{\M},\tau)  = \sumn\ln\N\summ\left(\mathbf{x}_{n,:}^{(m)}\Wm^\top, \tau\right) \\
    & = \sumn \Bigg(-\unmed\lnpi + \unmed\ln \tau - \frac{\tau}{2}\left(\Zn- \summ \mathbf{x}_{n,:}^{(m)}\Wm^\top\right) \left(\Zn - \summ \mathbf{x}_{n,:}^{(m)}\Wm^\top\right)^\top\Bigg).
\end{split}
\end{equation}

As we will further apply the expectation over $\mathbf{W}^\M$, we can rewrite Eq. (\ref{eq:z_para_a}) as:
\begin{equation}
\begin{split}
\label{eq:w_first}
    & \ln p(\mathbf{Z}|\mathbf{W}^{\M},\tau) = \\
    & = \sumn\Bigg(-\frac{\tau}{2}\Bigg(-2\Zn\summ \left(\mathbf{x}_{n,:}^{(m)}\Wm^\top\right)^\top + \summ \left(\mathbf{x}_{n,:}^{(m)}\Wm^\top\right)\summ \left(\mathbf{x}_{n,:}^{(m)}\Wm^\top\right)^\top\Bigg)\Bigg) + \const \\
    & = \sumn\Bigg(-\frac{\tau}{2}\Bigg(-2\Zn\summ \left(\mathbf{x}_{n,:}^{(m)}\Wm^\top\right)^\top + \summ\sum_{m'}^M \left(\mathbf{x}_{n,:}^{(m)}\Wm^\top\right)\left(\mathbf{x}_{n,:}^{(m')}\mathbf{W}^{(m')\top}\right)^\top\Bigg)\Bigg) + \const \\
    & = \sumn\Bigg(-\frac{\tau}{2}\Bigg(-2\Zn\summ \left(\mathbf{x}_{n,:}^{(m)}\Wm^\top\right)^\top + \summ\sum_{m'}^M\Bigg(\Xnm\Wm\mathbf{W}^{(m')\top}\mathbf{x}_{\rm n,:}^{(m')\top}\Bigg)\Bigg)\Bigg) + \const \\
    & = \sumn\sumk\Bigg(-\frac{\tau}{2}\Bigg(-2 z_{\rm n,k}\summ \mathbf{x}_{\rm n,:}^{(m)}\mathbf{w}_{\rm k,:}^{(m)^\top} + \summ\sum_{m'}^M\Bigg(\mathbf{w}_{\rm k,:}^{(m)}\mathbf{x}_{\rm n,:}^{(m')\top}\mathbf{x}_{\rm n,:}^{(m')}\mathbf{w}_{\rm k,:}^{(m')\top}\Bigg)\Bigg)\Bigg) + \const\\
    & = \sumn\sumk\Bigg(\unmed\ln\tau -\frac{\tau}{2}\Bigg(-2 z_{\rm n,k}\summ \mathbf{x}_{\rm n,:}^{(m)}\mathbf{w}_{\rm k,:}^{(m)^\top} \\
    & + \summ\sum_{m' \neq m}^M \Bigg(\mathbf{w}_{\rm k,:}^{(m)}\mathbf{x}_{\rm n,:}^{(m)^\top}\mathbf{x}_{\rm n,:}^{(m')}\mathbf{w}_{\rm k,:}^{(m')\top}\Bigg) + \summ \Bigg(\mathbf{w}_{\rm k,:}^{(m)}\mathbf{x}_{\rm n,:}^{(m)^\top}\mathbf{x}_{\rm n,:}^{(m)}\mathbf{w}_{\rm k,:}^{(m)^\top}\Bigg)\Bigg)\Bigg) + \const
\end{split}
\end{equation}

Second term is analogous to Eq. (\ref{eq:vy_second}), and can be developed as
\begin{equation}
\begin{split}
\label{eq:w_second}
    & \ln p\left(\mathbf{W}^{(m)}\vert\boldsymbol{\phi}^{(m)}, \boldsymbol{\gamma}^{(m)}\right) = 
    \sum_k^K \ln \N\left(\mathbf{w}_{\rm k,:}^{(m)}\vert 0, \phi_k^{(m)-1}\Lambda_{\boldsymbol{\gamma}^{(m)}}^{-1}\right) \\
    & = \sum_k^K \left(-\unmed\lnpi + \unmed\ln(\phi_k^{(m)}\Lambda_{\boldsymbol{\gamma}^{(m)}}) - \frac{\phi_k^{(m)}}{2}\left(\mathbf{w}_{\rm k,:}\Lambda_{\boldsymbol{\gamma^{(m)}}}\mathbf{w}_{\rm k,:}^\top\right)\right) \\
    & = -\frac{1}{2}\sum_k^K \left(\phi_k^{(m)}\mathbf{w}_{\rm k,:}\Lambda_{\boldsymbol{\gamma}^{(m)}}\mathbf{w}_{\rm k,:}^\top\right) + \const.
\end{split}
\end{equation}

Now, if we join both Eq. (\ref{eq:w_first}) and (\ref{eq:w_second}) we obtain:
\begin{equation}
\begin{split}
\label{eq:final_aa}
    &\ln q(\mathbf{w}_{k,:}^{(m)}) = \E\Bigg[\sumn\sumk\Bigg(\mathbf{w}_{\rm k,:}^{(m)}\left(-\frac{\tau}{2}\mathbf{x}_{\rm n,:}^{(m)}\mathbf{x}_{\rm n,:}^{(m)^\top}- \frac{\phi_k^{(m)}}{2}\mathbf{\tilde{x}}_{\rm n,:}^{(m)}\Lambda_{\boldsymbol{\gamma}^{(m)}}\mathbf{\tilde{x}}_{\rm n,:}^{(m)^\top}\right)\mathbf{w}_{\rm k,:}^{(m)^\top} \\
    & + \sumk\Bigg(\tau\mathbf{z}_{\rm :,k}^\top\mathbf{X}^{(m)} - \tau\sum_{m' \neq m}^M\left(\mathbf{w}_{\rm k,:}^{(m')}\mathbf{X}^{(m')\top}\right)\mathbf{X}^{(m)}\Bigg)\mathbf{w}_{\rm k,:}^{(m)^\top}\Bigg] + \const.
\end{split}
\end{equation}

Finally, if we apply the expectation over Eq. \eqref{eq:final_aa} and identify terms, we can parameterize the approximated posterior $q(\mathbf{w}_{k,:}^{(m)})$ as:
\begin{equation}
    q(\mathbf{w}_{k,:}^{(m)}) = \N(\langle\mathbf{w}_{k,:}^{(m)}\rangle, \Sigma_{\mathbf{w}_{k,:}^{(m)}}^{-1})
\end{equation}
where 
\begin{align}
   \Sigma_{\Wk}^{(m)-1} & = \tauE \XmT\Xm + \langle\phi_k^{(m)} \rangle \Lambda_{\langle\boldsymbol{\gamma}^{(m)}\rangle}
        \\ \langle \Wkm\rangle & = \tauE\sum_n^N \left[\langle z_{\rm n, k}\rangle - \sum_{m' \neq m}^{M}\langle\mathbf{w}_{\rm k,:}^{(m')}\rangle \mathbf{x}^{(m') \top}_{\rm n,:}\right] \mathbf{x}_{\rm n,:}^{(m)}\Sigma_{\Wk}^{\rm (m)}
\end{align}

\subsubsection{\texorpdfstring{Mean field approximation of $\tau$}{Mean field approximation of tau}}

As the mean-field approximation states, we use Eq. (\ref{eq:meanfield}) for $\tau$ as:
\begin{equation}
\label{eq:E_tau}
    \ln q(\tau) = \E\left[\ln p(\mathbf{Z}|\mathbf{W}^{\M},\tau) + \ln p(\tau)\right].
\end{equation}

The first term can be developed similarly as Eq. (\ref{eq:final_aa}); that is
\begin{equation}
\begin{split}
    & \ln p(\mathbf{Z}|\mathbf{W}^{\M},\tau) = \\
    & = \sumn\Bigg(\unmed\ln\tau -\frac{\tau}{2}\Bigg(\Zn\ZnT -2\Zn\summ \left(\mathbf{x}_{n,:}^{(m)}\Wm^\top\right)^\top + \summ \left(\mathbf{x}_{n,:}^{(m)}\Wm^\top\right)\summ \left(\mathbf{x}_{n,:}^{(m)}\Wm^\top\right)^\top\Bigg)\Bigg) \\
    & + const. \\
    & = \sumn\Bigg(\unmed\ln\tau -\frac{\tau}{2}\Bigg(\Zn\ZnT -2\Zn\summ\left(\mathbf{x}_{n,:}^{(m)}\Wm^\top\right)^\top  + \summ\sum_{m'}^M \left(\mathbf{x}_{n,:}^{(m)}\Wm^\top\right)\left(\mathbf{x}_{n,:}^{(m')}\mathbf{W}^{(m')\top}\right)^\top\Bigg)\Bigg) \\
    & + const. \\
    & = \sumn\Bigg(\unmed\ln\tau -\frac{\tau}{2}\Bigg(\Zn\ZnT-2\Zn\summ \left(\mathbf{x}_{n,:}^{(m)}\Wm^\top\right)^\top + \summ\sum_{m'}^M\Bigg(\Xnm\Wm\mathbf{W}^{(m')\top}\mathbf{x}_{\rm n,:}^{(m')\top}\Bigg)\Bigg)\Bigg) \\
    & + const.
\end{split}
\end{equation}

To apply further expectations over the model variables, we will slightly operate over the whole equation. Thus, the final equation will look as follows
\begin{equation}
\label{eq:tau_first}
\begin{split}
    & \ln p(\mathbf{Z}|\mathbf{W}^{\M},\tau) = \\
    & = \sumn\sumk\Bigg(\unmed\ln\tau -\frac{\tau}{2}\Bigg(z_{\rm n,k}^{2}-2 z_{\rm n,k}\summ \mathbf{x}_{\rm n,:}^{(m)}\mathbf{w}_{\rm k,:}^{(m)^\top} + \summ\sum_{m'}^M\Bigg(\mathbf{w}_{\rm k,:}^{(m)}\mathbf{x}_{\rm n,:}^{(m')\top}\mathbf{x}_{\rm n,:}^{(m')}\mathbf{w}_{\rm k,:}^{(m')\top}\Bigg)\Bigg)\Bigg) \\
    & + const. \\
    & = \sumn\sumk\Bigg(\unmed\ln\tau -\frac{\tau}{2}\Bigg(z_{\rm n,k}^{2}-2 z_{\rm n,k}\summ \mathbf{x}_{\rm n,:}^{(m)}\mathbf{w}_{\rm k,:}^{(m)^\top} \\
    & + \summ\sum_{m' \neq m}^M \Bigg(\mathbf{w}_{\rm k,:}^{(m)}\mathbf{x}_{\rm n,:}^{(m)^\top}\mathbf{x}_{\rm n,:}^{(m')}\mathbf{w}_{\rm k,:}^{(m')\top}\Bigg) + \summ \Bigg(\mathbf{w}_{\rm k,:}^{(m)}\mathbf{x}_{\rm n,:}^{(m)^\top}\mathbf{x}_{\rm n,:}^{(m)}\mathbf{w}_{\rm k,:}^{(m)^\top}\Bigg)\Bigg)\Bigg)
     + \const. 
\end{split}
\end{equation}

Second term is analogous to Eq. (\ref{eq:eta_first}) and can be developed as
\begin{equation}
\begin{split}
\label{eq:tau_second}
    \ln p(\tau) = \ln \Gamma(\alpha_0^\tau,\beta_0^\tau) = -\bcerotau\taus + (\acerotau -1)\lntau + \const.
\end{split}
\end{equation}

Now, if we combine both Eq. (\ref{eq:tau_first}) and (\ref{eq:tau_second}), apply the expectation and identify terms, we can parameterize the approximated posterior distribution of $\tau$ as
\begin{equation}
    q(\tau) = \Gamma(\alpha^\tau, \beta^\tau),
\end{equation}
where
\begin{align}
    \alpha^{\tau} &= \frac{NK}{2} + \alpha_0^{\tau}\\
    \label{eq:beta_tau}
    \beta^{\tau} &= \beta_0^{\tau} + \unmed\sum_n^N\langle\mathbf{z}_{\rm n,:}\mathbf{z}_{\rm n,:}^\top\rangle - Tr\left(\langle\mathbf{Z}\rangle\sum_m^{M}\left(\mathbf{X}^{(m)}\langle\mathbf{W}^{(m)^\top}\rangle\right)^\top\right) \\ & + \unmed Tr\left(\sum_m^M\sum_{m'}^M \left(\mathbf{X}^{(m)}\langle\mathbf{W}^{(m)^\top}\rangle\right)\left(\mathbf{X}^{(m')}\langle\mathbf{W}^{(m')\top}\rangle\right)^\top\right) \nonumber
\end{align}

\subsubsection{\texorpdfstring{Mean field approximation of $\psi^\M$}{Mean field approximation of psi M}}

We can use Eq. (\ref{eq:meanfield}) for each $\psi^{(m)}$ as
\begin{equation}
\ln q(\psi^{(m)}) = \E\left[\ln p(\Xm\vert\Vm,\G,\psim) + \ln p(\psim)\right]
\end{equation}

Similarily to Eq. (\ref{eq:g_second}), we develop the first term:
\begin{align} 
&\lnp{p\p*{\Xm | \Wm,\G,\psim}} = \sumn \lnp{\N\p*{ \Gn \VmT, \p*{\psim}^{-1}\Is}} \nonumber \\
\eqeq \sumn\sumd \left(\frac{1}{2} \ln\left|\psim\right| \right.  \left. -\frac{\psim}{2}\p*{ \Xndm - \Gn \VdmT}^{2} \right)+ \const \nonumber \\
\eqeq \frac{D^{(m)} N}{2} \lnp{\psim} - \frac{\psim}{2}\sumn\sumd \left( \Xndm^2 \right. \left.- 2 \Vdm \GnT \Xndm + \p*{\Gn \VdmT}^{2}\right) + \const \nonumber\\
\eqeq \frac{D^{(m)} N}{2} \lnp{\psim} - \frac{\psim}{2} \left(\sumn\sumd \Xndm^2 \right. \nonumber \\
\eqline \hspace{-0.1cm} \left. - 2 \sumd \Vdm \GT \Xdm + \sumd \Vdm \GT \G \VdmT\right) + \const \nonumber \\
\eqeq \frac{D^{(m)} N}{2} \lnp{\psim} - \frac{\psim}{2} \left(\sumn\sumd \Xndm^2  \right. \nonumber \\
\eqline \hspace{-0.1cm} \left. - 2 \Tr\llav*{\Vm \GT \Xm} + \Tr\llav*{\Vm \GT \G \VmT}\right) + \const 
\end{align} 

The second term can be expressed as 
\begin{align} 
\E\cor*{\lnp{p\p*{\psim}}} \eqeq \lnp{p\p*{\psim}} = - \beta_0^{\psim} \psim + \p*{\alpha_0^{\psim} -1}\lnp{\psim}+ \const
\end{align} 

Joining both terms and applying expectations we obtain:
\begin{align} 
\lnp{q^*\p*{\psim}} 
\eqeq \frac{D^{(m)} N}{2} \lnp{\psim} - \frac{\psim}{2} \left(\sumn\sumd \Xndm^2 \right. \nonumber \\ 
\eqline \hspace{-0.1cm} \left. - 2 \Tr\llav*{\ang{ \Vm} \ang{ \GT} \Xm} + \Tr\llav*{\ang{ \VmT\Vm} \ang{ \GT \G}}\right) \nonumber \\ 
\eqline - \beta_0^{\tau} \psim + \p*{\alpha_0^{\tau} -1}\lnp{\psim} + \const 
\end{align} 

Thus, if we idenfity terms, we can parameterize
\begin{equation}
    q (\psim) = \Gamma\left(\alpha^{\psim},\beta^{\psim}\right),
\end{equation}
where
\begin{equation}
    \alpha^{\psim} = \frac{D^{(m)} N}{2} + \alpha_0^{\psim}
\end{equation}
and
\begin{equation}
    \beta^{\psim} = \beta_0^{\psim} + \frac{1}{2} \p*{\sumn\sumd \Xndm^2 - 2 \Tr\llav*{\ang{  \Vm}   \ang{  \GT}   \Xm} + \Tr\llav*{\ang{  \VmT\Vm}   \ang{  \GT \G}  }} \label{eq:bPsi}
\end{equation}

\subsubsection{\texorpdfstring{Mean field approximation of $\boldsymbol{\phi}^\M$}{Mean field approximation of phi M}}

We apply Eq. (\ref{eq:meanfield}) to $\boldsymbol{\phi}$ as:
\begin{equation}
\label{eq:E_delta}
    \ln q(\boldsymbol{\phi}^\M) = \E\left[\ln p(\mathbf{W}^{\M}|\boldsymbol{\phi}^\M, \boldsymbol{\gamma}^\M) + \ln p(\boldsymbol{\delta}^\M)\right].
\end{equation}

The first term is analogous to Eq. (\ref{eq:w_first}) as
\begin{equation}
\begin{split}
\label{eq:phi_first}
    & \ln p\left(\mathbf{W}^{(m)}\vert\boldsymbol{\phi}^{(m)}, \boldsymbol{\gamma}^{(m)}\right) = 
    \sum_k^K \ln \N\left(\mathbf{w}_{\rm k,:}^{(m)}\vert 0, \phi_k^{(m)-1}\Lambda_{\boldsymbol{\gamma}^{(m)}}^{-1}\right) \\
    & = \sum_k^K \left(-\unmed\lnpi + \unmed\ln(\phi_k^{(m)}\Lambda_{\boldsymbol{\gamma}^{(m)}}) - \frac{\phi_k^{(m)}}{2}\left(\mathbf{w}_{\rm k,:}\Lambda_{\boldsymbol{\gamma^{(m)}}}\mathbf{w}_{\rm k,:}^\top\right)\right) \\
    & = \sumk\sumd\left(\unmed\ln(\phi_k^{(m)}\gamma_d^{(m)})- \frac{(\phi_k^{(m)}\gamma_d^{(m)})}{2} w_{\rm k,d}^{(m)2}\right)+ \const.
\end{split}
\end{equation}

Moreover, the second term, as it follows a gamma distribution, can be expressed as
\begin{equation}
\label{eq:phi_second}
    \ln p(\boldsymbol{\phi}^{(m)}) = \Gamma(\boldsymbol{\alpha}_0^{\phim},\boldsymbol{\beta}_{0}^{\phim}) = \sumk\left(-\beta_{0 }^{\phim}\delta_k^{(m)} + (\alpha_{0 }^{\phim} -1)\ln \phi_k^{(m)} \right) + \const
\end{equation}

Now, if we include both terms in Eq. (\ref{eq:E_delta}) and apply expectations, we will obtain the final posterior approximation of each $k-th$ element of $\mathbf{\delta}^{(m)}$ as:
\begin{equation}
    q(\phi_k^{(m)}) = \Gamma(\alpha_{\rm k}^{\phim},\beta_{\rm k}^{\phim}),
\end{equation}
where
\begin{align}
    \boldsymbol{\alpha}^{\phim} & = \frac{D^{\rm (m)}}{2} + \alpha_0^{\phim}\\
    \beta_{\rm k}^{\phim} =& \beta_0^{\phim} + \unmed\sum_{\rm d}^{D^{\rm (m)}}\langle\gamma_d^{\rm (m)}\rangle\langle w_{\rm k,d}^{(m)^{2}}\rangle 
\end{align}

\subsubsection{\texorpdfstring{Mean field approximation of $\boldsymbol{\gamma}^\M$}{Mean field approximation of gamma M}}

The inference of each $\gamma_d^{(m)}$ will follow an analogous development to random variable $\boldsymbol{\phi}^{(m)}$. That is, the final approximated posterior will look as follows:
\begin{equation}
    q(\gamma_d^{(m)}) = \Gamma(\alpha_{\rm d}^{\gamma^{(m)}},\beta_{\rm d}^{\gamma^{(m)}}),
\end{equation}
where
\begin{align}
    \boldsymbol{\alpha}^{\gamma^{(m)}} &= \frac{K}{2} + \alpha_0^{\gamma^{(m)}}\\
    \beta_{\rm k}^{\gamma^{(m)}} &= \beta_0^{\gamma^{(m)}} + \unmed\sum_{\rm k}^{K}\langle\phi_k^{\rm (m)}\rangle\langle w_{\rm k,d}^{(m)^{2}}\rangle
\end{align}

\subsubsection{\texorpdfstring{Mean field approximation of $\mathbf{V}^\M$}{Mean field approximation of V M}}

We use Eq. (\ref{eq:meanfield}) for each $\mathbf{V}^{(m)}$ as
\begin{equation}
    \ln q\left(\mathbf{V}^{(m)}\right) = \E\left[\ln p\left(\mathbf{X}^{(m)}\vert\mathbf{V}^{(m)}, \G, \psi^{(m)}\right) + \ln p\left(\mathbf{V}^{(m)}\vert \boldsymbol{\delta}^{(m)}, \boldsymbol{\lambda}^{(m
    )}\right)\right].
\end{equation}

The first term can be expressed as
\begin{align} 
&\ln p\left(\mathbf{X}^{(m)}\vert\mathbf{V}^{(m)}, \G, \psi^{(m)}\right) = \nonumber \\
\eqeq - \frac{\psi^{(m)}}{2} \sumn \left(- 2 \Gn \VmT \XnmT + \Gn \VmT \Vm \GnT \right) + \const. \nonumber\\
\eqeq - \frac{\psi^{(m)}}{2} \sumn\sumd \left(- 2 \Gn \VdmT \XndmT + \Gn \VdmT \Vdm \GnT  \right) + \const.\nonumber\\
\eqeq - \frac{\psi^{(m)}}{2} \sumn\sumd \left(- 2 \XndmT \Gn \VdmT  + \Vdm \GnT \Gn \VdmT \right)+ \const. \nonumber.
\label{eq:v_first}
\end{align} 

The second term can be expressed as
\begin{equation}
\ln p\left(\mathbf{V}^{(m)}\vert \boldsymbol{\delta}^{(m)}, \boldsymbol{\lambda}^{(m
    )}\right)
=  \sumd \sum_s^S \p*{ \frac{1}{2}\lnp{\delta_s^{(m)} \lambda_d^{(m)}} - \frac{1}{2} \delta_s^{(m)}\lambda_d^{(m)} \text{v}_{\rm d,s}^{(m)}}+ \const.
\label{eq:v_second}
\end{equation} 

If we put both terms together and apply expectations, we get
\begin{equation}
\begin{split}
    &\ln q\left(\mathbf{V}^{(m)}\right) = \sumd \sum_s^S \p*{ - \frac{1}{2} \ang{\lambda_d^{(m)}}\ang{\delta_s^{(m)}}\langle\text{v}_{\rm d,s}^{(m)}\rangle} + \sumn\sumd \Big( \ang{\psim}  \\
    & \Xndm \ang{\Gn}\VdmT - \frac{\ang{\psim}}{2} \sumd \p*{\Vdm \ang{\GnT \Gn}\VdmT}\Big) + \const \\
    &= \sumd( - \frac{1}{2} \Vdm \p*{\Lambda_{\langle\boldsymbol{\delta}^{(m)}\rangle}\ang{\lambda_d^{(m)}} + \ang{\psim}\ang{\GnT \Gn}}\VdmT \nonumber\\
    & + \ang{\psim} \sumn\p*{\Xndm \ang{\Gn} }\VdmT + \const.
\end{split}
\end{equation}

Now, if we join both terms and apply expectation over them, we can identify terms and parameterize
\begin{equation}
    q\left(\mathbf{v}_{d ,:}^{(m)}\right) = \N(\mathbf{v}_{d ,:}^{(m)}|\langle \mathbf{v}_{d ,:}^{(m)}\rangle,\Sigma_{\mathbf{v}_{d ,:}}^{\rm (m)}),
\end{equation}
where
\begin{align}
    \Sigma_{\mathbf{v}_{d ,:}}^{\rm (m)-1} & = \langle\lambda_{d }^{(m)}\rangle\Lambda_{\langle\boldsymbol{\delta}^{(m)}\rangle} + \langle\psi^{(m)}\rangle\langle\mathbf{G}^\top \mathbf{G}\rangle \\
    \langle \mathbf{v}_{d ,:}^{(m)}\rangle & = \langle\psi^{(m)}\rangle\mathbf{x}_{\rm :, d }^{(m)^\top}\langle\mathbf{G}\rangle\Sigma_{\mathbf{v}_{d ,:}}^{\rm (m)}
\end{align}

\subsubsection{\texorpdfstring{Mean field inference of $\boldsymbol{\delta}^\M$}{Mean field inference of delta M}}

The update of each $\boldsymbol{\delta}^{(m)}$ follows the expression
\begin{equation}
\label{eq:delta}
    \ln q\left(\boldsymbol{\delta}^{(m)}\right) = \E\left[\ln p\left(\mathbf{V}^{(m)}\vert\boldsymbol{\delta}^{(m)}\boldsymbol{\lambda}^{(m)}\right) + \ln p\left(\boldsymbol{\delta}^{(m)}\right)\right].
\end{equation}

The first term can be expressed as
\begin{equation}
\label{eq:delta_first}
    \ln p\left(\mathbf{V}^{(m)}\vert\boldsymbol{\delta}^{(m)}\boldsymbol{\lambda}^{(m)}\right) = \sumd \sum_s^S \p*{ \frac{1}{2}\lnp{\delta_s^{(m)} \lambda_d^{(m)}} - \frac{1}{2} \delta_s^{(m)}\lambda_d^{(m)} \text{v}_{\rm d,s}^{(m)}}+ \const,
\end{equation}
and the second as
\begin{equation}
\label{eq:delta_second}
    \ln p\left(\boldsymbol{\delta}^{(m)}\right) = \sum_s^S\left(-\beta_{0}^{\deltam}\delta_s^{(m)} + (\alpha_{0 }^{\deltam} -1)\ln \delta_s^{(m)} \right).
\end{equation}

Now, we can merge both terms of Eq. (\ref{eq:delta_first}) and (\ref{eq:delta_second}) in Eq. (\ref{eq:delta}) and apply expectations to parameterize
\begin{equation}
    q(\delta_s^{(m)}) = \Gamma(\alpha_{\rm s}^{\deltam},\beta_{\rm s}^{\deltam}),
\end{equation}
where
\begin{align}
    \alpha_{\rm s}^{\deltam} &= \frac{D^{(m)}}{2} + \alpha_0^{\deltam} \\
    \beta_{\rm s}^{\deltam} & = \beta_0^{\deltam} + \frac{1}{2}\sumd\left(\langle\lambda_d\rangle\langle\text{v}_{\rm d,s}^{(m)2}\rangle\right) \label{eq:deltaBeta}
\end{align}

\subsubsection{\texorpdfstring{Mean field approximation of $\boldsymbol{\lambda}^\M$}{Mean field approximation of lambda M}}

The update closely follows the derivation of the $q(\boldsymbol{\delta}^{(m)})$ update. From
\begin{equation}
\label{eq:lambda}
    \ln q\left(\boldsymbol{\lambda}^{(m)}\right) = \E\left[\ln p\left(\mathbf{V}^{(m)}\vert\boldsymbol{\delta}^{(m)}\boldsymbol{\lambda}^{(m)}\right) + \ln p\left(\boldsymbol{\delta}^{(m)}\right)\right],
\end{equation}
we can easily obtain that
\begin{equation}
    q(\lambda_d^{(m)}) = \Gamma(\alpha_{\rm d}^{\lambda^{(m)}},\beta_{\rm d}^{\lambda^{(m)}}),
\end{equation}
where
\begin{align}
    \alpha_{\rm s}^{\lambda^{(m)}} &= \frac{S}{2} + \alpha_0^{\lambda^{(m)}} \\
    \beta_{\rm s}^{\lambda^{(m)}} & = \beta_0^{\lambda^{(m)}} + \frac{1}{2}\sum_s^S\left(\langle\delta_s\rangle\langle\text{v}_{\rm d,s}^{(m)2}\rangle\right)
\end{align}

\subsubsection{\texorpdfstring{Mean field approximation of $\mathbf{Y}$}{Mean field approximation of Y}}

To parameterize the posterior distribution of the soft output view, i.e., $q\left(\mathbf{Y}\right)$, we have to use Eq. (\ref{eq:meanfield}), obtaining
\begin{equation}
\label{eq:E_y_ant}
    \ln q(\mathbf{Y}) = \E\left[\ln p(\mathbf{T}|\mathbf{Y}) + \ln p(\mathbf{Y}|\mathbf{Z}, \U, \G,\mathbf{V}^{\mathbf{Y}},\eta)\right].
\end{equation}

However, Eq. (\ref{eq:meanfield}) requires the distributions to be conjugated, so we have to slightly modify $p(\mathbf{T}|\mathbf{Y})$. To do so, as presented in \cite{jaakkola2000bayesian}, we lower-bound this distribution based on a first-order Taylor series expansion as
\begin{equation}
    \ln p(\mathbf{t}_{\rm n,:} = 1|\mathbf{y}_{\rm n,:}) = 
    e^{\mathbf{y}_{\rm n,:}}\sigma(-\mathbf{y}_{\rm n,:}) \geqslant h(\mathbf{y}_{\rm n,:}, \xi_{\rm n,:}) = 
    e^{\mathbf{y}_{\rm n,:} \mathbf{t}_{\rm n,:}}\sigma(\xi_{\rm n,:})e^{-\frac{\mathbf{y}_{\rm n,:} + \xi_{\rm n,:}}{2} - \lambda(\xi_{\rm n,:})(\mathbf{y}_{\rm n,:}^2 - \xi_{\rm n,:}^2)},
\end{equation}

Hence, we can rewrite Eq. (\ref{eq:E_y_ant}) as
\begin{equation}
\label{eq:E_y}
    \ln q(\mathbf{Y}) = \E\left[\ln h(\mathbf{y}_{\rm n,:}, \xi_{\rm n,:}) + \ln p(\mathbf{Y}|\mathbf{Z}, \U, \G,\mathbf{V}^{\mathbf{Y}},\eta)\right].
\end{equation}

Thus, we can develop the first term as
\begin{align}
\label{eq:y_first}
&\E\cor*{\lnp{h\p*{\Xm,\bm{\xi}}}} = \\&= \E\cor*{\sumn\sumd\p*{\lnp{\sigma\p*{\xi_{n,d}}}+\Xndm \tndm - \frac{1}{2}\p*{\Xndm + \xi_{n,d}}- \lambda\p*{\xi_{n,d}}\p*{\Xndm^2 - {\xi_{n,d}}^2}}} \nonumber\\
&= \sumn\sumd\p*{\Xndm \tndm - \frac{1}{2}\Xndm - \lambda\p*{\xi_{n,d}}\Xndm^2} +\const \nonumber\\
&= \sumn\p*{\p*{\tnm  - \frac{1}{2}} \XnmT - \Xnm\Lambda_{\bm{\xi}_{n,:}}\XnmT} +\const\nonumber ,
\end{align}
being $\lambda(a) = \frac{1}{2a}(\sigma(a) - \unmed)$ and $\xi_{\rm n,c}$ the center of the Taylor series around $y_{\rm n,c}$. Moreover, $\bm{\xi}_{n,:}$ is a learnable parameter that will be adjusted during the model training through variational inference.

Moreover, the second element can be developed as
\begin{equation}
\begin{split}
\label{eq:y_second}
    &\ln p(\mathbf{Y}|\mathbf{Z}, \mathbf{U},\mathbf{G}, \mathbf{V}^{\mathbf{Y}},\eta)  = \sum_n^N \ln \N\left(\Zn\UT + \Gn\mathbf{V}^{\mathbf{Y}\top}, \eta^{-1}\Ic 
    \right) \\
    & =\sum_n^N\left(-\frac{1}{2}\ln(2\pi) + \frac{1}{2}\ln(\eta) - \frac{\eta}{2}\left(\Yn - \Zn\UT - \Gn\mathbf{V}^{\mathbf{Y}\top}\right)\left(\Yn - \Zn\UT - \Gn\mathbf{V}^{\mathbf{Y}\top}\right)^\top\right)\\
    & = \sum_n^N-\frac{\eta}{2}\Big( \Yn\Yn^\top -2\Yn\U\Zn^\top - 2\Yn\mathbf{V}^{\mathbf{Y}}\Gn^\top\Big) + \const.
\end{split}
\end{equation}

Now, we can join both Eq. (\ref{eq:y_first}) and (\ref{eq:y_second}) into Eq. (\ref{eq:E_y}) and apply expectations to parameterize the final distribution
\begin{equation}
    q(\Yn) = \N(\langle \mathbf{y}_{\rm n,:}\rangle,\Sigma_{\mathbf{y}_{\rm n,:}}),
\end{equation}
where
\begin{align}
    \Sigma_{\mathbf{y}_{\rm n,:}}^{-1} & = \langle\psi\rangle\Ic + 2\Lambda_{\xi_{\rm n,:}} \\
    \langle\mathbf{y}_{\rm n,:}\rangle &= \left(\mathbf{t}_{\rm n,:} - \unmed + \langle\eta\rangle\left(\langle\mathbf{z}_{\rm n,:}\rangle\langle\mathbf{U}\rangle^\top + \langle\mathbf{g}_{\rm n,:}\rangle\langle\mathbf{V}^{\mathbf{Y}}\rangle^\top\right)\right)\Sigma_{\mathbf{y}_{\rm n,:}}.
\end{align}

\subsubsection{\texorpdfstring{Variational parameter calculation of $\boldsymbol{\xi}$}{Variational parameter calculation of xi}}

To optimize $\bm{\xi}$, we need to calculate the gradient of the lower bound $L(q)$ with respect to $\bm{\xi}$ and set it to zero. Thus, the only terms that depend on $\bm{\xi}$ are
\begin{align}
 L_{\boldsymbol{\xi}} = \E_{q}\cor*{\lnp{h\p*{\Xm,\bm{\xi}}}} + \E_{q}\cor*{\lnp{q\p*{\Xm}}},
\end{align}
where
\begin{align}
\E_{q}\cor*{\lnp{p\p*{\tm|\Xm}}} 
\eqeq \E_{q}\cor*{\lnp{h\p*{\Xm,\bm{\xi}}}} \nonumber\\
\eqeq \sumn\sumd \left(\lnp{\sigma\p*{\xi_{n,d}}} + \ang{\Xndm} \tndm - \frac{1}{2}\p*{\ang{\Xndm} + \xi_{n,d}} \right. \nonumber\\
\E_{q}\cor*{\lnp{q\p*{\Xm}}} 
\eqeq  \sumn\p*{\frac{D^{(m)}}{2} \lnp{2\pi e} + \frac{1}{2} \ln|\Sigma_{\Xnm}|} \label{eq:ElogpXtwo}
\end{align}

We can now set the derivative with respect to the parameter $\xi_{n,d}$ equal to zero:
\begin{align}
\frac{\partial L_{\boldsymbol{\xi}}}{\partial \xi_{n,d}} \eqeq \lambda'\p*{\xi_{n,d}}\p*{\E\cor*{\Xndm^2} - {\xi_{n,d}}^2} = 0
\end{align}
where $\lambda \p*{a} = \frac{1}{2a}\p*{\sigma\p*{a} - \frac{1}{2}}$ is a monotonic function of $\xi_{n,d}$ for $\xi_{n,d} \geq 0$. Hence, if we ignore the $\xi_{n,d}=0$ solution, we have
\begin{align}
\lambda'\p*{\xi_{n,d}} \neq 0 \longrightarrow {\xi_{n,d}^{new}}^2 = \E\cor*{\Xndm^2}  = \ang{\Xnm}\ang{\Xnm}^\top + \Sigma_{\Xnm}
\end{align}

\newpage

\subsubsection{Summary results}

Tables \ref{tab:q_dist_main_disc} and \ref{tab:q_dist_main_gen} present the update rules for the r.v. related to the discriminative and generative parts, respectively. In the following subsections, we present the mathematical derivations leading to these approximated posterior distributions for each model's random variable.

\begin{table*}[ht!]
\caption{$q^{*}$ update rules for variables relative to the \textbf{discriminative} part of the model obtained using the mean-field approximation. 
}
    \begin{adjustbox}{max width=\textwidth}
    \renewcommand{\arraystretch}{2}
        \centering
        \setlength{\tabcolsep}{1pt}
        \begin{tabular}{c c c}
        \toprule
        {\bf Variable} & $\mathbf{q^{*}}$ {\bf distribution} & {\bf Parameters}\\
        
        \midrule
        $\mathbf{z}_{\rm n,:}$ & $\N(\mathbf{z}_{\rm n,:}|\langle \mathbf{z}_{\rm n,:}\rangle,\Sigma_{\mathbf{Z}})$ & \begin{tabular}{@{}c@{}}$\Sigma_{\mathbf{Z}}^{-1} = \langle\tau\rangle \Ik + \langle\eta\rangle\langle\mathbf{U}^\top\mathbf{U}\rangle$
        
        \\ $\langle\mathbf{z}_{\rm n,:}\rangle = \left(\langle\tau\rangle\sum_m^M\mathbf{x}_{\rm n,:}^{(m)}\langle\mathbf{W}^{(m)}\rangle^\top  +\langle \eta \rangle \left(\langle\mathbf{y}_{\rm n,:}\rangle - \langle\mathbf{g}_{\rm n,:} \rangle \langle\mathbf{V}^{\mathbf{Y} }  \rangle^\top \right)\langle\mathbf{U}\rangle \right)\Sigma_{\mathbf{Z}}$\end{tabular}\\

        \midrule
        $\Wkm$ & $\N(\Wkm|\langle \Wkm\rangle,\Sigma_{\Wk}^{\rm (m)})$ & \begin{tabular}{@{}c@{}}$\Sigma_{\Wk}^{(m)-1} = \tauE \XmT\Xm + \langle\phi_k^{(m)} \rangle \Lambda_{\langle\boldsymbol{\gamma}^{(m)}\rangle}$
        
        \\ $\langle \Wkm\rangle = \tauE\sum_n^N \left[\langle z_{\rm n, k}\rangle - \sum_{m' \neq m}^{M}\langle\mathbf{w}_{\rm k,:}^{(m')}\rangle \mathbf{x}^{(m') \top}_{\rm n,:}\right] \mathbf{x}_{\rm n,:}^{(m)}\Sigma_{\Wk}^{\rm (m)}$\end{tabular}\\

        \midrule
        $\mathbf{u}_{\rm c,:}$ & $\N(\mathbf{u}_{\rm c,:}|\langle \mathbf{u}_{\rm c,:}\rangle,\Sigma_{\mathbf{u}_{\rm c,:}})$ & \begin{tabular}{@{}c@{}}$\Sigma_{\mathbf{u}_{\rm c,:}}^{-1} = \Ik + \langle\eta\rangle\langle\ZT\Z\rangle$
        
        \\ $\langle\mathbf{u}_{\rm c,:}\rangle = \left[\langle\eta\rangle\langle\mathbf{y}_{\rm :,c}^\top\rangle\langle\Z\rangle - \langle\eta\rangle\langle\mathbf{v}_{\rm c,:}^{\mathbf{Y}}\rangle\langle\GT\rangle\langle\Z\rangle\right]\Sigma_{\mathbf{u}_{\rm c,:}}$\end{tabular}\\
        
        \midrule
        $\eta$ & $\Gamma(\alpha^\eta, \beta^\eta)$ & \begin{tabular}{@{}c@{}}$\alpha^\eta  = \frac{NC}{2} + \alpha^\eta_0$
        
        \\ $\begin{aligned}
\beta^\eta = \beta_0^\eta + \frac{1}{2}\sum_n^N\Big(
&\langle\Yn\Yn^\top\rangle -2\langle\Yn\rangle\langle\U\rangle\langle\Zn^\top\rangle 
-2 \langle\Yn\rangle\langle\mathbf{V}^{\mathbf{Y}}\rangle\langle\Gn^\top\rangle \\
&+ \operatorname{Trace}\left(\langle\UT\U\rangle\langle\Zn^\top\Zn\rangle\right) 
+ \operatorname{Trace}\left(\langle\mathbf{V}^{\mathbf{Y}\top}\mathbf{V}^{\mathbf{Y}}\rangle\langle\Gn^\top\Gn\rangle\right) \\
&+ 2\langle\Zn\rangle\langle\UT\rangle\langle\mathbf{V}^{\mathbf{Y}}\rangle\langle\Gn^\top\rangle
\Big)
\end{aligned}$
     \end{tabular}\\

     \midrule
        $\tau$ & $\Gamma(\alpha^\tau, \beta^\tau)$ & \begin{tabular}{@{}c@{}}$\alpha^{\tau} = \frac{NK}{2} + \alpha_0^{\tau}$
        
        \\ $\beta^{\tau} = \beta_0^{\tau} + \unmed\sum_n^N\langle\mathbf{z}_{\rm n,:}\mathbf{z}_{\rm n,:}^\top\rangle - Tr\left(\langle\mathbf{Z}\rangle\sum_m^{M}\left(\mathbf{X}^{(m)}\mathbf{W}^{(m)^\top}\right)^\top\right) + \unmed Tr\left(\sum_m^M\sum_{m'}^M \left(\mathbf{X}^{(m)}\mathbf{W}^{(m)^\top}\right)\left(\mathbf{X}^{(m')}\mathbf{W}^{(m')\top}\right)^\top\right)$
        \end{tabular}\\

        \midrule
        $\phi_k^{(m)}$ & $\Gamma(\alpha_{\rm k}^{\phi(m)},\beta_{\rm k}^{\phi(m)})$ & \begin{tabular}{@{}c@{}}$\boldsymbol{\alpha}^{\phi(m)} = \frac{D^{\rm (m)}}{2} + \alpha_0^{\phi(m)}$
        
        \\ $\beta_{\rm k}^{\delta(m)} = \beta_0^{\phi(m)} + \unmed\sum_{\rm d}^{D^{\rm (m)}}\langle\gamma_d^{\rm (m)}\rangle\langle w_{\rm k,d}^{(m)^{2}}\rangle $\end{tabular}\\

        \midrule
        $\gamma_d^{(m)}$ & $\Gamma(\alpha_{\rm d}^{\gamma(m)},\beta_{\rm d}^{\gamma(m)})$ & \begin{tabular}{@{}c@{}}$ \boldsymbol{\alpha}^{\gamma(m)} = \frac{K}{2} + \alpha_0^{\gamma(m)}$
        
        \\ $\beta_{\rm k}^{\delta(m)} = \beta_0^{\delta(m)} + \unmed\sum_{\rm k}^{K}\langle\phi_k^{\rm (m)}\rangle\langle w_{\rm k,d}^{(m)^{2}}\rangle$\end{tabular}\\

        \midrule
        $\mathbf{y}_{\rm n,:}$ & $\N(\mathbf{y}_{\rm n,:}|\langle \mathbf{y}_{\rm n,:}\rangle,\Sigma_{\mathbf{y}_{\rm n,:}})$ & \begin{tabular}{@{}c@{}c@{}c@{}}$\Sigma_{\mathbf{y}_{\rm n,:}}^{-1} = \langle\eta\rangle\Ic + 2\Lambda_{\boldsymbol{\xi}_{\rm n,:}}$ ~~ where ~~ $\boldsymbol{\xi}_{\rm n,:} = \sqrt{\langle\mathbf{y}_{\rm n,:}\rangle^2 + diag\left(\Sigma_{\mathbf{y}_{\rm n,:}}\right)}$  \\ 
        
        $\langle\mathbf{y}_{\rm n,:}\rangle = \left(\mathbf{t}_{\rm n,:} - \unmed + \langle\eta\rangle\left(\langle\mathbf{z}_{\rm n,:}\rangle\langle\mathbf{U}\rangle^\top + \langle\mathbf{g}_{\rm n,:}\rangle\langle\mathbf{V}^{\mathbf{Y}}\rangle^\top\right)\right)\Sigma_{\mathbf{y}_{\rm n,:}}$ \end{tabular}\\
        \bottomrule

        \end{tabular}
    \end{adjustbox}
    \label{tab:q_dist_main_disc}
\end{table*}

\begin{table*}[ht!]
\caption{$q^{*}$ update rules for variables relative to the \textbf{generative} part of the model obtained using the mean-field approximation. 
}
    \begin{adjustbox}{max width=\textwidth}
    \renewcommand{\arraystretch}{2}
        \centering
        \setlength{\tabcolsep}{1pt}
        \begin{tabular}{c c c}
        \toprule
        {\bf Variable} & $\mathbf{q^{*}}$ {\bf distribution} & {\bf Parameters}\\
        
        \midrule
        $\mathbf{g}_{\rm n,:}$ & $\N(\mathbf{g}_{\rm n,:}|\langle \mathbf{g}_{\rm n,:}\rangle,\Sigma_{\mathbf{G}})$ & \begin{tabular}{@{}c@{}}$\Sigma_{\mathbf{G}}^{-1} = \Is + \langle\eta\rangle\langle\mathbf{V}^{\mathbf{Y}\top} \mathbf{V}^{\mathbf{Y}}\rangle + \sum_m^{M}\langle\psi^{(m)}\rangle\langle\mathbf{V}^{(m)\top}\mathbf{V}^{(m)}\rangle$
        
        \\ $\langle\mathbf{g}_{\rm n,:}\rangle = \left( \sum_m^M\left(\langle\psi^{(m)}\rangle\mathbf{x}_{\rm n,:}^{(m)}\langle\mathbf{V}^{(m)}\rangle\right) + \langle\eta\rangle   \left( \langle\mathbf{y}_{n,:}\rangle -\langle\mathbf{z}_{\rm n,:}\rangle\langle\mathbf{U}\rangle^\top \right)\langle\mathbf{V}^{\mathbf{Y}}\rangle \right)\Sigma_{\mathbf{G}}$\end{tabular}\\

        \midrule
        $\mathbf{v}_{\rm c,:}^{\mathbf{Y}}$ & $\N(\mathbf{v}_{\rm c,:}^{\mathbf{Y}}|\langle \mathbf{v}_{\rm c,:}^{\mathbf{Y}}\rangle,\Sigma_{\mathbf{v}_{\rm c,:}^{\mathbf{Y}}})$ & \begin{tabular}{@{}c@{}}$\Sigma_{\mathbf{v}_{\rm c,:}^{\mathbf{Y}}}^{-1}  = \langle\eta\rangle \langle\GT\G\rangle + \langle\gamma_c^{(M+1)}\rangle\Lambda_{\langle\boldsymbol{\phi}\rangle}$
        
        \\ $\langle\mathbf{v}_{\rm c,:}^{\mathbf{Y}}\rangle  = \langle\eta\rangle\sum_n^N\langle\Gn\rangle\langle y_{\rm n,c}\rangle - \langle\eta\rangle\sum_n^N\langle\Zn\rangle\langle\mathbf{u}_{\rm c,:}^\top\rangle\langle\Gn\rangle$\end{tabular}\\
        
        \midrule
        $\mathbf{v}_{d ,:}^{(m)}$ & $\N(\mathbf{v}_{d ,:}^{(m)}|\langle \mathbf{v}_{d ,:}^{(m)}\rangle,\Sigma_{\mathbf{v}_{d ,:}}^{\rm (m)})$ & \begin{tabular}{@{}c@{}}$ \Sigma_{\mathbf{v}_{d ,:}}^{\rm (m)-1} = \langle\lambda_{d }^{(m)}\rangle\Lambda_{\langle\boldsymbol{\delta}^{(m)}\rangle} + \langle\psi^{(m)}\rangle\langle\mathbf{G}^\top \mathbf{G}\rangle$
        
        \\ $\langle \mathbf{v}_{d ,:}^{(m)}\rangle = \langle\psi^{(m)}\rangle\mathbf{x}_{\rm :, d }^{(m)\top}\langle\mathbf{G}\rangle\Sigma_{\mathbf{v}_{d ,:}}^{\rm (m)}$\end{tabular}\\

        \midrule
        $\lambda_d^{(m)}$ & $\Gamma(\alpha_{\rm d}^{\lambda(m)},\beta_{\rm d}^{\lambda(m)})$ & \begin{tabular}{@{}c@{}}$\alpha_{\rm s}^{\lambda(m)} = \frac{S}{2} + \alpha_0^{\lambda(m)}$
        
        \\ $\beta_{\rm s}^{\lambda(m)}  = \beta_0^{\lambda(m)} + \frac{1}{2}\sum_s^S\left(\langle\delta_s\rangle\langle\text{v}_{\rm d,s}^{(m)2}\rangle\right)$\end{tabular}\\

        \midrule
        $\delta_s^{(m)}$ & $\Gamma(\alpha_{\rm s}^{\delta(m)},\beta_{\rm s}^{\delta(m)})$ & \begin{tabular}{@{}c@{}}$\alpha_{\rm s}^{\delta(m)} = \frac{D^{(m)}}{2} + \alpha_0^{\delta(m)}$
        
        \\ $\beta_{\rm s}^{\delta(m)}  = \beta_0^{\delta(m)} + \frac{1}{2}\sumd\left(\langle\lambda_d\rangle\langle\text{v}_{\rm d,s}^{(m)2}\rangle\right)$\end{tabular}\\

        \midrule
        $\psim$ & $\Gamma\left(\alpha^{\psim},\beta^{\psim}\right)$ & \begin{tabular}{@{}c@{}}$\alpha^{\psim} = \frac{D_m N}{2} + \alpha_0^{\psim}$
        
        \\ $\beta_{\psim} = \beta_0^{\psim} + \frac{1}{2} \p*{\sumn\sumd \Xndm^2 - 2 \Tr\llav*{\ang{  \Vm}   \ang{  \GT}   \Xm} + \Tr\llav*{\ang{  \VmT\Vm}   \ang{  \GT \G}  }}$\end{tabular}\\

        \midrule
        $\mathbf{y}_{\rm n,:}$ & $\N(\mathbf{y}_{\rm n,:}|\langle \mathbf{y}_{\rm n,:}\rangle,\Sigma_{\mathbf{y}_{\rm n,:}})$ & \begin{tabular}{@{}c@{}c@{}c@{}}$\Sigma_{\mathbf{y}_{\rm n,:}}^{-1} = \langle\eta\rangle\Ic + 2\Lambda_{\boldsymbol{\xi}_{\rm n,:}}$ ~~ where ~~ $\boldsymbol{\xi}_{\rm n,:} = \sqrt{\langle\mathbf{y}_{\rm n,:}\rangle^2 + diag\left(\Sigma_{\mathbf{y}_{\rm n,:}}\right)}$  \\ 
        
        $\langle\mathbf{y}_{\rm n,:}\rangle = \left(\mathbf{t}_{\rm n,:} - \unmed + \langle\eta\rangle\left(\langle\mathbf{z}_{\rm n,:}\rangle\langle\mathbf{U}\rangle^\top + \langle\mathbf{g}_{\rm n,:}\rangle\langle\mathbf{V}^{\mathbf{Y}}\rangle^\top\right)\right)\Sigma_{\mathbf{y}_{\rm n,:}}$ \end{tabular}\\
        \bottomrule

        \end{tabular}
    \end{adjustbox}
    \label{tab:q_dist_main_gen}
\end{table*}

\newpage

\subsection{Evidence Lower Bound}

To measure the quality of the approximated posterior distributions $q(\boldsymbol{\Theta})$ of our model, the mean-field inference approximation procedure maximizes a lower bound $L(q)$, which is proportional to the Kullback-Leibler (KL) divergence between the approximated posteriors $q(\boldsymbol{\Theta})$ and the true posterior $p(\boldsymbol{\Theta}\vert\mathbf{X}^\M,\mathbf{t})$. We can derive the expression of $L(q)$ as
\begin{equation}
\begin{split}
    L(q) & = \int q(\bm{\Theta}) \ln\left( \frac{p(\bm{\Theta},\mathbf{t},\X^\M)}{q(\bm{\Theta})}\right) d\bm{\Theta} = \int \prod_{i}q_i\left[\ln(p(\bm{\Theta},\mathbf{t},\X^\M)) - \sum_i \ln(q_i)\right]d\bm{\Theta} \\
    & = \int \prod_i q_i \ln(p(\bm{\Theta},\mathbf{t},\X^\M))d\bm{\Theta} - \int \prod_i q_i \sum_i \ln(q_i)d\bm{\Theta} \\
    & = \int q_j \prod_{i \neq j} q_i \ln(p(\bm{\Theta},\mathbf{t},\X^\M)) d\bm{\Theta} - \int q_j \prod_{i \neq j} q_i \left(\ln(q_j) + \sum_{i \neq j}\ln(q_i)\right)d\bm{\Theta} \\
    & = \int q_j \prod_{i \neq j} q_i \ln(p(\bm{\Theta},\mathbf{t},\X^\M)) d\bm{\Theta} - \int q_j \prod_{i \neq j} q_i \sum_{i \neq j}\ln(q_i) d\bm{\Theta} - \int q_j \prod_{i \neq j} q_i \ln(q_j)d\bm{\Theta}
    \\& = \int q_j \left[\int \prod_{i \neq j} q_i \ln p(\mathbf{t},\X^\M,\bm{\Theta})d\bm{\Theta}_i \right]d\bm{\Theta}_j - \int q_j \ln q_jd\bm{\Theta}_j + \const \\
    & = \int q_j\ln(f_j)d\bm{\Theta}_j - \int q_j\ln(q_j)d\bm{\Theta}_j + \const.
    \label{eq:MFA}
\end{split}
\end{equation}
where $\ln (f_j) = \mathbb{E}_{-q_j}[\ln p(\textbf{t},\textbf{X},\bm{\Theta})]$, and $\mathbb{E}_{-q_j}$ means the expectation over all the r.v. in $q$ except the $j$-th one. Hence, we can express the $L(q)$ as:
\begin{equation}
\label{eq:lp_tot}
    L(q) = - \E_{q}\left[\ln (q(\bm{\Theta}))\right] + \E_{q}\left[\ln (p(\bm{\Theta}, \mathbf{t}, \X^{\M}))\right],
\end{equation}
where $\E_{q}\left[\ln (p(\bm{\Theta}, \mathbf{t}, \mathbf{X}^{\M}))\right]$ is the expectation of the joint posterior distribution w.r.t. $q(\mathbf{\Theta})$ and $\E_{q}[\ln 
 q(\bm{\Theta})]$ the entropy of $q(\mathbf{\Theta})$.

\subsubsection{\texorpdfstring
  {Terms associated with $E_{q}\left[\ln (p(\boldsymbol{\Theta}, \mathbf{t}, \mathbf{X}^{\M}))\right]$}
  {Terms associated with Eqn (log p(Theta, t, X M))}}

This term comprises the following elements:
\begin{equation}
\begin{split}
\label{eq:full_elbo}
    &\E_{q}\left[\ln (p(\bm{\Theta}, \mathbf{t}, \mathbf{X}^{\M}))\right] = \E\Big[\ln p(\Y\vert\Z,\mathbf{U},\G,\V^{\boldsymbol{Y}},\eta) + \ln p(\mathbf{T}\vert\mathbf{Y}) + \ln p(\mathbf{V}^{\boldsymbol{Y}}\vert\boldsymbol{\delta}^{(M+1)},\boldsymbol{\lambda}^{(M+1)}) \\ &+ \ln p(\boldsymbol{\delta}^{(M+1)}) + \ln p(\boldsymbol{\lambda}^{(M+1)}) 
    + \ln p(\mathbf{Z}|\mathbf{W}^{\M},\tau) + \ln p(\G)+ \ln p(\U) + \ln p(\eta) + \ln p(\tau) \\ &+ \summ\Big(\ln p(\mathbf{W}^{(m)}|\boldsymbol{\phi}^{(m)}, \boldsymbol{\gamma}^{(m)}) +\ln p(\mathbf{V}^{(m)}|\boldsymbol{\delta}^{(m)}, \boldsymbol{\lambda}^{(m)})
    + \ln p(\boldsymbol{\delta}^{(m)}) + \ln p(\boldsymbol{\gamma}^{(m)}) \\
    & + \ln p(\boldsymbol{\lambda}^{(m)}) + \ln p(\boldsymbol{\phi}^{(m)})\Big) \Big]
\end{split}
\end{equation}
 
Thus, to analyze Eq. (\ref{eq:full_elbo}), we will separately develop each term and subsequently apply the expectation over all the variables.

Let's start by calculating the terms related to the generative space. The latent generative variable can be calculated as
\begin{align} 
\label{eq:G_q}
\E_{q}\cor*{\lnp{p\p*{\G}}} 
\eqeq \sumn \E_{q}\cor*{-\frac{S}{2}\lnp{2\pi}-\frac{1}{2}\Gn \GnT}
= -\frac{N S}{2}\lnp{2\pi}-\frac{1}{2} \Tr\{\ang{\G \GT}\}.
\end{align} 

Next, the term corresponding to the projection matrix $\Vm$ can be calculated as
\begin{align} 
\label{eq:V_q}
&\E_{q}\cor*{\lnp{p\p*{\Vm|\dm, \lm}}} \nonumber \\
&= \sums\sumd \E_{q}\cor*{-\frac{1}{2}\lnp{2\pi}+\frac{1}{2}\lnp{\dsm}+\frac{1}{2}\lnp{\ldm}-\frac{1}{2} \Vdsm  \ldm \dsm \Vdsm } \nonumber \\
&= -\frac{S D^{(m)}}{2}\lnp{2\pi}+\frac{D^{(m)}}{2}\sums \E_{q}\cor*{\lnp{\dsm}}+\frac{S}{2}\sumd \E_{q}\cor*{\lnp{\ldm}} \nonumber \\
&= - \frac{1}{2} \sumk\sumd\p*{\E_{q}\cor*{ \ldm}\E_{q}\cor*{ \dsm} \E_{q}\cor*{\Vdsm  \Vdsm }},
\end{align} 
where we the expectation of the different variables is:
\begin{align} 
\E_{q(\delta)}\cor*{\ln({\dsm})} 
\eqeq \psi\p*{\alpha_s^{\dm}} - \lnp{\beta_s^{\dm}} \label{eq:ElogDelta}
\end{align} 
\begin{align} 
\E_{q(\lambda)}\cor*{\ln({\ldm})} 
\eqeq \psi\p*{\alpha_d^{\lm}} - \lnp{\beta_d^{\lm}} \label{eq:ElogLambda}
\end{align} 
\begin{align} 
\E_{q(\delta)}\cor*{\dsm}
\eqeq \frac{\alpha_s^{\dm}}{\beta_s^{\dm}} \label{eq:EDelta}
\end{align} 
\begin{align} 
\E_{q(\lambda)}\cor*{\ldm}
\eqeq \frac{\alpha_d^{\lm}}{\beta_d^{\lm}} \label{eq:ELambda}
\end{align} 
\begin{align} 
\E_{q(V)}\cor*{\Vdsm \Vdsm }
\eqeq \ang{\Vdsm , \Vdsm}
\end{align} 

Moreover, the development of $\mathbf{V}^{\mathbf{Y}}$ will be analogous to Eq. (\ref{eq:V_q}) as
\begin{align} 
\E_{q}\cor*{\lnp{p\p*{\mathbf{V}^{\mathbf{Y}}|\dm, \lm}}} 
\eqeq -\frac{S C}{2}\lnp{2\pi}+\frac{C}{2}\sums \E_{q}\cor*{\lnp{\delta_s^{(M+1)}}}+\frac{S}{2}\sum_c^C \E_{q}\cor*{\lnp{\lambda_c^{(M+1)}}} \nonumber \\
\eqline - \frac{1}{2} \sum_s^S\sum_c^C\p*{\E_{q}\cor*{ \lambda_c^{(M+1)}}\E_{q}\cor*{ \delta_s^{(M+1)}} \E_{q}\cor*{\text{v}_{c,s}^{\mathbf{Y}2} }}.
\end{align}

Moving to the sparsity-inducing parameters, we get
\begin{align} 
\E_{q}\cor*{\lnp{p\p*{\dm}}} 
\eqeq \sums \left(- \beta_0^{\dm} \E_{q}\cor*{\dsm} + \alpha_0^{\dm} \lnp{\beta_0^{\dm}} \right.\nonumber \\
& \left.+ \p*{\alpha_0^{\dm}-1} \E_{q}\cor*{\lnp{\dsm}} - \lnp{\Gamma\p*{\alpha_0^{\dm}}}\right), \nonumber
\end{align} 
and
\begin{align} 
\E_{q}\cor*{\lnp{p\p*{\lm}}} 
\eqeq \sumd \left(- \beta_0^{\lm} \E_{q}\cor*{\ldm} + \alpha_0^{\lm} \lnp{\beta_0^{\lm}}\right.\nonumber \\
& \left.+ \p*{\alpha_0^{\lm}-1} \E_{q}\cor*{\lnp{\ldm}} - \lnp{\Gamma\p*{\alpha_0^{\lm}}}\right), \nonumber
\end{align} 
where the expectations are determined in Equations \eqref{eq:ElogDelta} and \eqref{eq:EDelta} for $\dm$ 
and in Equations \eqref{eq:ElogLambda} and \eqref{eq:ELambda} for $\lm$.

Looking at the generation of the observations, we have
\begin{align} 
\E_{q}&\cor*{\lnp{p\p*{\Xm|\Vm,\G,\psim}}} \nonumber \\ 
\eqline \sumn \E_{q}\left[-\frac{D^{(m)}}{2}\lnp{2\pi}+\frac{D^{(m)}}{2}\lnp{\psim} \right. \left.- \frac{1}{2}\p*{\Xnm-\G \VmT}\psim\p*{\Xnm-\G \VmT}^\top\right] \nonumber\\
\eqeq \sumn \left(-\frac{D^{(m)}}{2}\lnp{2\pi}+\frac{D^{(m)}}{2}\E_{q}\cor*{\lnp{\psim}}  \right. \left.- \frac{1}{2} \E_{q}\cor*{\psim} \E_{q}\left[\Xnm \XnmT  \right.\right. \nonumber \\ 
\eqline \left.\hspace{-0.1cm} \left.- 2\Xnm\Vm\GnT + \Gn \VmT \Vm \GnT\right]\right) \nonumber \\
\eqeq -\frac{N D^{(m)}}{2}\lnp{2\pi}+\frac{D^{(m)}}{2}\sumn \p*{\E_{q}\cor*{\lnp{\psim}}} - \frac{1}{2}\E_{q}\cor*{\psim} \sumn \left(\Xnm \XnmT \right. \nonumber \\ 
\eqline \left.- 2 \Tr\left\{\Xnm\ang{ \Vm} \ang{ \GnT}  \right\} +\Tr\left\{\ang{ \VmT,\Vm} \ang{ \GnT,\Gn}  \right\}\right).
\end{align}


Considering the noise, we have that
\begin{align} 
\E_{q}&\cor*{\lnp{p\p*{\psim}}} \nonumber \\
\eqeq - \beta_0^{\psim}\E_{q}\cor*{\psim} + \alpha_0^{\psim}\lnp{\beta_0^{\psim}} + \p*{\alpha_0^{\psim}-1}\E_{q}\cor*{\lnp{\psim}} - \lnp{\Gamma\p*{\alpha_0^{\psim}}} \nonumber \\
\eqeq \alpha_0^{\psim}\lnp{\beta_0^{\psim}} - \lnp{\Gamma\p*{\alpha_0^{\psim}}} - \beta_0^{\psim}\frac{\alpha^{\psim}}{\beta^{\psim}} + \p*{\alpha_0^{\psim}-1}\p*{\psi\p*{\alpha^{\psim}} - \lnp{\beta^{\psim}}},\nonumber \\
\label{eq:ElogpTau}
\end{align} 
where $\taum$ follows a Gamma distribution, so the expectation of its parameters will be equivalent to Equations \eqref{eq:ElogDelta} and \eqref{eq:EDelta}.

If we now consider the discriminative space, we have that term $\ln p(\mathbf{Z}|\mathbf{W}^{\M},\tau)$ can be developed as:
\begin{equation}
\label{eq:full_elbo_z}
\begin{split}
    & \ln p(\mathbf{Z}|\mathbf{W}^{\M},\tau) = -\frac{N}{2}\lnpi + \frac{N}{2}\ln \vert\tau\vert \\ 
    & - \sumn\unmed\tau\Bigg(\Zn -\summ\Bigg(\Xnm\WmT\Bigg)\Bigg)\Bigg(\Zn -\summ\Bigg(\Xnm\WmT\Bigg)\Bigg)^\top\\
    & = \sumn\unmed\tau\Bigg(\Zn\ZnT - 2\Zn\mathbf{J}_{\rm n,:}^\top - \mathbf{J}_{\rm n,:}\mathbf{J}_{\rm n,:}^\top\Bigg),
\end{split}
\end{equation}
where
\begin{equation}
\mathbf{J}_{\rm n,:} = \Zn -\summ\Bigg(\Xnm\WmT\Bigg).
\end{equation}

Now, if we operate and return to Eq. (\ref{eq:full_elbo_z}), we can apply the expectation over the model variables and obtain:
\begin{equation}
\label{eq:casi_fin_z}
\begin{split}
    & \E\left[\ln p(\mathbf{Z}|\mathbf{W}^{\M},\tau)\right] = -\frac{N}{2}\lnpi + \frac{N}{2}\E\left[\ln \vert\tau\vert\right]  \\
    & -2\langle\Zn\rangle\E\left[\mathbf{U}_{\rm n,:}\right] + \summ\left[\Xnm\langle\WmT\Wm\rangle\XnmT\right] \\
    & + \summ\sum_{m'\neq m}^M \Bigg(\Xnm\langle\WmT\rangle\langle\mathbf{W}^{(m')}\rangle\mathbf{x}_{\rm n,:}^{(m')\top}\Bigg).
\end{split}
\end{equation}

Also note that $\E\left[\ln(\tau)\right] = dg(\alpha^\tau) - \ln \beta^\tau$, and $dg()$ is the digamma function and $\E\left[\tau\right] = \frac{\alpha^\tau}{\beta^\tau}$.

Hence, with Eq. (\ref{eq:beta_tau}) we can simplify and rewrite Eq. (\ref{eq:casi_fin_z}) as:
\begin{equation}
\begin{split}
    \E\left[\ln p(\mathbf{Z}|\mathbf{W}^{\M},\tau)\right] = & -\frac{N}{2}\lnpi + \frac{N}{2}\left(dg(\alpha^\tau) - \ln \beta^\tau\right) \\ &-\unmed\frac{\alpha^\tau}{\beta^\tau}\left(\beta^\tau - \beta^\tau_0\right).
\end{split}
\end{equation}

Now, the development of $\ln p(\Y\vert\Z,\U,\G,\mathbf{V}^{(M+1)},\eta)$ will be analogous to $\ln p(\mathbf{Z}|\mathbf{W}^{\M},\tau)$. That is, we can develop the output term as:
\begin{equation}
    \E\left[\ln p(\Y\vert\Z,\U,\G,\mathbf{V}^{(M+1)},\eta)\right] = -\frac{N}{2}\lnpi + \frac{N}{2}\left(dg(\alpha^\eta) - \ln \beta^\eta\right) -\unmed\frac{\alpha^\eta}{\beta^\eta}\left(\beta^\eta - \beta^\eta_0\right).
\end{equation}

Also, if we focus in $\ln p(\mathbf{T}\vert\mathbf{Y})$, we will have to apply the expectation over the logarithm of the lower bounded expression depicted in Eq. (\ref{eq:y_first}). Thus, we will finally achieve:
\begin{equation}
    \E(\ln(h(\mathbf{Y}, \bm{\xi}))) = \sumn\sumc(\ln(\sigma(\xi_{\rm n,c})) + \langle y_{\rm n,c}\rangle t_{\rm n,c} - \unmed(\langle y_{\rm n,c}\rangle + \xi_{\rm n,c}) - \lambda(\xi_{\rm n,c})(\langle\y_{\rm n,c}^2\rangle - \xi_{\rm n,c}^2)).
\end{equation}

The development of $p(\U)$ is analogous to Eq. (\ref{eq:G_q}), thus
\begin{align} 
\E_{q}\cor*{\lnp{p\p*{\U}}} 
\eqeq \sum_k^K \E_{q}\cor*{-\frac{C}{2}\lnp{2\pi}-\frac{1}{2}\mathbf{u}_{\rm :,k}^\top \mathbf{u}_{\rm :,k}}
= -\frac{K C}{2}\lnp{2\pi}-\frac{1}{2} \Tr\{\ang{\U \UT}\}. \label{eq:ElogpUpsilon}
\end{align}

Furthermore, we can develop the expectation over the output noise $\eta$ as:
\begin{equation}
\label{eq:eta_jon}
\begin{split}
    & \E\left[\ln p(\eta)\right] = \E\left[\alpha_0^\eta\ln\beta_0^\eta - \ln\left(\Gamma(\alpha_0^\eta)\right) - \beta_0^\eta \eta + (\alpha_0^\eta -1)\ln(\tau)\right] \\
    & = \alpha_0^\eta\ln\beta_0^\eta - \ln\left(\Gamma(\alpha_0^\eta)\right) - \beta_0^\eta \frac{\alpha_0^\eta}{\beta_0^\eta} + (\alpha_0^\eta -1)(dg(\alpha_0^\eta) - \ln \beta_0^\eta).
\end{split}
\end{equation}

Also, for $\tau$ we follow the same process as in Eq. (\ref{eq:eta_jon}). That is
\begin{equation}
\begin{split}
    & \E\left[\ln p(\tau)\right] = \E\left[\alpha_0^\tau\ln\beta_0^\tau - \ln\left(\Gamma(\alpha_0^\tau)\right) - \beta_0^\tau \tau + (\alpha_0^\tau -1)\ln(\tau)\right] \\
    & = \alpha_0^\tau\ln\beta_0^\tau - \ln\left(\Gamma(\alpha_0^\tau)\right) - \beta_0^\tau \frac{\alpha_0^\tau}{\beta_0^\tau} + (\alpha_0^\tau -1)(dg(\alpha_0^\tau) - \ln \beta_0^\tau),
\end{split}
\end{equation}

Also, the development of the $m$-th view of $p(\mathbf{W}^{(m)}\vert\boldsymbol{\psi}^{(m)}, \boldsymbol{\gamma}^{(m)})$ can be expressed as
\begin{equation}
\begin{split}
\label{eq:www}
    &\ln p(\mathbf{W}^{(m)}\vert\boldsymbol{\psi}^{(m)}, \boldsymbol{\gamma}^{(m)}) = -\frac{K D^{(m)}}{2}\lnpi + \unmed\sumk\sumd(\ln(\gamma_d\phi_k)) - \unmed\sumk\sumd(w_{\rm k,d}^2\delta_k\gamma_d) \\
    & = -\frac{K D^{(m)}}{2}\lnpi + \frac{D^{(m)}}{2}\sumk\ln\phi_k + \frac{K}{2}\sumd\ln\gamma_d - \unmed\sumk\sumd(w_{\rm k,d}^2\phi_k\gamma_d). 
\end{split}
\end{equation}

If we apply expectations over Eq. (\ref{eq:www}), we obtain:
\begin{equation}
\begin{split}
    &\E\left[\ln p(\mathbf{W}^{(m)}\vert \boldsymbol{\phi}^{(m)}, \boldsymbol{\gamma}^{(m)})\right] = -\frac{K D^{(m)}}{2}\lnpi + \frac{D^{(m)}}{2}\sumk(dg(\alpha_k^\phi) - \ln \beta_k^\phi) \\ &+ \frac{K}{2}\sumd(dg(\alpha_d^\gamma) - \ln \beta_d^\gamma)
     - \unmed\sumk\sumd\left[\frac{\alpha_k^\phi\alpha_d^\gamma}{\beta_k^\phi\beta_d^\gamma}\langle w_{\rm k,d}^2\rangle\right].
\end{split}
\end{equation}

The developments of $p(\boldsymbol{\phi}^\M)$ and $p(\boldsymbol{\gamma}^\M)$ are anlogous to Eq. (\ref{eq:eta_jon}). Thus, we obtain
\begin{equation}
\label{eq:alphaaa}
\begin{split}
    \E\left[\boldsymbol{\phi^{(m)}}\right] &= K\left(\alpha_0^{\phi^{(m)}}\ln \beta_0^{\phi^{(m)}} - \ln \left(\Gamma(\alpha_0^{\phi^{(m)})}\right)\right) \\
    & + \sumk\left(-\beta_0^{\phi^{(m)}}\frac{\alpha_k^{\phi^{(m)}}}{\beta_k^{\phi^{(m)}}} + (\alpha_0^{\phi^{(m)}}-1)(dg(\alpha_k^{\phi^{(m)}}) - \ln \beta_k^{\phi^{(m)}})\right),
\end{split}
\end{equation}
and
\begin{equation}
\label{eq:gamaaa}
\begin{split}
    \E\left[\boldsymbol{\gamma^{(m)}}\right] &= D\left(\alpha_0^{\gamma^{(m)}}\ln \beta_0^{\gamma^{(m)}} - \ln \left({\gamma^{(m)}}(\alpha_0^{\gamma^{(m)}})\right)\right) \\
    & + \sumd\left(-\beta_0^{\gamma^{(m)}}\frac{\alpha_d^{\gamma^{(m)}}}{\beta_d^{\gamma^{(m)}}} + (\alpha_0^{\gamma^{(m)}}-1)(dg(\alpha_d^{\gamma^{(m)}}) - \ln \beta_d^{\gamma^{(m)}})\right),
\end{split}
\end{equation}

\subsubsection{\texorpdfstring{Terms associated with $\E_{q}\left[\ln (q(\boldsymbol{\Theta}))\right]$}{Terms associated with Eq[log(q(Theta))]}}

The term comprises the following elements:
\begin{equation}
\begin{split}
    &\E_{q}\left[\ln (q(\bm{\Theta}))\right] = \E\Big[\ln q(\Y) +\ln q(\Z) +\ln q(\G) + q(\mathbf{V}^{\mathbf{Y}}) + \ln q(\U) + \ln q(\tau) + \ln q(\eta) \\
    &  + \ln q(\boldsymbol{\delta}^{(M+1)}) + \ln q(\boldsymbol{\lambda}^{(M+1)}) \sum_m^M\Big(\ln q(\Wm) + \ln q(\Vm) + \ln q(\deltam) + \ln q(\boldsymbol{\lambda}^{(m)})\\
    & + \ln q(\phim) + \ln q(\boldsymbol{\gamma}^{(m)}) + \ln q(\psim))\Big)\Big]
\end{split}
\end{equation}

Also, as in previous $\E_{q}\left[\ln (p(\bm{\Theta}, \mathbf{t}, \X^{\M}))\right]$ each element can be developed independently. For the discriminative variables:
\begin{align}
    \ln q(\mathbf{Y}) & = \sumn\left(\unmed\ln 2\pi e + \unmed\ln \vert\Sigma_{\mathbf{y}_{\rm n,:}}\vert\right) = \frac{N}{2}\ln 2\pi e + \frac{N}{2}\ln \vert\Sigma_{\mathbf{Y}}\vert\\
    \ln q(\mathbf{U}) & = \sumc\left(\unmed\ln 2\pi e + \unmed\ln \vert\Sigma_{\mathbf{U}}\vert\right) = \frac{C}{2}\ln 2\pi e + \frac{C}{2}\ln \vert\Sigma_{\mathbf{U}}\vert \\
    \ln q(\boldsymbol{\phi^{(m)}}) & = \sumk\left(\alpha_k^{\phi^{(m)}} + \ln \Gamma(\alpha_k^{\phi^{(m)}}) - (1-\alpha_k^{\phi^{(m)}})dg(\alpha_k^{\phi^{(m)}}) - \ln \beta_k^{\phi^{(m)}}\right) \\
    \ln q(\boldsymbol{\gamma^{(m)}}) & = \sumd\left(\alpha_d^{\gamma^{(m)}} + \ln \Gamma(\alpha_d^{\gamma^{(m)}}) - (1-\alpha_d^{\gamma^{(m)}})dg(\alpha_d^{\gamma^{(m)}}) - \ln \beta_d^{\gamma^{(m)}}\right) \\
    \ln q(\eta) & = \alpha^\eta + \ln \Gamma(\alpha^\eta) - (1-\alpha^\eta)dg(\alpha^\eta) - \ln \beta^\eta \\
    \ln q(\tau) & = \alpha^\tau + \ln \Gamma(\alpha^\tau) - (1-\alpha^\tau)dg(\alpha^\tau) - \ln \beta^\tau \\
    \ln q(\mathbf{Z}) &= \frac{N}{2}\ln 2\pi e + \frac{N}{2}\ln\vert\Sigma_{\mathbf{Z}}\vert\\
    \ln q(\mathbf{W}^{(m)}) &= \frac{K}{2}\ln 2\pi e + \unmed\sumk\ln\vert\Sigma_{\mathbf{W}^{(m)}_{\rm k,:}}\vert
\end{align}

For the generative variables:
\begin{align}
    \ln q(\mathbf{V}^{(M+1)}) & = \sumc\left(\unmed\ln 2\pi e + \unmed\ln \vert\Sigma_{\mathbf{V}^{(M+1)}}\vert\right) = \frac{C}{2}\ln 2\pi e + \frac{C}{2}\ln \vert\Sigma_{\mathbf{V}^{(M+1)}}\vert \\
    \ln q(\mathbf{V}^{(m)}) & = \sumc\left(\unmed\ln 2\pi e + \unmed\ln \vert\Sigma_{\mathbf{V}^{(m)}}\vert\right) = \frac{C}{2}\ln 2\pi e + \frac{C}{2}\ln \vert\Sigma_{\mathbf{V}^{(m)}}\vert \\
    \ln q(\boldsymbol{\delta^{(m)}}) & = \sum_s^S\left(\alpha_s^{\delta^{(m)}} + \ln \Gamma(\alpha_s^{\delta^{(m)}}) - (1-\alpha_s^{\delta^{(m)}})dg(\alpha_s^{\delta^{(m)}}) - \ln \beta_s^{\delta^{(m)}}\right) \\
    \ln q(\boldsymbol{\lambda^{(m)}}) & = \sumd\left(\alpha_d^{\lambda^{(m)}} + \ln \Gamma(\alpha_d^{\lambda^{(m)}}) - (1-\alpha_d^{\lambda^{(m)}})dg(\alpha_d^{\lambda^{(m)}}) - \ln \beta_d^{\lambda^{(m)}}\right) \\
    \ln q(\boldsymbol{\delta^{(M+1)}}) & = \sum_s^S\left(\alpha_s^{\delta^{(M+1)}} + \ln \Gamma(\alpha_s^{\delta^{(M+1)}}) - (1-\alpha_s^{\delta^{(M+1)}})dg(\alpha_s^{\delta^{(M+1)}}) - \ln \beta_s^{\delta^{(M+1)}}\right) \\
    \ln q(\boldsymbol{\lambda^{(M+1)}}) & = \sumd\left(\alpha_d^{\lambda^{(M+1)}} + \ln \Gamma(\alpha_d^{\lambda^{(M+1)}}) - (1-\alpha_d^{\lambda^{(M+1)}})dg(\alpha_d^{\lambda^{(M+1)}}) - \ln \beta_d^{\lambda^{(M+1)}}\right) \\
    \ln q(\psim) & = \alpha^{\psim} + \ln \Gamma(\alpha^{\psim}) - (1-\alpha^{\psim})dg(\alpha^{\psim}) - \ln \beta^{\psim} \\
    \ln q(\mathbf{G}) &= \frac{N}{2}\ln 2\pi e + \frac{N}{2}\ln\vert\Sigma_{\mathbf{G}}\vert
\end{align}

\subsubsection{Complete Lower Bound}

Finally, we can introduce both $E_{q}\left[\ln (p(\boldsymbol{\Theta}, \mathbf{t}, \mathbf{X}^{\M}))\right]$ and $\E_{q}\left[\ln (q(\boldsymbol{\Theta}))\right]$ in Eq. (\ref{eq:lp_tot}) and simplify to obtain the final $L(q)$ as:
\begin{equation}
\label{eq:Lq}
    \begin{split}
        L(q) = &\frac{N}{2}\ln |\Sigma_{\mathbf{Z}}|- (2 + \frac{N}{2} - \alpha_0^\tau)\ln(\beta^\tau) - (2 + \frac{N}{2} - \alpha_0^\eta)\ln(\beta^\eta)  + \frac{N}{2}\ln|\Sigma_{\mathbf{y}}| + \frac{C}{2}\ln|\Sigma_{\mathbf{U}}| \\ &  + \sum_m^M\bigg[\left(\frac{D}{2} + \alpha_0^{\gamma^{\rm (m)}} -2\right)\sum_{\rm d}^{D^{\rm (m)}}\ln(\beta_{\rm d}^{\gamma^{\rm (m)}}) \\ & + \left(\frac{K}{2} + \alpha_0^{\phi^{\rm (m)}} -2\right)\sum_{\rm k}^K\ln(\beta_{\rm k}^{\phi^{\rm (m)}}) + \sum_{\rm d}^{D^{\rm (m)}}\left(\beta_0^{\gamma^{\rm (m)}}\frac{\alpha_{\rm d}^{\gamma^{\rm (m)}}}{\beta_{\rm d}^{\gamma^{\rm (m)}}}\right) + \sum_{\rm k}^K\left(\beta_0^{\phi^{\rm (m)}}\frac{\alpha_{\rm k}^{\phi^{\rm (m)}}}{\beta_{\rm k}^{\phi^{\rm (m)}}}\right) \\ & + \unmed\sum_{\rm k}^K\sum_{\rm d}^{D^{\rm (m)}}\left(\frac{\alpha_{\rm k}^{\delta^{\rm (m)}} \alpha_{\rm d}^{\gamma^{\rm (m)}}}{\beta_{\rm k}^{\delta^{\rm (m)}} \beta_{\rm d}^{\gamma^{\rm (m)}}}\langle\Wkd\rangle\right) \Bigg]\\  & - \sum_n^N\left(\ln(\sigma(\xi_{\rm n,:})) + \langle\mathbf{y}_{\rm n,:}\rangle\mathbf{t}_{\rm n,:}^\top - \unmed(\langle\mathbf{y}_{\rm n,:}\rangle + \xi_{\rm n,:}) - \lambda(\xi_{\rm n,:})(\langle\mathbf{y}_{\rm n,:}^2\rangle - \xi_{\rm n,:}^2)\right)\\
        & -\frac{1}{2} \Tr\{\ang{\GT \G}\} - \frac{N}{2} \ln|\Sigma_{\G}| \\
        & + \sum_m^M\Bigg[\left(\frac{D^{(m)}}{2} + \alpha_0^{\lambda^{\rm (m)}} -2\right)\sum_{\rm d}^{D^{\rm (m)}}\ln(\beta_{\rm d}^{\lambda^{\rm (m)}}) +\left(\frac{S}{2} + \alpha_0^{\delta^{\rm (m)}} -2\right)\sum_{\rm s}^{S^{\rm (m)}}\ln(\beta_{\rm s}^{\delta^{\rm (m)}})\\
        &  - (2 + \frac{N}{2} - \alpha_0^{\psim})\ln(\beta^{\psim})  -\left( \frac{D}{2}\ln\left|\Sigma_{\Vm}\right| \right)\Bigg]
    \end{split}
\end{equation}

\newpage

\subsection{Posterior predictive distribution}

Since the model incorporates two intermediate stages between the input data $\mathbf{X}^\M$ and the final classification determined by $\mathbf{t}$, predictions are performed in a hierarchical manner. Thus, to calculate the posterior predictive distribution, we begin marginalizing the posterior distribution of $\mathbf{t}$ over $\mathbf{y}$ for a new data point $\xpredin$ as
\begin{equation}
    p(\tpredc = 1|\xpredin) = \int_{\ypredc} p(\tpredc = 1|\ypredc)p(\ypredc|\xpredin)d\ypredc,
    \label{eq:dist_clas}
\end{equation}
where $\ypredc$ is the $c$-th column of the regression output of the model. Furthermore, as
\begin{equation}
    p(\mathbf{t}_{\rm n,:}|\mathbf{y}_{\rm n,:}) = e^{\mathbf{y}_{\rm n,:} \mathbf{t}_{\rm n,:}}\sigma(-\mathbf{y}_{\rm n,:}),
    \label{eq:reg_mod}
\end{equation}
we can express Eq. \eqref{eq:dist_clas} as
\begin{equation}
\label{eq:true_post}
    p(\tpredc = 1|\xpredin) = \int_{\ypredc} \sigma( \ypredc)p(\ypredc|\xpredin)d\ypredc,
\end{equation}
where $\sigma(\cdot)$ is the sigmoid function.

The challenge we face is that Eq. \eqref{eq:true_post} does not have an analytical or closed-form solution. Thus, we will perform the approximation presented in \cite{bishop2006pattern} that suggests that $\sigma(\ypredc)$ can be approximated by the Gaussian cumulative distribution function $\Phi(\kappa \ypredc)$, where $\kappa$ acts as a scaling parameter on the horizontal axis of the probit function to align it with a sigmoid function that maps $\ypredc$ into the interval $(0,1)$. Now, the integral of Eq. \eqref{eq:true_post} can be approximated as follows:
\begin{equation}
\label{eq:pre_prob}
    p(\tpredc = 1|\xpredin) \simeq \int_{\chi_{\ypredc}} \Phi(\kappa \ypredc)p(\ypredc|\xpredin)d\ypredc = \Phi\left(\frac{\langle \ypredc\rangle}{(\kappa^{-2} +\Sigma_{\ypredc})^{\frac{1}{2}}}\right),
\end{equation}
where $\langle \ypredc\rangle$ and $\Sigma_{\ypredc}$ represent the mean and variance of $\ypredc$, respectively. Moreover, 
given that the similarity between the sigmoid and the probit functions is maximized when $\kappa^2 = \frac{\pi}{8}$, we can rewrite Eq. \eqref{eq:pre_prob} as 
\begin{equation}
\label{eq:final_pred}
    p(\tpredc = 1|\xpredin) \simeq \sigma\left(\frac{\langle \ypredc\rangle }{(1 + \frac{\pi}{8}\Sigma_{\ypredc})^{\unmed}}\right).
\end{equation}
where $\Phi(\cdot)$ is the probit function.

Now, to infer the parameters $\langle \ypredc\rangle$ and $\Sigma_{\ypredc}$, it is necessary to parameterize $p(\ypredc|\xpredin)$. This involves marginalizing over the main model variables as
\begin{equation}
\label{eq:y_arr}
    p(\ypredc|\xpredin) = \int p(\ypredc|\boldsymbol{\Theta}) p\p*{\boldsymbol{\Theta}|\xpredin} d\boldsymbol{\Theta},
\end{equation}
where $\boldsymbol{\Theta} = \{\Z, \G, \U, \mathbf{V}^{\Y}, \eta\}$. Having developed mean-field inference over the model variables, we can rewrite Eq. \eqref{eq:y_arr} employing the obtained approximate posterior distributions as follows:
\begin{equation}
\begin{split}
\label{eq:yyy}
    p(\ypredc|\xpredin) = & \int_{\mathbf{z}_{\rm *, :}}\int_{\mathbf{g}_{\rm *,:}}\int_{\mathbf{u}_{\rm c,:}}\int_{\mathbf{v}_{\rm c,:}^{\Y}}\int_\eta N(\ypredc|\mathbf{z}_{\rm *,:} \mathbf{u}_{\rm c,:}^\top + \mathbf{g}_{\rm *,:} \mathbf{v}_{\rm c,:}^{\mathbf{Y}\top},\eta^{-1})q(\mathbf{z}_{\rm *, :})q(\mathbf{g}_{\rm *,:}) \\
    & q(\mathbf{u}_{\rm c,:})q(\mathbf{v}_{\rm c,:}^{\Y})q(\eta) d\mathbf{z}_{\rm *,:} d\mathbf{g}_{\rm *,:} d\mathbf{u}_{\rm c,:} d\mathbf{v}_{\rm c,:}^{\mathbf{Y}\top} d\eta.
\end{split}
\end{equation}

Although the integration over $\eta$ is intractable, due to the asymptotic behavior of the variance of the gamma distribution, $\eta$ will be concentrated around its mean as the sample size increases, i.e., $\langle \eta^2 \rangle - \langle \eta \rangle^2 = \frac{a^{\eta}}{b^{\eta 2}} \sim \mathcal{O}(\frac{1}{N})$ (see \cite{bishop2013variational}). Therefore, Eq. \eqref{eq:yyy} can be approximated as:
\begin{equation}
\begin{split}
    p(\ypredc|\xpredin) = & \int_{\mathbf{z}_{\rm *, :}}\int_{\mathbf{g}_{\rm *,:}}\int_{\mathbf{u}_{\rm c,:}}\int_{\mathbf{v}_{\rm c,:}^{\Y}} N(\ypredc|\mathbf{z}_{\rm *,:} \mathbf{u}_{\rm c,:}^\top + \mathbf{g}_{\rm *,:} \mathbf{v}_{\rm c,:}^{\mathbf{Y}\top},\langle\eta^{-1}\rangle)q(\mathbf{z}_{\rm *, :})q(\mathbf{g}_{\rm *,:}) \\
    & q(\mathbf{u}_{\rm c,:})q(\mathbf{v}_{\rm c,:}^{\Y}) d\mathbf{z}_{\rm *,:} d\mathbf{g}_{\rm *,:} d\mathbf{u}_{\rm c,:} d\mathbf{v}_{\rm c,:}^{\mathbf{Y}\top}.
\end{split}
\end{equation}

Finally, since only a product of Gaussian distributions remains, the resulting distribution will also be Gaussian. Moreover, by exploiting the properties of integrals, these can be treated as convolutions of Gaussian functions, allowing the multiple integral to be computed step-by-step. Hence, the we can calculate the mean of $p(\ypredc|\xpredin)$ as:
\begin{equation}
\label{eq:mean_yy}
    \E[\ypredc] = \E[\mathbf{z}_{\rm *,:} \mathbf{u}_{c,:}^\top + \mathbf{g}_{\rm *,:} \mathbf{v}_{c,:}^{\mathbf{Y}\top} + \epsilon_{Y}] = \langle\mathbf{z}_{\rm *,:}\rangle\langle\mathbf{u}_{c,:}^\top\rangle + \langle\mathbf{g}_{\rm *,:}\rangle\langle\mathbf{v}_{c,:}^{\mathbf{Y}\top}\rangle,
\end{equation}
and the variance
\begin{equation}
    Var[\ypredc] = Var[\mathbf{z}_{\rm *,:} \mathbf{u}_{c,:}^\top] + Var[\mathbf{g}_{\rm *,:} \mathbf{v}_{c,:}^{\mathbf{Y}\top}] + \langle\eta\rangle,
\end{equation}
where 
\begin{equation}
\label{eq:var_z}
\begin{split}
    & Var[\mathbf{z}_{\rm *,:} \mathbf{u}_{\rm c,:}^\top]  = Var\left(\sum_k^{K}\text{z}_{\rm *,k} \text{u}_{\rm c,k}\right) = \sum_k^K\Big(\E[\text{z}_{\rm *,k}^2]\E[\text{u}_{\rm c,k}^2] - \E[\text{z}_{\rm *,k}]\E[\text{z}_{\rm *,k}]\E[\text{u}_{\rm c,k}]\E[\text{u}_{\rm c,k}]\Big)\\
    & = \sum_k^K\Big((Var[\text{z}_{\rm *,k}] + \E[\text{z}_{*,k}]\E[\text{z}_{\rm *,k}])(Var[\text{u}_{\rm c,k}] + \E[\text{u}_{\rm c,k}]\E[\text{u}_{\rm c,k}]) - \E[\text{z}_{\rm *,k}]\E[\text{u}_{\rm c,k}]\E[\text{u}_{\rm c,k}]\Big) \\
    & = \sum_k^K\Big(Var[\text{z}_{\rm *,k}]Var[\text{u}_{\rm c,k}] + \E[\text{z}_{\rm *,k}]\E[\text{z}_{\rm *,k}] Var[\text{u}_{\rm c,k}] + \E[\text{u}_{\rm c,k}]\E[\text{u}_{\rm c,k}] Var[\text{z}_{\rm *,k}]\Big) \\
    & = \sum_k^K\Big(\Sigma_{\text{z}_{\rm *,k}}\Sigma_{\text{u}_{\rm c,k}} + \langle\text{z}_{\rm *,k}\rangle\langle\text{z}_{\rm *,k}\rangle\Sigma_{\text{u}_{\rm c,k}} + \langle \text{u}_{\rm c,k}\rangle\langle\text{u}_{\rm c,k} \rangle\Sigma_{\mathbf{z}_{\rm *,k}}\Big),
\end{split}
\end{equation}
where $\Sigma_{\text{z}_{\rm *,k}}$ is the value of the cell at position $(\text{k},\text{k})$ of $\Sigma_{\text{z}_{\rm *,:}}$, and $\Sigma_{\text{u}_{\rm c,k}}$ analog to $\Sigma_{\text{u}_{\rm c,:}}$.

Note that the development of $Var[\mathbf{g}_{\rm *,:} \mathbf{v}_{c,:}^{\mathbf{Y}\top}]$ is analogue to Eq. \eqref{eq:var_z}. Thus, the final expression of $Var[\ypredc]$ will be
\begin{equation}
\begin{split}
\label{eq:var_yy}
    Var[\ypredc] & = \langle\eta\rangle + \sum_k^K\Big(\Sigma_{\text{z}_{\rm *,k}}\Sigma_{\text{u}_{\rm c,k}} + \langle\text{z}_{\rm *,k}\rangle\langle\text{z}_{\rm *,k}\rangle\Sigma_{\text{u}_{\rm c,k}} + \langle \text{u}_{\rm c,k}\rangle\langle\text{u}_{\rm c,k} \rangle\Sigma_{\mathbf{z}_{\rm *,k}}\Big) \\
    & + \sum_{s}^{S}\Big(\Sigma_{\mathbf{g}_{\rm *,s}}\Sigma_{\mathbf{v}_{\rm c,s}^{Y}} + \langle\mathbf{g}_{\rm *,s}\rangle\langle\mathbf{g}_{\rm *,s}\rangle\Sigma_{\mathbf{v}^{Y}_{\rm c,s}} + \langle \mathbf{
v}^{Y}_{\rm c,s}\rangle\langle\mathbf{v}^{Y}_{\rm c,s} \rangle\Sigma_{\mathbf{g}_{\rm *,s}}\Big).
\end{split}
\end{equation}

Thus, $p(\ypredc|\xpredin)$ will be parameterized with expressions of Eq. \eqref{eq:mean_yy} and \eqref{eq:var_yy}. 

\newpage

\subsection{Missing values imputation}

To address missing values and enable semi-supervised training, we treat the missing entries as additional r.v. to be inferred. Let $U$ denote the set of unobserved features and $O$ the set of observed ones. For a given data point in modality $m$, we write $\mathbf{x}_{*,U}^{(m)}$ for the unobserved features and $\mathbf{x}_{*,O}^{(m)}$ for the observed ones. The posterior distribution of the missing entries can then be expressed as
\begin{equation}
    p(\text{x}_{*,U}^{(m)}\vert\text{x}_{*,O}^{(m)}, \boldsymbol{\Theta}) = \prod_{m}^{\M} p(\text{x}_{*,U}^{(m)}\vert \mathbf{x}_{*,O}^{\M}, \boldsymbol{\Theta}),
\end{equation}
where
\begin{equation}
\label{eq:miss}
    p(\text{x}_{*,U}^{(m)}\vert\text{x}_{*,O}^{\M}, \boldsymbol{\Theta}) = \int p(\mathbf{x}^{(m)}_{*,U} \vert \mathbf{g}_{*,:}, \boldsymbol{\Theta}) p(\mathbf{g}_{*,:} \vert \text{x}_{*,O}^{\M},\boldsymbol{\Theta})d\boldsymbol{\Theta},
\end{equation}
Here, $p(\mathbf{x}_{*,U}^{(m)} \mid \mathbf{g}_{*,:}, \boldsymbol{\Theta})$ follows Eq.~\eqref{eq:x}. Expanding Eq.~\eqref{eq:miss}, we obtain
\begin{align}
    p(\mathbf{x}^{(m)}_{*,U}\vert \mathbf{x}_{*,O}^{\M}, \boldsymbol{\Theta}) = \int_{\mathbf{g}_{*,:}}\int_{\mathbf{V}_{U,:}^{(m)}} \int_{\psi^{(m)}} &\N(\mathbf{x}^{(m)}_{*,:} \vert \mathbf{g}_{\rm *,:} \mathbf{V}_{U,:}^{(m)\top}, \psi^{(m)-1})q(\mathbf{g}_{*,:}) \nonumber\\
    & q(\mathbf{V}_{U,:}^{(m)})q(\psi^{(m)})d \mathbf{g}_{*,:} d\mathbf{V}_{U,:}^{(m)} d\psi^{(m)}.
\end{align}

Following analogous steps to those in the previous section, the posterior mean and variance of the unobserved features are given by
\begin{equation}
    \langle\mathbf{x}_{*,U}^{(m)}\rangle = \langle\mathbf{g}_{\rm *,:}\rangle\langle\V_{U,:}^{(m)\top}\rangle,
\end{equation}
and
\begin{equation}
    \Sigma_{\text{x}_{*,u}^{(m)}} = \langle\psi^{(m)}\rangle^{-1} + \Sigma_{\mathbf{g}_{*,:}}\Sigma_{\mathbf{V}_{u,:}^{(m)}} + \langle\mathbf{g}_{*,:}\mathbf{g}_{*,:}^\top\rangle\Sigma_{\mathbf{V}_{u,:}^{(m)}} + \langle\mathbf{v}_{u,:}^{(m)}\mathbf{V}_{u,:}^{(m)\top}\rangle\Sigma_{\mathbf{g}_{*,:}}.
\end{equation}

Finally, the marginal posterior distribution of $\mathbf{g}_{*,:}$ is computed as
\begin{equation}
    \Sigma_{\mathbf{G}}^{-1} = \Is + \langle\eta\rangle\langle\mathbf{V}^{\mathbf{Y}\top} \mathbf{V}^{\mathbf{Y}}\rangle + \sum_{m}^{\M}\langle\psi^{(m)}\rangle\langle\mathbf{V}_{U,:}^{(m)\top}\mathbf{V}_{U,:}^{(m)}\rangle 
\end{equation}
with mean
\begin{equation}
    \langle\mathbf{g}_{\rm *,:}\rangle = \left( \sum_{m}^{\M}\langle\psi^{(m)}\rangle\langle\mathbf{x}_{\rm *,U}^{(m)}\rangle\langle\mathbf{V}_{U,:}^{(m)}\rangle + \langle\eta\rangle   \left( \langle\mathbf{y}_{*,:}\rangle -\langle\mathbf{z}_{\rm *,:}\rangle\langle\mathbf{U}\rangle^\top \right)\langle\mathbf{V}^{\mathbf{Y}}\rangle \right)\Sigma_{\mathbf{G}}.
\end{equation}

\newpage

\section{Generative stable factors analysis}

In this appendix, we present the stable generative factors identified by the model on the ADNI dataset, which comprises 13 modalities. The model has inferred 13 latent factors. A factor is considered stable if it appears in at least 8 out of the 10 cross-validation folds. Additionally, two factors are considered the same if their absolute cosine similarity exceeds 0.95.

Given the methodological focus of this work, we do not provide a detailed clinical interpretation of each latent factor. However, we include the factor loadings to illustrate the model’s ability to uncover both intra- and inter-modality relationships.

Furthermore, following the idea presented in \cite{ferreira2022hierarchical}, for each latent factor $i$, we only report the modalities that explain at least 10$\%$ of the factor’s variance, defined by
\begin{equation}
    var_i = \frac{\mathbf{w_{\rm i,:}}^{(m)}\mathbf{w_{\rm i,:}}^{(m)\top}}{\mathbf{w_{\rm i,:}}^{\M}\mathbf{w_{\rm i,:}}^{\M\top}},
\end{equation}
where $\mathbf{w_{\rm i,:}}^{(m)}$ is the $(m)$-th modality of the $i$-th stable latent factor, and $\mathbf{w_{\rm i,:}}^{\M}$ all the modalities concatenated of the $i$-th stable factor. This selection is facilitated by the sparsity induced by the model’s priors. Furthermore, for each relevant modality, we display only the top 10 most relevant features according to the feature’s loadings. The captions of each figure describe the modalities represented within the corresponding figure. Table \ref{tab:general} describes and summarizes the preprocessing pipeline used for each modality while Table \ref{tab:descriptions} provides a description of the top features highlighted by each factor.

\begin{figure}[htbp]
    \centering
    \includegraphics[width=0.50\textwidth]{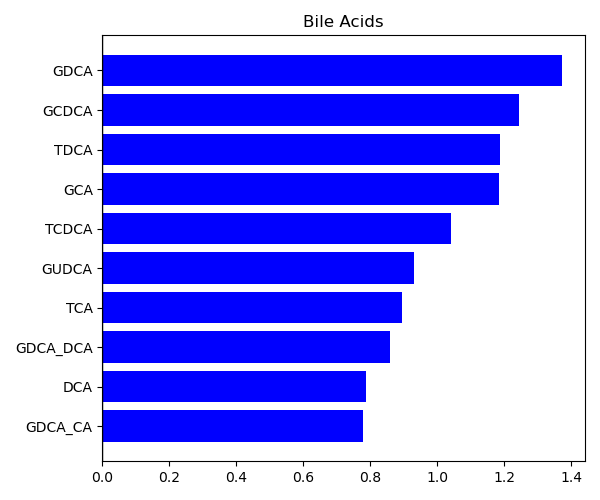}
    \caption{Factor that captures the variability of the bile acids measures across the sample.}
    \label{fig:bar_0}
\end{figure}

\begin{figure}[htbp]
    \centering
    \includegraphics[width=0.50\textwidth]{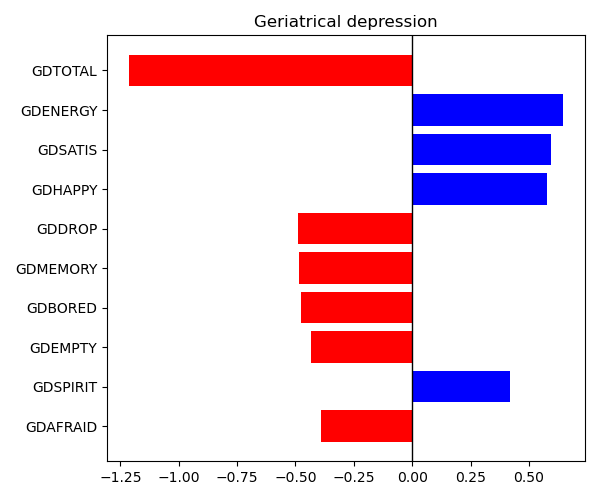}
    \caption{Factor that captures the variability of the depression measures across the sample.}
    \label{fig:bar_1}
\end{figure}


\begin{table}[htp]
    \centering
    \caption{General description and preprocessing methods applied to each modality used in the ADNI experiment.}
    \label{tab:general}
    \begin{adjustbox}{max width=\textwidth}
        \footnotesize
        \begin{tabular}
        {>{\centering\arraybackslash}m{3cm}  m{11cm}}
        \toprule
        \textbf{Modality} & \textbf{Description} \\
        \midrule
         \textbf{Bile Acids}& Bile acids were quantified from serum samples using the Biocrates Bile Acids assay, based on UPLC-MS/MS (ultra-high pressure
        liquid chromatography tandem with mass spectrometry) with Multiple Reaction Monitoring for high specificity. \\
        \hline
         \textbf{Geriatric depression} & Depression Scale (GDS-15) composed by 15-item self-report questionnaire administered during clinical visits. Participants answered yes/no questions reflecting mood over the past week. Total scores (0–15) were calculated according to ADNI protocol to assess depressive symptoms \cite{montorio1996geriatric}.  \\
         \hline
         \textbf{Montreal Cognitive Assessment}& Cognitive Assessment that evaluate multiple cognitive domains, including attention, executive function, memory, language, visuospatial skills, and orientation. Total scores range from 0 to 30, with lower scores indicating greater cognitive impairment \cite{nasreddine2005montreal}.  \\
         \hline
         \textbf{Hachinski Ischemia Scale}&   This scale, composed of 13 items, was designed to help differentiate vascular dementia from other dementia types \cite{hachinski1975cerebral}. The scale evaluates clinical history, cognitive symptoms, and vascular factors through clinician interviews and informant reports.\\
         \hline
         \textbf{Demographical data}& Demographic data, including age, sex, education level, and ethnicity, collected from ADNI participants during baseline clinical assessments.  \\
         \hline
         \textbf{\textbf{Neuro\-morphometrical ROIs}}& Structural MRI images were preprocessed using the CAT12 toolbox \cite{gaser2024cat}, followed by extraction of regional volumes for neuromorphometric ROIs based on the Harvard-Oxford atlas. These ROIs were segmented into three tissue classes: gray matter, white matter, and cerebrospinal fluid (CSF). Additionally, global brain measures were derived to complement the regional volumetric analysis.  \\
         \hline
         \textbf{Cortical Parcellation} & Structural MRI images were preprocessed using the CAT12 toolbox, followed by cortical parcellation based on the Destrieux atlas \cite{destrieux2010automatic} (dividing the brain into gyri and sulci).
         \\
         \hline
         \textbf{Amyloid PET} & Amyloid PET images were preprocessed with CAT12 toolbox. Regional ROIs were defined using the Desikan-Killiany atlas \cite{desikan2006automated} via FreeSurfer segmentation \cite{fischl2012freesurfer}. For all ROIs, standardized uptake value ratios (SUVR) were extracted to quantify amyloid burden, while volumetric measures obtained were normalized by the maximum activation value within each region. This combined approach enables analysis of both amyloid deposition and normalized regional volume.\\
        \bottomrule
        \end{tabular}
    \end{adjustbox}
\end{table}


\begin{landscape}
\begin{table}[htbp]
\centering
\scriptsize 
\caption{List of top features identified by the factors and their corresponding definitions.}
\renewcommand{\arraystretch}{1.1} 
\begin{tabularx}{\linewidth}{>{\raggedright\arraybackslash}p{0.08\linewidth}%
                          >{\raggedright\arraybackslash}X%
                          >{\raggedright\arraybackslash}p{0.08\linewidth}%
                          >{\raggedright\arraybackslash}X%
                          >{\raggedright\arraybackslash}p{0.08\linewidth}%
                          >{\raggedright\arraybackslash}X}
\toprule
\textbf{Feature} & \textbf{Description} &
\textbf{Feature} & \textbf{Description} &
\textbf{Feature} & \textbf{Description} \\
\midrule

\multicolumn{6}{l}{\textbf{Bile Acids}} \\[0.3em]
GDCA & Glycodeoxycholic Acid & GCDCA & Glycochenodeoxycholic Acid & TDCA & Taurodeoxycholic Acid \\
GCA & Glycocholic Acid & TCDCA & Taurochenodeoxycholic Acid & GUDCA & Glycoursodeoxycholic Acid \\
TCA & Taurocholic Acid & GDCA\_DCA & Ratio GDCA / DCA & DCA & Deoxycholic Acid \\
GDCA\_CA & Ratio GDCA / Cholic Acid & & & & \\

\multicolumn{6}{l}{\textbf{Geriatrical Depression (GDS)}} \\[0.3em]
GDTOTAL & Total score & GDENERGY & Do you feel full of energy? & GDSATIS & Are you basically satisfied with your life? \\
GDHAPPY & Do you feel happy most of the time? & GDDROP & Dropped activities/interests? & GDMEMORY & More problems with memory? \\
GDBORED & Do you often get bored? & GDEMPTY & Do you feel life is empty? & GDSPIRIT & In good spirits most of the time? \\
GDAFRAID & Afraid something bad will happen? & GDHOPE & Do you feel your situation is hopeless? & GDHOME & Prefer to stay at home? \\
GDUNABL & Unable to complete GDS & GDBETTER & Do you think most people are better off than you? & GDHELP & Do you often feel helpless? \\

\multicolumn{6}{l}{\textbf{Montreal Cognitive Assessment (MoCA)}} \\[0.3em]
RHINO & Rhinoceros & SERIAL1 & Serial 7: 1st subtraction & FFLUENCY & Letter Fluency (F) \\
CUBE & Copy cube & LETTERS & List of letters – errors & CLOCKHAN & Clock drawing – hands \\
SERIAL5 & Serial 7: 5th subtraction & IMMT2W5 & Immediate \#2: Red & ABSTRAN & Abstraction: train–bicycle \\
DIGBACK & Digits backward & DELW2 & Delayed: Church & DELW3 & Delayed: Daisy \\
SERIAL2 & Serial 7: 2nd subtraction & LION & Lion & IMMT2W2 & Immediate \#2: Velvet \\
SERIAL3 & Serial 7: 3rd subtraction & IMMT1W3 & Immediate \#1: Church & DELW1 & Delayed: Face \\
IMMT2W3 & Immediate \#2: Church & REPEAT1 & Repeat sentence & IMMT1W4 & Immediate \#1: Daisy \\
DELW4 & Delayed: Daisy & CLOCKNO & Clock contour & IMMT2W4 & Immediate \#2: Daisy \\
IMMT1W2 & Immediate \#1: Velvet & IMMT2W1 & Immediate \#2: Face & IMMT2W2 & Immediate \#2: Velvet \\

\multicolumn{6}{l}{\textbf{Demographics}} \\[0.3em]
PTETHCAT & Ethnicity category & PTRACCAT & Race category & PTMARRY & Married situation \\
PTGENDER & Gender & PTEDUCAT & Level of education & & \\

\multicolumn{6}{l}{\textbf{General Measurements}} \\[0.3em]
surf\_TSA & Total surface area & vol\_tiv & Total intracranial volume & vol\_abs\_WMH & WMH absolute volume \\
IQR & Interquartile range & & & & \\

\multicolumn{6}{l}{\textbf{Hachinski Ischemia Scale}} \\[0.3em]
HMONSET & Abrupt dementia onset & HMSTEPWS & Stepwise deterioration & HMSOMATC & Somatic complaints \\
HMEMOTIO & Emotional incontinence & HMHYPERT & History of hypertension & HMSTROKE & History of stroke \\
HMNEURSM & Focal neurologic symptoms & HMNEURSG & Focal neurologic signs & HMSCORE & Total score \\
SOURCE & Information source & & & & \\

\bottomrule
\end{tabularx}
\label{tab:descriptions}
\end{table}
\end{landscape}

\begin{figure}[htbp]
    \centering
    \includegraphics[width=\textwidth]{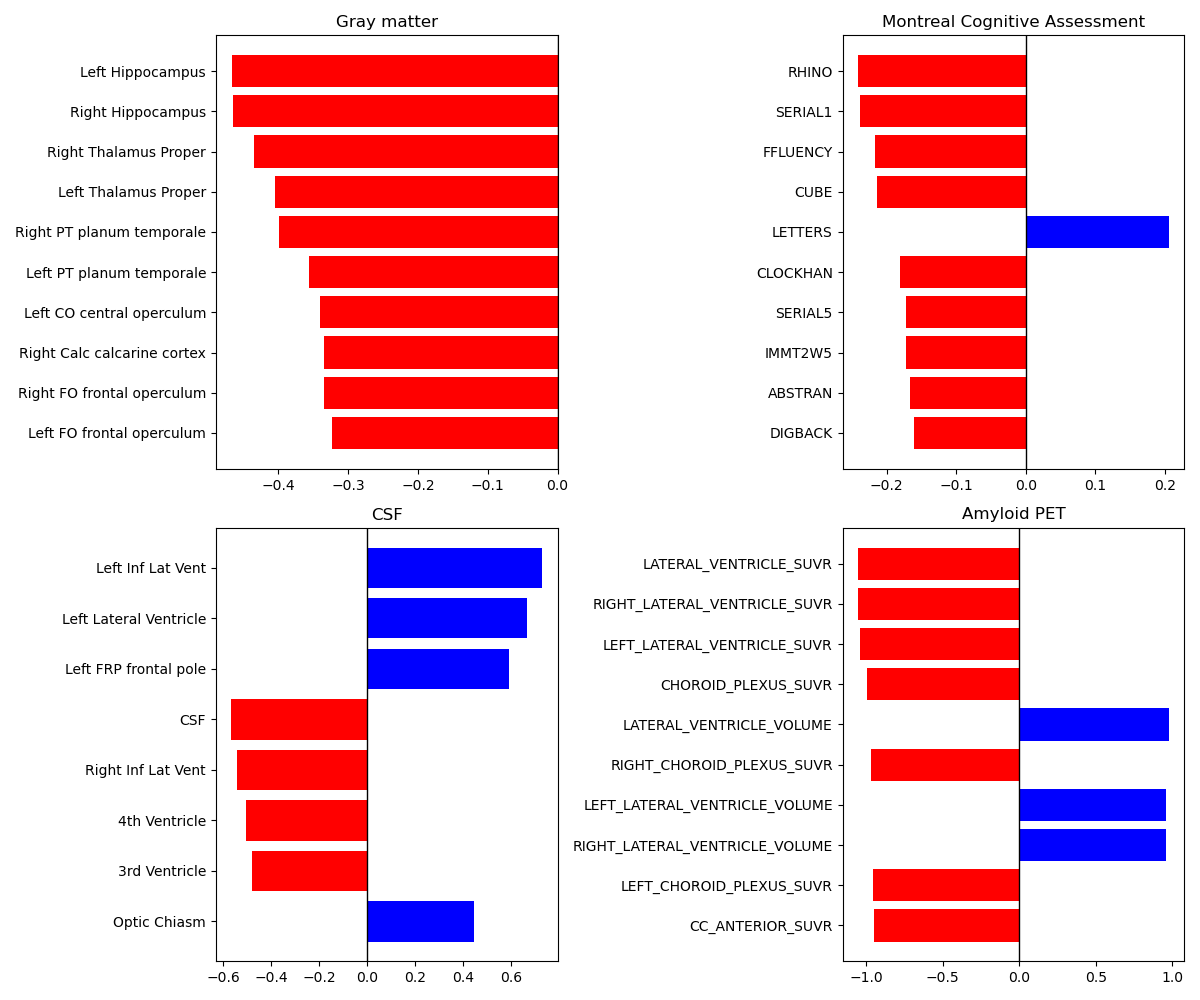}
    \caption{Factor that captures an association between gray matter volumes, CSF volumes, Amyloid PET measures in several brain regions, and cognitive assessments (attention, memory, language, etc) from the Montreal Cognitive Assessment. List of acronyms: CC (Corpus Callosum), and SUVR (Standarized Uptake Value Ratio).}
    \label{fig:bar_2}
\end{figure}

\begin{figure}[htbp]
    \centering
    \includegraphics[width=\textwidth]{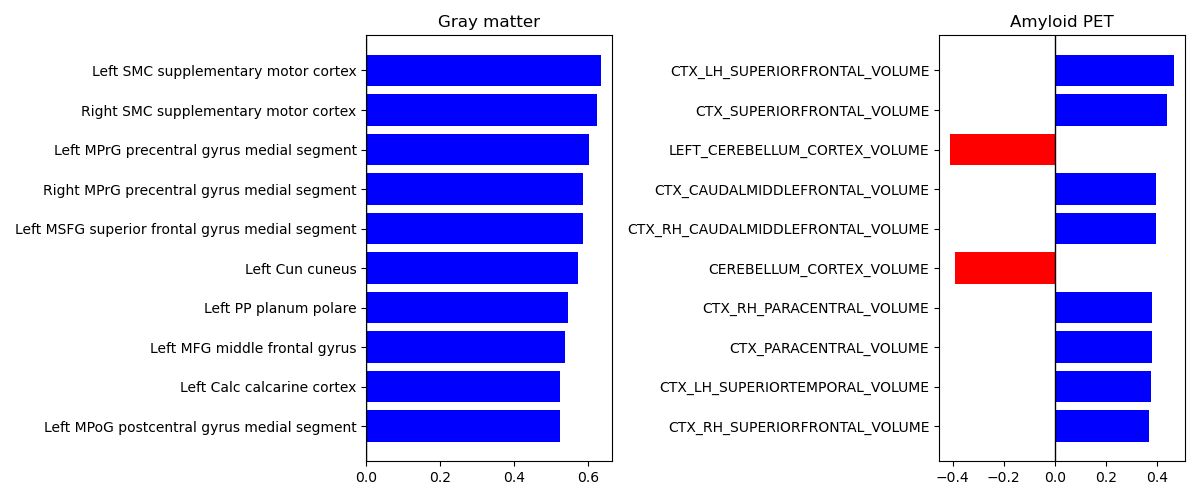}
    \caption{Factor that captures associations between neuroimaging modalities: gray matter volumes and Amyloid PET. List of acronyms: CTX (cortex), RH (right hemisphere), and LH (left hemisphere).}
    \label{fig:bar_3}
\end{figure}

\begin{figure}[htbp]
    \centering
    \includegraphics[width=0.50\textwidth]{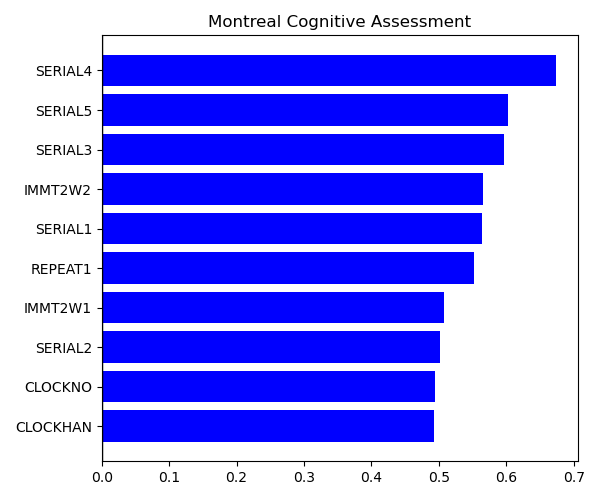}
    \caption{Factor that captures the cognitive variability across the sample.}
    \label{fig:bar_4}
\end{figure}

\begin{figure}[htbp]
    \centering
    \includegraphics[width=\textwidth]{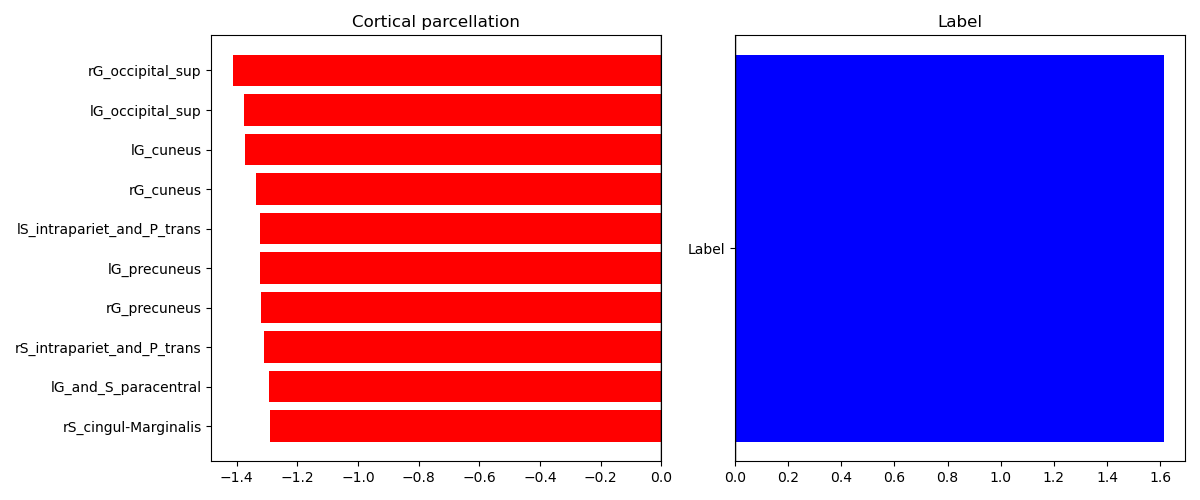}
    \caption{Factor that captures associations between the label (MCI) and certain cortical regions. List of acronyms: r (right), l (left), G (gyrus), S (sulcus), P (Parietal), sup (superior), and trans (transverse).}
    \label{fig:bar_5}
\end{figure}

\begin{figure}[htbp]
    \centering
    \includegraphics[width=0.50\textwidth]{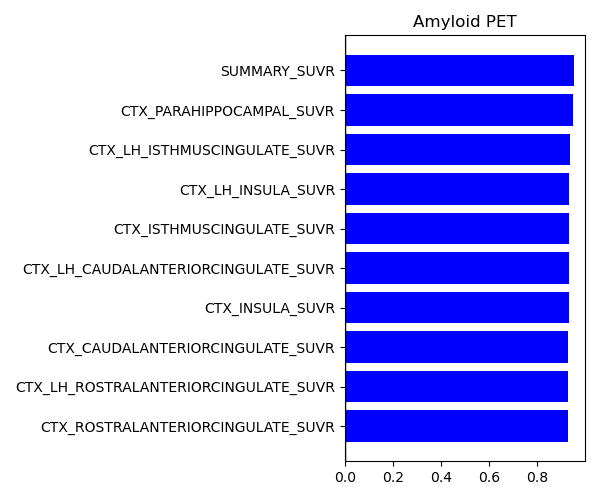}
    \caption{Factors that describe the variability of the Amyloid PET measures across the sample. List of acronyms: CTX (cortex), LH (left hemisphere), and SUVR (Standarized Uptake Value Ratio).}
    \label{fig:bar_6}
\end{figure}

\begin{figure}[htbp]
    \centering
    \includegraphics[width=\textwidth]{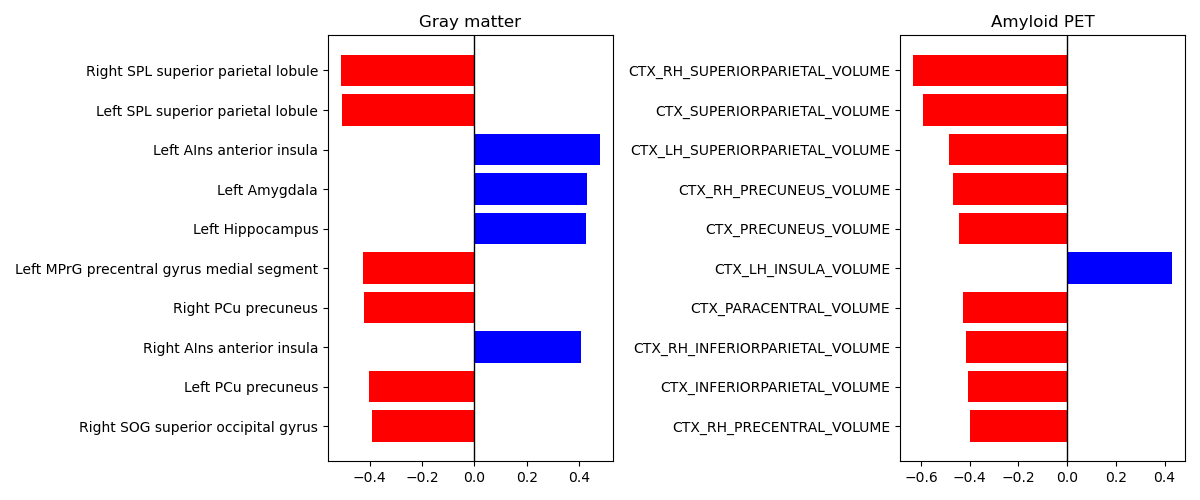}
    \caption{Factor that captures associations between gray matter volume in certain brain regions and Amyloid PET measures. List of acronyms: CTX (cortex), RH (right hemisphere), and LH (left hemisphere).}
    \label{fig:bar_7}
\end{figure}

\begin{figure}[htbp]
    \centering
    \includegraphics[width=\textwidth]{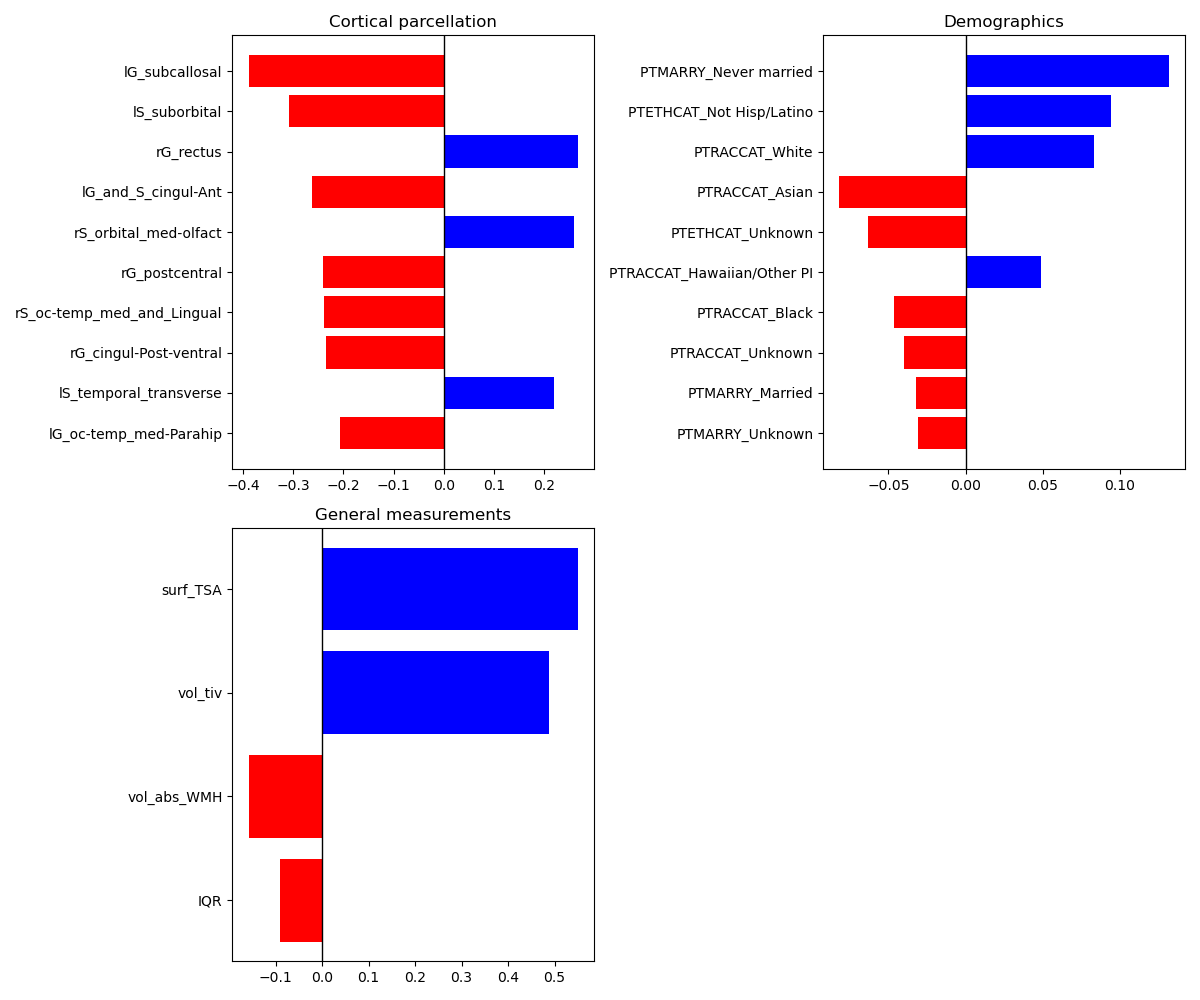}
    \caption{Factor that captures associations between the volume in certain cortical regions, general measurements (related to brain volume), and demographic information. List of acronyms: r (right), l (left), G (gyrus), S (sulcus), P (Parietal), ant (anterior), temp (temporal), and med (medial).}
    \label{fig:bar_8}
\end{figure}

\begin{figure}[htbp]
    \centering
    \includegraphics[width=0.9\textwidth]{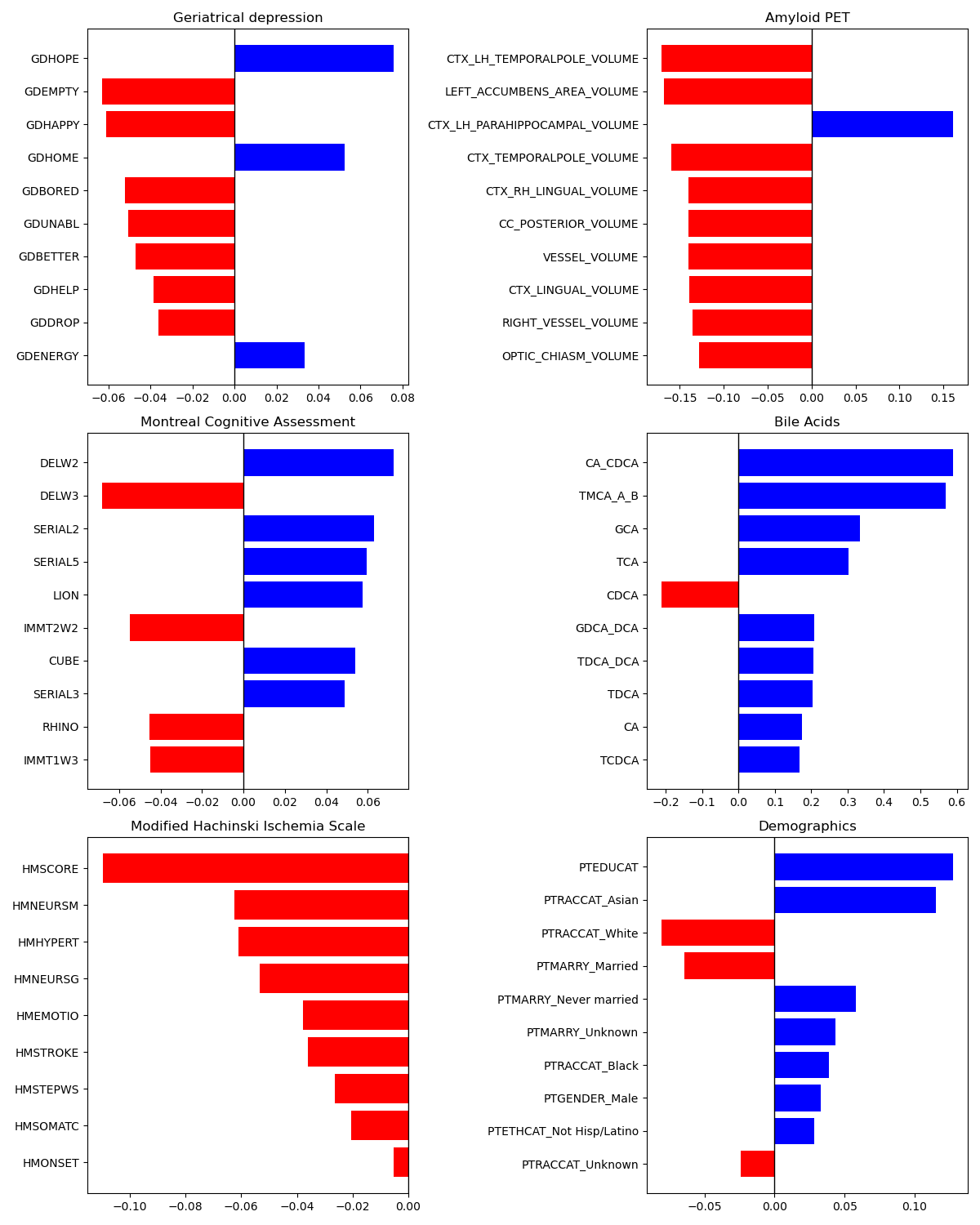}
    \caption{Factor that captures associations between psychological measures from the Geriatric Depression Scale, cognitive measures from the Montreal Cognitive Assessment, demographic data, clinical measures (Hachinski Ischemia Scale and bile acid) and neuroimaging measures (Amyloid PET). List of acronyms: CTX (cortex), RH (right hemisphere), and LH (left hemisphere).}
    \label{fig:bar_9}
\end{figure}

\begin{figure}[htbp]
    \centering
    \includegraphics[width=\textwidth]{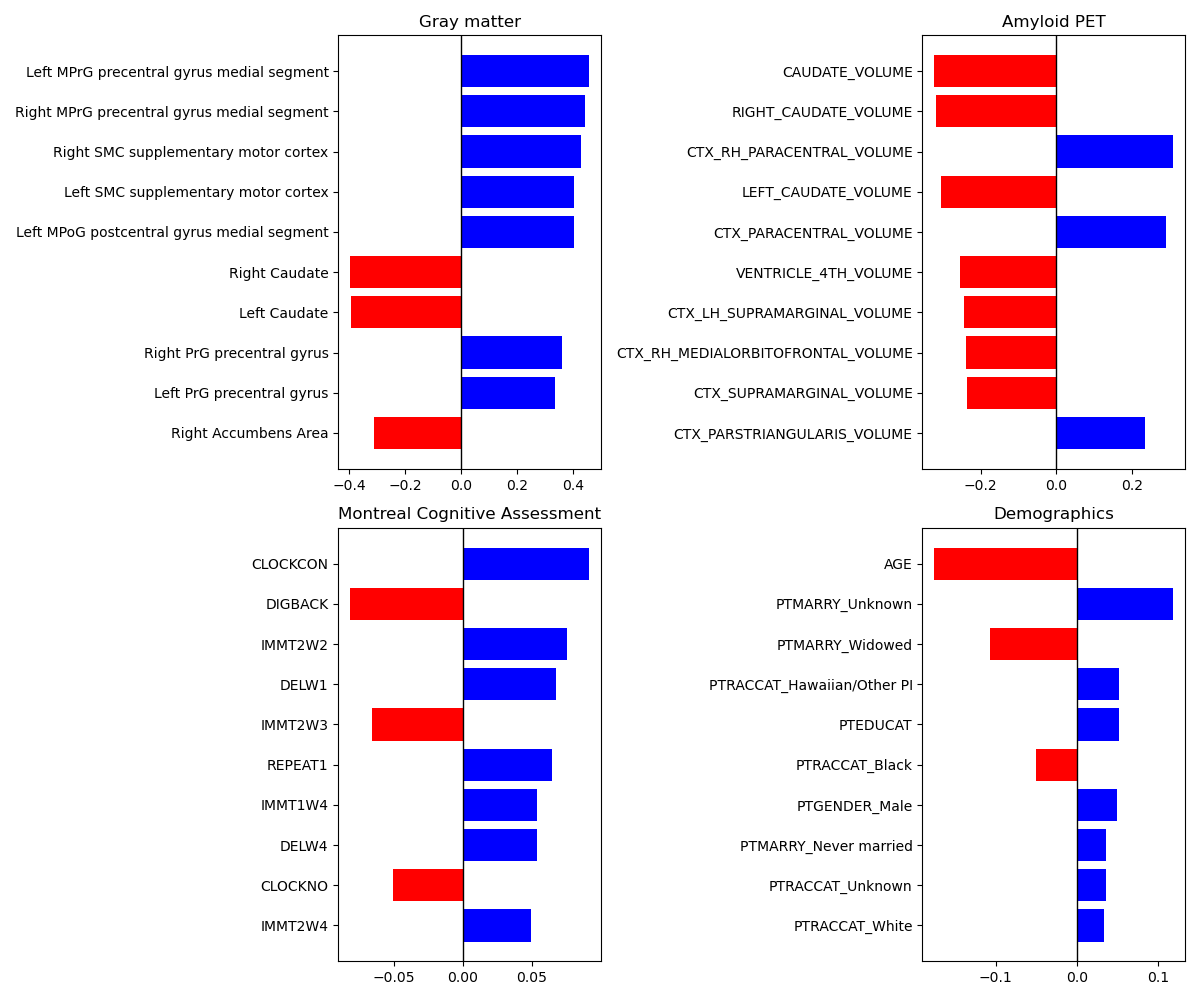}
    \caption{Factor that captures associations between cognitive measures from the Montreal Cognitive Assessment, demographic and neuroimaging measures (such as gray matter volumes and amyloid PET measures). List of acronyms: CTX (cortex), RH (right hemisphere), and LH (left hemisphere).}
    \label{fig:bar_10}
\end{figure}

\begin{figure}[htbp]
    \centering
    \includegraphics[width=0.9\textwidth]{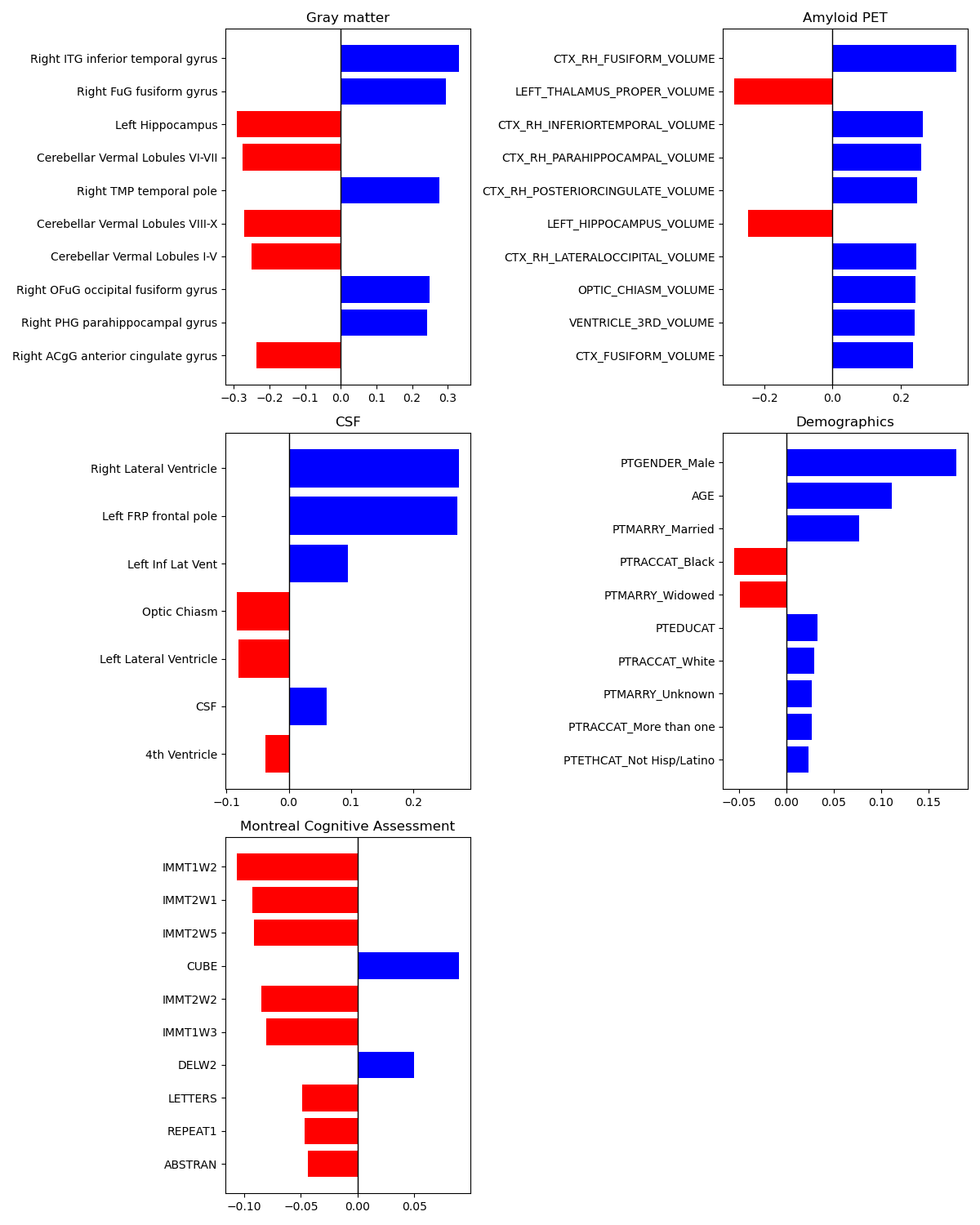}
    \caption{Factor that captures associations between cognitive measures from the Montreal Cognitive Assessment, demographic, and neuroimaging data (such as gray matter volumes, CSF, and amyloid PET measures). List of acronyms: CTX (cortex), and RH (right hemisphere).}
    \label{fig:bar_11}
\end{figure}

\begin{figure}[htbp]
    \centering
    \includegraphics[width=\textwidth]{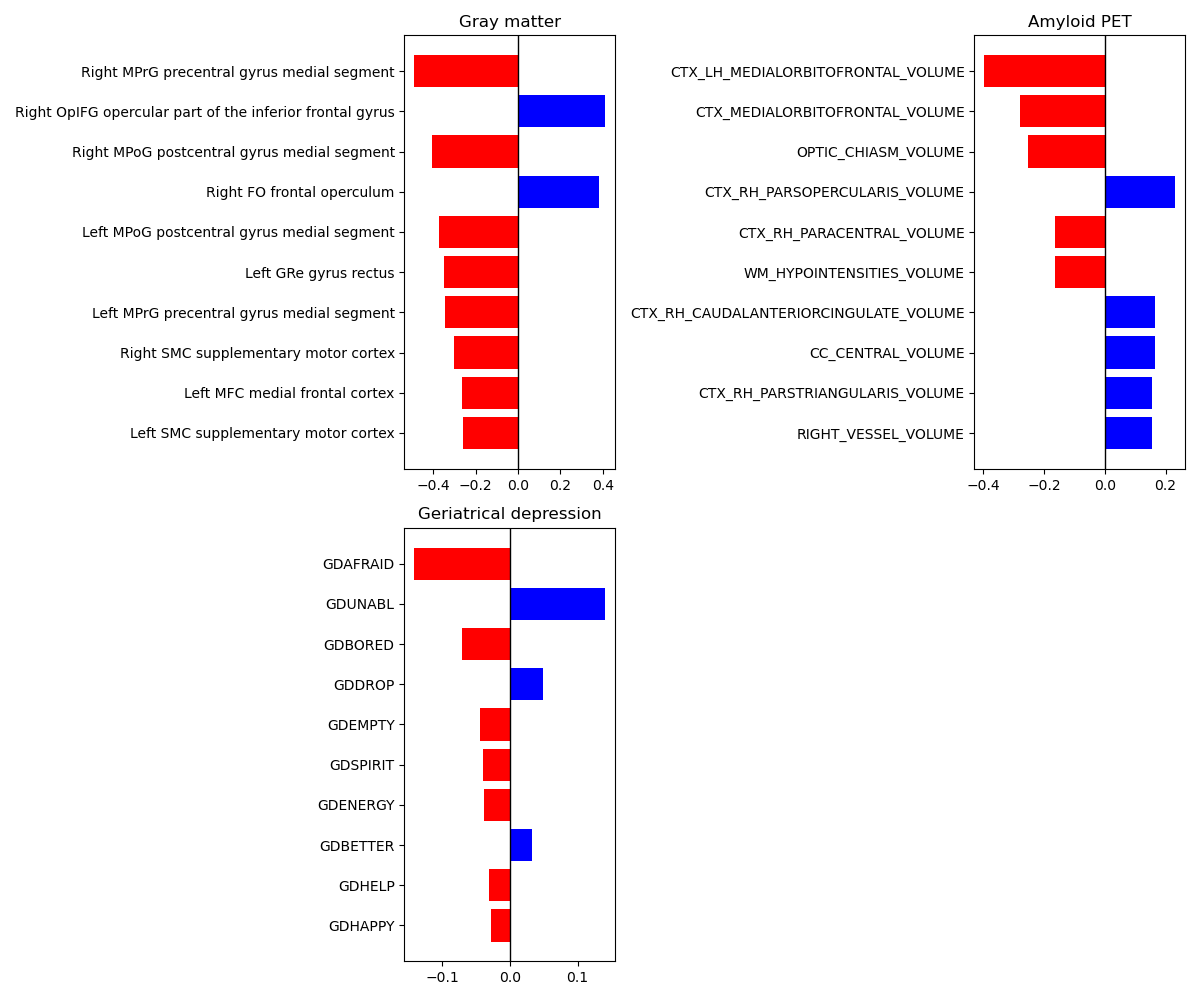}
    \caption{Factor that captures associations between psychological measures from the Geriatric Depression Scale and neuroimaging measures (such as gray matter ROIs and amyloid PET measures). List of acronyms: CTX (cortex), RH (right hemisphere), LH (left hemisphere), and WM (white matter).}
    \label{fig:bar_12}
\end{figure}

\bibliographystyle{abbrv}
\bibliography{bibliography.IF.bib}

\end{document}